\documentclass[10pt,twocolumn,letterpaper]{article}

\PassOptionsToPackage{table}{xcolor} %
\usepackage[pagenumbers]{cvpr} %

\definecolor{cvprblue}{rgb}{0.21,0.49,0.74}
\usepackage[pagebackref,breaklinks,colorlinks,allcolors=cvprblue]{hyperref}

\makeatletter
\renewcommand\paragraph{\@startsection{paragraph}{4}{\z@}{0.2ex}{-1em}{\normalfont\normalsize\bfseries}}
\makeatother
\usepackage[utf8]{inputenc}
\usepackage{newunicodechar}
\newunicodechar{−}{-}
\newunicodechar{’}{\textquoteright}
\newunicodechar{‘}{\textquoteleft}
\newunicodechar{“}{\textquotedblleft}
\newunicodechar{”}{\textquotedblright}
\newunicodechar{–}{\textendash}
\newunicodechar{—}{\textemdash}
\usepackage{textcomp}
\usepackage[table]{xcolor}
\usepackage[capitalize]{cleveref}
\crefname{researchquestion}{research question}{research questions}
\Crefname{researchquestion}{Research Question}{Research Questions}

\usepackage{graphicx}
\usepackage{adjustbox}
 
\usepackage{xcolor}
\usepackage{dashrule}
\usepackage[most]{tcolorbox}
\newlength{\gap}

\newlength{\colw}

\definecolor{promptbg}{HTML}{FDEBD3}      %
\definecolor{promptaccent}{HTML}{C97A2B}  %

\newtcbox{\promptbox}[1][]{%
  on line, boxrule=0pt, colback=promptbg, colframe=promptbg,
  left=4pt, right=4pt, top=2pt, bottom=2pt,
  boxsep=0pt, arc=2pt,
  leftrule=2pt, colframe=promptaccent,
  #1
}

\newcommand{\hficon}{%
  \raisebox{-0.2\height}{\includegraphics[height=0.95em]{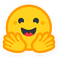}}%
}

\newcommand{\prompttext}[1]{%
  \promptbox{\scriptsize\textcolor{promptaccent}{\textbf{Expression:}}\ \itshape #1}%
}

\newtcolorbox{promptblock}[1][]{%
  enhanced, breakable, colback=promptbg, colframe=promptaccent,
  boxrule=0pt, leftrule=2pt, arc=2pt,
  left=4pt, right=4pt, top=2pt, bottom=2pt,
  before skip=4pt, after skip=4pt,
  fontupper=\small\itshape,
  #1
}

\newcommand{\emojiovb}{\raisebox{-0.2ex}{\includegraphics[height=1.2em]{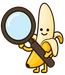}}}

\newcommand{\fluxknineb}{FLUX.2 Klein 9B Base}
\newcommand{\fluxkninebshort}{F2K9B Base}
\newcommand{\fluxkfourb}{FLUX.2 Klein 4B Base}
\newcommand{\fluxkfourbshort}{F2K4B Base}

\usepackage{booktabs}
\usepackage{graphicx}
\usepackage{xcolor}
\usepackage{pifont}
\newcommand{\cmark}{\ding{51}}
\usepackage{amsmath,amssymb}
\usepackage{float}
\usepackage{pifont}
\usepackage{multirow}
\usepackage{makecell}
\usepackage{tabularx}
\usepackage{array}
\usepackage{fontawesome5}
\usepackage[T1]{fontenc}

\newcommand{\proprietary}{{\scriptsize\textcolor{BrickRed}{\faLock}}}
\newcommand{\public}{{\scriptsize\textcolor{ForestGreen}{\faLockOpen}}}

\newcommand{\modelfull}{Open Vision Banana (OVB)}
\newcommand{\modellong}{Open Vision Banana}
\newcommand{\model}{OVB}
\newcommand{\best}[1]{\textbf{#1}}
\newcommand{\second}[1]{\underline{#1}}

\newcommand{\na}{--}

\newcolumntype{Y}{>{\centering\arraybackslash}X}

\newcounter{researchquestion}

\newcommand{\rqprefix}{}

\newcommand{\rqcolor}{gray}

\newcommand{\setrqprefix}[2]{%
  \renewcommand{\rqprefix}{#1}%
  \renewcommand{\rqcolor}{#2}%
  \setcounter{researchquestion}{0}%
  \renewcommand{\theresearchquestion}{%
    \rqprefix\arabic{researchquestion}%
  }%
}

\newcommand{\researchquestion}[2]{%
  \refstepcounter{researchquestion}%
  \label{rq:#1}%
  \par\smallskip
  \noindent
  \colorbox{\rqcolor!15}{%
    \parbox{\dimexpr\linewidth-2\fboxsep\relax}{%
      \textbf{RQ-\theresearchquestion.}\ %
      \emph{#2}%
    }%
  }%
  \par\smallskip
}

\newcommand{\rqref}[1]{%
  \hyperref[rq:#1]{RQ-\ref*{rq:#1}}%
}

\usepackage[most]{tcolorbox}   %

\newcommand{\takeaway}[2]{%
  \begin{tcolorbox}[
    enhanced, breakable,
    colback=\rqcolor!8,
    frame hidden,          %
    sharp corners,
    boxsep=0pt,
    left=\fboxsep, right=\fboxsep,
    top=\fboxsep,  bottom=\fboxsep,
    before skip=\smallskipamount,
    after skip=\smallskipamount,
  ]
    \textbf{Takeaway
      (\hyperref[rq:#1]{\ref*{rq:#1}}).}\ %
    #2%
  \end{tcolorbox}%
}

\newcommand{\rqcaption}[2][]{%
  \if\relax\detokenize{#1}\relax
    \caption{#2}%
  \else
    \captionsetup{labelsep=space}%
    \caption{\protect\hyperref[rq:#1]{(\protect\ref*{rq:#1})}. #2}%
  \fi
}

\usepackage{xcolor}
\usepackage{colortbl}
\usepackage{pgf}

\newcommand{\HeatRange}[3]{%
  \expandafter\def\csname heatmin@#1\endcsname{#2}%
  \expandafter\def\csname heatmax@#1\endcsname{#3}%
}

\newcommand{\Heat}[2]{%
  \pgfmathparse{10 + 40*(#2-\csname heatmin@#1\endcsname)/
    (\csname heatmax@#1\endcsname-\csname heatmin@#1\endcsname)}%
  \edef\temp{%
    \noexpand\cellcolor{\rqcolor!\pgfmathresult}%
  }%
  \temp #2%
}

\definecolor{defaultcolor}{RGB}{220,220,220}
\definecolor{segmentationreferencecolor}{RGB}{220,220,220}
\definecolor{geometryreferencecolor}{RGB}{252,222,195}
\definecolor{qiecolor}{HTML}{E1F0DA}
\definecolor{ovbzeroshot}{RGB}{249,240,206}      %
\definecolor{ovbnonzeroshot}{RGB}{206,226,249}  %

\def\confName{CVPR}
\def\confYear{2026}

\title{Open Vision Banana~\emojiovb: A Fully Open Recipe for Generative Visual Perception}

\author{
Arpit Jadon$^{1,2}$\thanks{Equal contribution.}
\qquad
Valentinos Pariza$^{1*}$
\qquad
Jona Ruthardt$^{1}$
\\[0.5em]
Ryousuke Yamada$^{1,3}$
\qquad
A. Sophia Koepke$^{4,5}$
\qquad
Andrei Bursuc$^{6}$
\qquad
Yuki M. Asano$^{1}$
\\[0.8em]
{\footnotesize
$^1$Fundamental AI Lab, University of Technology Nuremberg
\enspace
$^2$DLR
\enspace
$^3$AIST
\enspace
$^4$University of Tübingen, Tübingen AI Center
\enspace
$^5$TUM, MCML
\enspace
$^6$valeo.ai}
}

\begin{document}

\setlength{\floatsep}{10pt plus 2pt minus 2pt}
\setlength{\dblfloatsep}{10pt plus 2pt minus 2pt}
\setlength{\textfloatsep}{12pt plus 2pt minus 2pt}
\setlength{\dbltextfloatsep}{12pt plus 2pt minus 2pt}
\setlength{\intextsep}{6pt plus 1pt minus 1pt}

\twocolumn[{%
\renewcommand\twocolumn[1][]{#1}%
\vspace{-5.5em} %
\maketitle

\vspace{-3.0em}
\begin{center}
    \captionsetup{type=figure}

    \includegraphics[width=\textwidth]{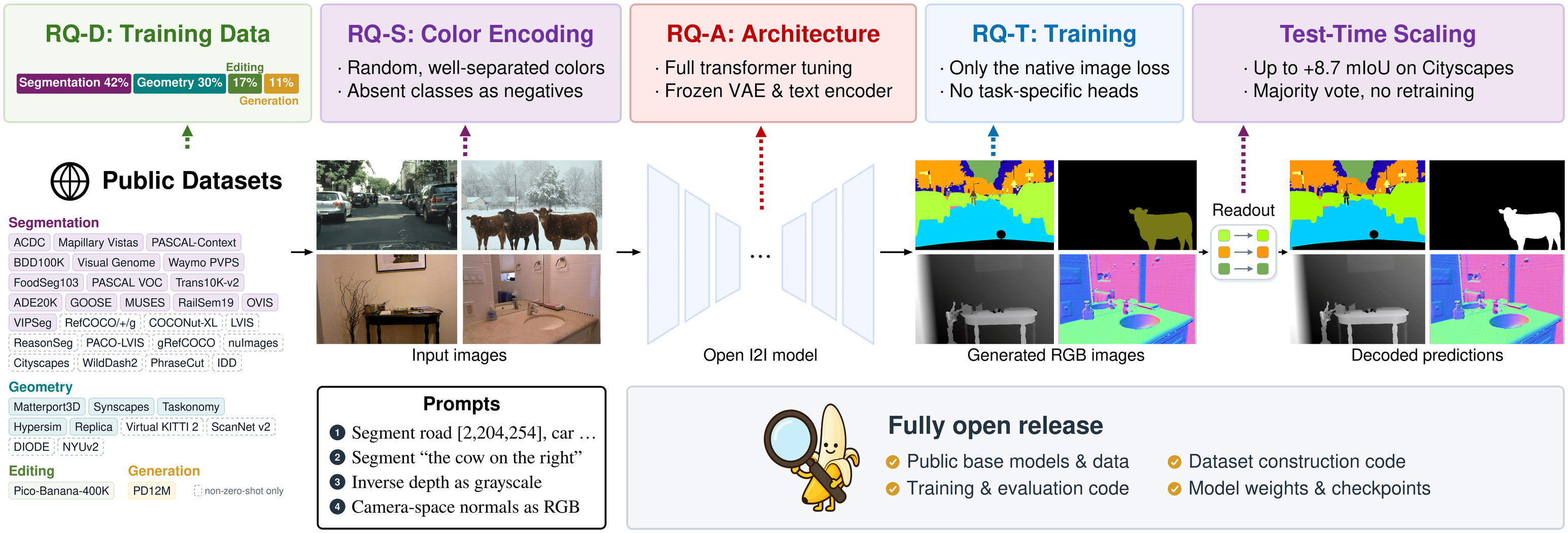}

    \vspace{-4pt}
    {\color{black!25}\hdashrule{0.995\textwidth}{0.4pt}{3pt 3pt}}
    \vspace{-4pt}

    \definecolor{teaserseg}{HTML}{6C4AB6}
\definecolor{teasergeo}{HTML}{1F8A70}
\definecolor{teasergen}{HTML}{D9730D}
\ifdefined\teaserw\else\newlength{\teaserw}\fi
\setlength{\teaserw}{0.118\textwidth}
\providecommand{\teaserchip}[2]{%
  \node[anchor=south,fill=#1,rounded corners=2pt,inner xsep=3pt,inner ysep=1.5pt,
    text=white,text height=1.5ex,text depth=0.25ex,font=\tiny\bfseries\sffamily]
    at (0.5\teaserw,1.3333\teaserw+2pt) {#2};}
\providecommand{\teasercaption}[1]{%
  \node[anchor=north,inner sep=0pt,text width=\teaserw,align=center,
    font=\fontsize{5.5}{6.5}\selectfont\itshape,text=black!70]
    at (0.5\teaserw,-2.5pt) {#1};}
\providecommand{\teasercell}[4]{%
  \begin{tikzpicture}[baseline=(current bounding box.north)]
    \begin{scope}
      \clip[rounded corners=4pt] (0,0) rectangle (\teaserw,1.3333\teaserw);
      \node[anchor=south west,inner sep=0] at (0,0)
        {\includegraphics[width=\teaserw,height=1.3333\teaserw]{#3}};
    \end{scope}
    \draw[#1,line width=0.8pt,rounded corners=4pt] (0,0) rectangle (\teaserw,1.3333\teaserw);
    \teaserchip{#1}{#2}
    \teasercaption{#4}
  \end{tikzpicture}}
\providecommand{\teaserplaceholder}[3]{%
  \begin{tikzpicture}[baseline=(current bounding box.north)]
    \draw[#1!60,fill=#1!6,line width=0.8pt,dashed,rounded corners=4pt]
      (0,0) rectangle (\teaserw,1.3333\teaserw);
    \node[font=\tiny\sffamily,text=#1!80] at (0.5\teaserw,0.6667\teaserw) {placeholder};
    \teaserchip{#1}{#2}
    \teasercaption{#3}
  \end{tikzpicture}}
\noindent\makebox[\textwidth][c]{%
  \teasercell{black!70}{Input}{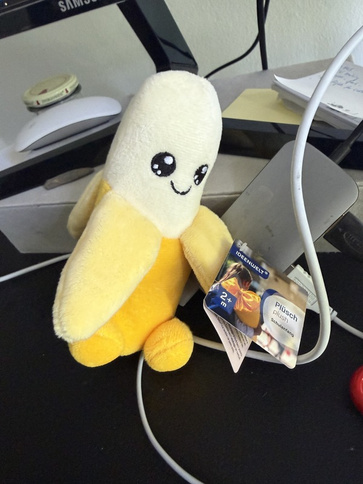}{}\hfill
  \teasercell{teaserseg}{Semantic seg.}
{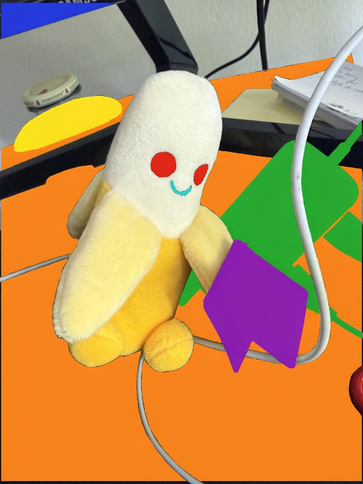}{``table, price tag, plushie eyes, screen, mouse, plushie smile, connection device''}\hfill
  \teasercell{teaserseg}{Instance seg.}{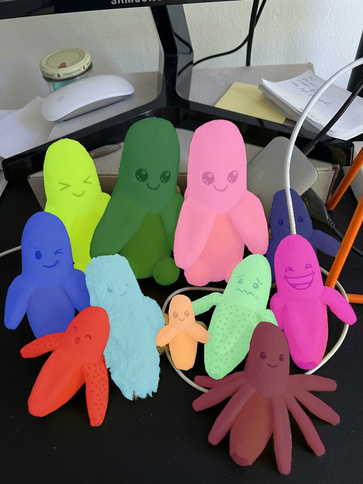}{``plushie''}\hfill
  \teasercell{teaserseg}{Referring seg.}{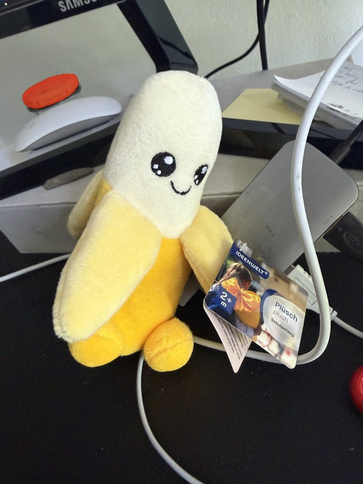}{``which round object behind the mouse could you use as a coaster?''}\hfill
  \teasercell{teasergeo}{Depth}{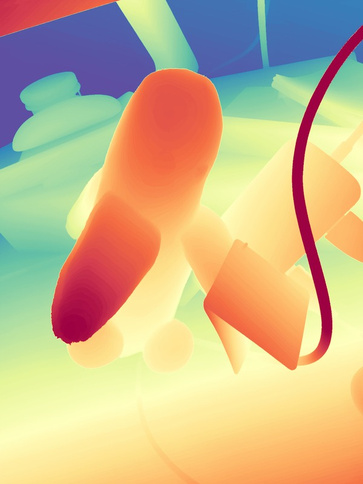}{``estimate a \mbox{relative} inverse-depth map''}\hfill
  \teasercell{teasergeo}{Surface normals}{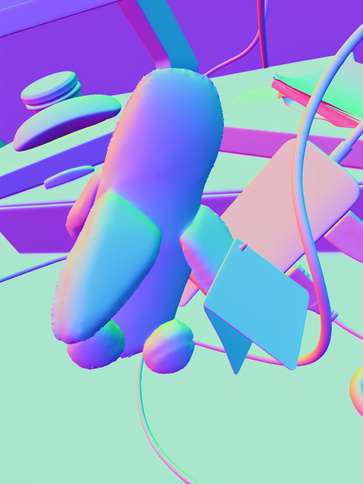}{``create a camera-space surface-normal map as RGB''}\hfill
  \teasercell{teasergen}{Image editing}{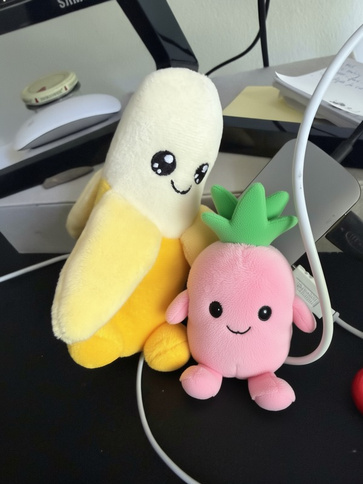}{``remove the price tag and add a pink pineapple plushie''}\hfill
  \teasercell{teasergen}{Generation}{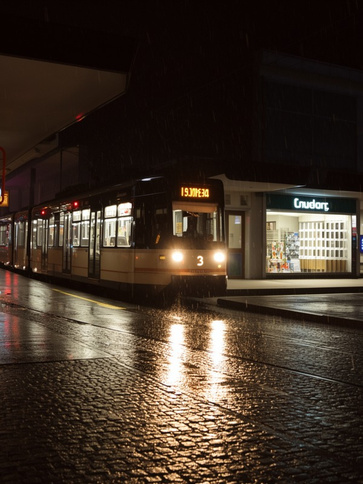}{``an empty tram at a cobblestone stop on a rainy night''}%
}

    \vspace{-0.75em}
    \caption{\textbf{Overview of the \modelfull{} recipe and analysis framework.}
    Open image generators are adapted on public datasets to express perception
    outputs as RGB images. The research questions (RQs, \cref{sec:ablations})
    cover target design, data composition, and model adaptation. Further analyses on
    prompt design, data scaling, and training duration are in the appendix. \emph{Bottom:} predictions of the zero-shot
    \model{} 9B on in-the-wild photos, with a shortened prompt version below the output (full templates in Sec.~\ref{sec:prompt-templates}).}
    \label{fig:analysis_overview}
\end{center}
}]

\begin{abstract}
\vspace{-2.5em}

Recent work suggests that pretrained image generators encode real-world understanding beyond image synthesis. Vision Banana exploits this by framing vision tasks as conditional image editing. However, its data, weights, code, and implementation details are not public. We present \modelfull{}, a fully open and reproducible recipe for converting pretrained generators into unified multi-task perception systems. 
Our analyses identify the design choices that matter most for generalization, including color encoding, prompt design, parameter adaptation, and task balancing. These choices interact, so individually helpful choices can hurt when combined with others.%

Starting from 4B and 9B parameter open-weight base models and using 2.1M--3.8M training pairs from public datasets, \model{} surpasses Vision Banana's reported Cityscapes zero-shot performance (73.9 vs.\ 69.9 mIoU) and rivals state-of-the-art affine-invariant depth estimators. Its non-zero-shot variants come within five points of Vision Banana on referring and reasoning segmentation but remain further behind on instance segmentation. All models largely retain the base model's image-editing abilities. 
With legend ensembling, we introduce a test-time scaling method for generator-based perception. It segments an image under several class--color legends and fuses the results by majority vote, correcting unstable color--class bindings and improving mIoU by up to $8.7$ points on Cityscapes and $15.0$ points on COCO-Object without retraining. 
We release the dataset construction and processing pipeline, the training and evaluation code\iftoggle{cvprfinal}{~\href{https://github.com/OpenVisionBanana/OpenVisionBanana}{\faGithub}}{}, as well as the final, intermediate, and ablation checkpoints\iftoggle{cvprfinal}{~\href{https://huggingface.co/FunAILab/Open-Vision-Banana}{\hficon}}{}.

\iftoggle{cvprfinal}{%
\noindent\textup{\textbf{Project website:} \href{https://openvisionbanana.github.io}{\nolinkurl{openvisionbanana.github.io}}}
}{}
\end{abstract}

\vspace{-1.25em}
\section{Introduction}
\label{sec:intro}

Image generation models have become increasingly capable of producing realistic and coherent visual content \citep{flux-2-2025, wu2025qwenimagetechnicalreport}. A growing line of work argues that this capability reflects general visual understanding that can be utilized for perception tasks~\citep{khangaonkar2026gen2seggenerativemodelsenable,ke2025marigoldaffordableadaptationdiffusionbased,zhao2025diceptiongeneralistdiffusionmodel,jadon2026test}.

Vision Banana~\citep{visionbanana2026} demonstrates that a sufficiently capable image generator can be adapted into a generalist visual perception model by rendering predictions as RGB images. 
It rivals domain specialists on segmentation and geometry while largely keeping the generator's editing abilities. However, its base model, data and code are proprietary, so it is unclear which components account for these results. Does the paradigm transfer to publicly available image generators, and how do design choices not elaborated in the Vision Banana report affect the resulting model? 

We introduce \textbf{\modelfull{}}~\emojiovb, a fully reproducible recipe and set of models for generative visual perception.
To derive this recipe, we systematically study the design choices that turn an image generator into a perception model. As shown in \cref{fig:analysis_overview}, the analysis spans the type and composition of training data, target representations and prompt design, the choice of parameters to adapt, and inference-time readout (decoding generated RGB raw output to task specific final format).

Our experiments show that these choices have large effects and can depend on one another. A model may locate the right region yet score poorly because it paints the wrong color or breaks an output convention. In particular, low-level decisions such as the palette and prompt wording strongly affect accuracy. Supervision for one task can help or hurt others, and choices that help individually may no longer help when combined (Appendix~\ref{sec:appendix-combined-choices}).

Finally, we introduce \emph{legend ensembling}, which segments an image under several randomly drawn color legends and fuses the passes by majority vote. Without retraining, it improves the zero-shot 9B model by up to 8.7 mIoU on Cityscapes and 15.0 mIoU on COCO-Object, with gains mostly from fixing colors bound to wrong classes.

\modellong{} makes four main contributions: 
\begin{itemize}
    \item \textbf{A fully open recipe} for adapting public image generators to four segmentation tasks, depth estimation, and surface normal estimation. The models are trained exclusively on public data, and we release the complete data pipeline, training code, and model checkpoints.
    \item \textbf{Controlled experiments} on performance-critical design choices, with an analysis of failure modes.
    \item \textbf{Legend ensembling}, a training-free test-time scaling method for generator-based segmentation.
    \item \textbf{Competitive open models} that surpass Vision Banana on Cityscapes, come close to it on referring and reasoning segmentation, perform well on depth and surface-normals, and largely retain their editing abilities.
\end{itemize}

\section{\modellong{} Recipe}
\label{sec:recipe}

\model{} turns a pretrained image generator into a perception model without task-specific heads: given an input image $x$ and an instruction $p$, it generates an RGB image $y$ that a deterministic, task-specific readout converts into a prediction.

\subsection{Base Generators}
\label{sec:recipe_base_models}

We release fine-tuned FLUX.2-Klein models at 4B and 9B scale~\cite{flux-2-2025}, open-weight generators that condition on an input image. All transformer parameters are trained, from the input and context embedders to the output layers. The VAE and text encoder stay frozen.

\subsection{RGB Task Modeling} %
\label{sec:recipe_task_interface}

Each task has a deterministic encoder that renders an annotation as an RGB target and a decoder that reads a generated image back into a prediction. Both use per-sample metadata such as the color legend, depth scaling and resizing.

\subsubsection{Color Mapping for Segmentation Tasks}
\label{sec:color_map_seg}

Prompts specify colors as RGB triplets, hexadecimal codes (64 most-separated colors of a 200-color palette, \cref{rq:color-segmentation}), or as one of 37 common color names. Black is reserved for background. Prompt templates and color notation vary across samples.

\paragraph{Semantic segmentation.}
The prompt lists the requested classes with their colors. Colors are assigned at random for each sample, using the same procedure during training and inference.
Given the classes present in an image, training examples request a random subset in 70\% of cases, all present classes in 15\%, and a subset augmented with absent negative classes in the remaining 15\% (\cref{fig:color_mapping_example}). Negative classes teach the model not to hallucinate absent masks. The target paints each prompted class in its assigned color and leaves all other pixels black. At inference, each generated pixel is assigned to the nearest prompted color or to background.

\paragraph{Instance segmentation.} 
Each prompt names one category and lists 32 colors. We keep the 32 largest instances of the category, sort them by their leftmost pixel and paint the $i$-th instance in the $i$-th color (Appendix~\ref{sec:supp-instance-prompting}). The fixed order makes training targets unique for each image and prompt. Evaluation matches predicted and ground-truth instances by overlap without penalizing alternative color assignments. The readout follows Vision Banana, adding color snapping, gap filling, and GrabCut refinement (Appendix~\ref{sec:instance-readout}).

\paragraph{Referring segmentation.}
Each prompt pairs the referring expression with a color sampled independently from the named palette. The model renders the referred region in that color, which is converted to a binary mask during readout.

\subsubsection{Color Mapping for Geometry Tasks}

\paragraph{Depth.}
Following prior work~\cite{birkl2023midas,ke2024repurposingdiffusionbasedimagegenerators}, depth is encoded as affine-invariant inverse depth: valid values are normalized per image to the RGB range and replicated across channels, so nearer points are brighter (figures use a spectral colormap for readability). At inference, the prediction is mapped back to inverse depth and evaluated after per-image affine alignment.

\paragraph{Surface normals.}
Normals are converted to a common camera-space convention and each vector component is mapped linearly to one normalized RGB channel. 
At inference, the RGB prediction is mapped back to camera-space vectors and normalized to unit length. 

For both depth and surface-normal tasks, the prompts explicitly state the expected output representation. 
Further geometric encoding details are provided in \cref{sec:supp-recipe-additional-taskrgb-details}.

\subsection{Public Data Mixture}
\label{sec:recipe_data}

Vision Banana fine-tunes a proprietary base model on proprietary vision data mixed with the base model's original training data. We only use public datasets. 
We mix \emph{Perception data} for the above tasks with replay data from public text-to-image, image-editing, and identity-reconstruction datasets.
The resulting pools contain 2.1M training pairs for the zero-shot models and 3.8M for the non-zero-shot models (Appendix~\ref{sec:data-protocol}).
Appendix Tables~\ref{tab:task-sampling-weights} and~\ref{tab:datasets} give the sampling weight for each task and the datasets and their shares within each task.

\subsection{Training}
\label{sec:recipe_training}
Because all targets are RGB images, we train with the base generator's rectified-flow objective~\cite{liu2022flowstraightfastlearning}. The only task-dependent element is the validity mask described below. We encode the target as \(z_y\), sample Gaussian noise \(\epsilon\) and a uniformly distributed interpolation coefficient \(t\in(0,1]\), and form
\[
z_t = (1-t)z_y + t\epsilon.
\]
Conditioned on the clean image \(x\) and text prompt \(p\), the transformer predicts \(\hat v=v_\theta(z_t,t,p,x)\) with flow target \(\epsilon-z_y\). Let \(m_j\) indicate whether latent position \(j\) is supervised and \(C\) be the number of latent channels. We use
\begin{equation}
\mathcal{L}_{FM}
=
\mathbb{E}_{(x,p,y),t,\epsilon}
\left[
\frac{\sum_j m_j\left\|
\hat v_j - (\epsilon-z_y)_j
\right\|_2^2}{C\sum_j m_j}
\right].
\end{equation}
Geometry annotations contain undefined regions encoded by placeholder colors that a model might learn to reproduce (\cref{fig:qual-validity-mask}). Such regions account for about 8\% of target pixels in the non-zero-shot pool and are therefore excluded from the loss using a validity mask. 
Zero-shot models (2\% invalid target regions) and non-geometry tasks use no mask and standard flow-matching. For masked training, the validity mask follows the image transforms and is area-resized to the latent grid; positions with at least 50\% valid coverage contribute to the loss. We use no erosion in the final models, exclude samples with no valid latent positions, and normalize each eligible sample by its valid area. 

The full-scale models are trained for 100K steps at a global batch size of 192 on aspect-ratio buckets of about $512^2$ pixels ($512\!\times\!512$, $576\!\times\!448$, $608\!\times\!416$, and their transposes). We use AdamW~\citep{loshchilov2019decoupledweightdecayregularization} (weight decay $10^{-4}$), FSDP, and bf16 autocast with fp32 parameters, using a constant learning rate after linear warmup: $5\!\times\!10^{-6}$ (500 warmup steps) for 4B and $2\!\times\!10^{-6}$ (1{,}000 steps) for 9B. Each epoch draws 1M samples with replacement, split across tasks according to Appendix Table~\ref{tab:task-sampling-weights}. Training thus spans about 19 epochs. Final models use weight-space averaging of a few late transformer checkpoints (Appendix~\ref{sec:supp-default-optimization}). The non-zero-shot runs took about 73\,h on 16 NVIDIA GH200 GPUs (4B) and 79\,h on 32 (9B).

\subsection{Evaluation Protocol and Data}
\label{sec:recipe_evaluation}

The evaluation protocol adopts standard metrics for each task and is specified in Appendix~\ref{sec:apdx-evaluation}. All models sample with 28 steps and a guidance scale of 3.5. The datasets used for evaluation are listed in Appendix Table~\ref{tab:datasets}.

\section{Evaluating Open Vision Banana}
\label{sec:scaled}

\providecommand{\model}{Open Vision Banana}
\providecommand{\modelfull}{Open Vision Banana}
\providecommand{\modellong}{Open Vision Banana}
\providecommand{\xmark}{\textcolor{red!80!black}{\ding{55}}}
\providecommand{\cmark}{\textcolor{green!55!black}{\ding{51}}}
\providecommand{\nmark}{\textcolor{black!50}{\textemdash}}
\providecommand{\namark}{\textcolor{black!50}{n/a}}
\providecommand{\tbdmark}{\textcolor{black!55}{\textit{TBD}}}
\providecommand{\mk}[1]{\rlap{$^{#1}$}}
\providecommand{\enstop}{9pt}
\providecommand{\ensgap}{1.5pt}
\providecommand{\ensbot}{2.5pt}
\providecommand{\ens}[2]{%
  \setbox0\hbox{\rule{0pt}{\enstop}#1\mk{\diamond}}%
  \setbox2\hbox{\tiny\textcolor{black!55}{(#2)}}%
  \dimen0=\wd0 \ifdim\wd2>\dimen0 \dimen0=\wd2 \fi
  \vtop{\hbox to\dimen0{\hfil\box0\hfil}\kern\ensgap\hbox to\dimen0{\hfil\box2\hfil}\kern\ensbot}}
\providecommand{\hsub}[1]{\textcolor{black!60}{#1}}               %
\providecommand{\grp}[1]{\textcolor{black!60}{\textit{#1}}}       %

\begin{table}[t]
\centering
\caption{\textbf{Comparison of OVB with specialist and generalist models} on six perception benchmarks, zero-shot (ZS~\cmark) and non-zero-shot (\xmark): semantic segmentation (Cityscapes val, mIoU), referring segmentation (RefCOCOg val-UMD, cIoU), reasoning segmentation (ReasonSeg val, gIoU), instance segmentation (SA-Co/Gold, cgF1), depth (NYUv2, $\delta_1$ in \%) and surface normals (NYUv2, mean angular error in degrees). Geometry protocols and per-dataset results are given in Sec.~\ref{sec:final-geometry-results}.}
\label{tab:main_perception}

\scriptsize
\setlength{\tabcolsep}{2pt}
\renewcommand{\arraystretch}{1.12}
\begin{tabularx}{\columnwidth}{@{}>{\raggedright\arraybackslash}X *{7}{c}@{}}
\toprule
\textbf{Model}
  & \makecell{\textbf{ZS}\\\hsub{}}
  & \makecell{\textbf{SemSeg}\\\hsub{mIoU$\uparrow$}}
  & \makecell{\textbf{RefSeg}\\\hsub{cIoU$\uparrow$}}
  & \makecell{\textbf{ReaSeg}\\\hsub{gIoU$\uparrow$}}
  & \makecell{\textbf{InstSeg}\\\hsub{cgF1$\uparrow$}}
  & \makecell{\textbf{Depth}\\\hsub{$\delta_1\uparrow$}}
  & \makecell{\textbf{Normal}\\\hsub{err.$\downarrow$}} \\
\midrule

\multicolumn{8}{@{}l}{\grp{Open generator-based generalists}} \\[1pt]

\rowcolor{ovbzeroshot}
\model{} 9B
    & \cmark
    & \ens{73.9}{65.6}
    & 57.4
    & 67.1\mk{\star}
    & \ens{\second{37.4}}{36.0}
    & 97.4
    & 19.85 \\

\rowcolor{ovbnonzeroshot}
\model{} 9B
    & \xmark
    & \ens{\second{75.7}}{69.7}
    & \second{69.2}
    & \second{68.6}\mk{\star}
    & \ens{\best{39.6}}{38.1}
    & \best{99.0}
    & \second{13.51} \\

\rowcolor{ovbzeroshot}
\model{} 4B
    & \cmark
    & \ens{71.3}{65.1}
    & 57.0
    & 67.9\mk{\star}
    & \ens{34.0}{33.0}
    & 97.5
    & 19.87 \\

\rowcolor{ovbnonzeroshot}
\model{} 4B
    & \xmark
    & \ens{\best{76.4}}{69.2}
    & 65.6
    & \best{74.6}\mk{\star}
    & \ens{35.9}{34.8}
    & \second{98.7}
    & \best{13.21} \\

\addlinespace[2pt]
ZS Vision Learners$^\ddagger$~{\tiny\citep{liu2026opensourceimageeditingmodels}}
    & \cmark & 25.7\mk{\dagger} & \namark & \namark & \namark & \nmark & 20.5\mk{\mathsf{A}} \\
RINO-Zero$^\ddagger$~{\tiny\citep{yang2026letrgblanguagevision}}
    & \cmark & 24.3\mk{\dagger} & 40.3 & \namark & \nmark & \nmark & 20.2\mk{\mathsf{A}} \\
DICEPTION~{\tiny\citep{zhao2025diceptiongeneralistdiffusionmodel}}
    & \cmark & \nmark & \nmark & \namark & \nmark & \nmark & 18.3 \\
SenseNova-Vision~{\tiny\citep{han2026visionunifiedmultimodalgeneration}}
    & \xmark & 71.2 & \best{80.3} & 63.2 & 30.0\mk{\S} & \nmark & 14.4 \\

\midrule
\multicolumn{8}{@{}l}{\grp{Proprietary generator-based generalists}} \\[1pt]
Vision Banana~{\tiny\citep{visionbanana2026}}
    & \cmark & 69.9 & 73.8 & 79.3\mk{\star} & 47.5\mk{\star} & 94.8 & 17.8 \\

\midrule
\multicolumn{8}{@{}l}{\grp{Aggregated task specialists}} \\[1pt]
Best specialist
    & \xmark & 84.2 & 83.8 & 65.0 & 61.2 & \nmark & \nmark \\
Best specialist
    & \cmark & 65.2 & 73.4 & 77.0\mk{\star} & 24.6 & 98.0 & 16.2 \\
\bottomrule
\end{tabularx}

\vspace{2pt}
\parbox{\columnwidth}{\scriptsize
Best and second-best are marked among open generator-based generalists only. \namark{} task not supported by the model \nmark{} no comparable setup.
$^\diamond$ legend ensembling, per-pass mean in small grey: 20 passes on Cityscapes and 11 on SA-Co/Gold. OVB SA-Co/Gold scores includes an MLLM gate.
$^\star$ MLLM used to simplify the task.
$^\dagger$ image-level oracle of present classes.
$^\ddagger$ frozen Qwen-Image-Edit.
$^\S$ released checkpoint, evaluated as in \cref{sec:instance_seg_perf}.
$^{\mathsf{A}}$ output axis calibrated before normal scoring.
OVB depth uses per-image disparity-affine alignment, Vision Banana reports metric depth.
}
\end{table}

\paragraph{Setup.}
{\setlength{\fboxsep}{1pt}

We evaluate four \model{} models, obtained by fine-tuning FLUX.2 Klein 4B and
9B under two data regimes. Following the dense-prediction
literature~\citep{zou2023generalized,khangaonkar2026gen2seggenerativemodelsenable,ke2024repurposingdiffusionbasedimagegenerators,yang2024depthanythingv2},
\colorbox{ovbzeroshot}{\emph{zero-shot}} refers to datasets excluded from training, while the tasks themselves are seen during training:
the zero-shot models are never trained on any evaluation benchmark. The \colorbox{ovbnonzeroshot}{\emph{non-zero-shot}}
models additionally train on the training splits of some of the evaluation datasets (Appendix Table~\ref{tab:datasets}) and compute
the geometry loss only on valid pixels (\cref{sec:recipe_training}). Unless noted, segmentation uses the random-draw
protocol that re-samples class--color assignments for every image.

\paragraph{Legend ensembling and test-time scaling in brief.} 
For semantic and instance segmentation, scores marked $^\diamond$ use legend ensembling (details in \cref{sec:test-time-scaling}): R passes with different color legends are fused by majority vote, and the mean single-pass score is
given in parentheses. Tables report the consensus at a fixed, pre-set number of passes (20 on Cityscapes, 90 on COCO-Object, 11 on SA-Co/Gold), whereas the ``up to'' gains quoted throughout the paper are the peak of the consensus curve over all evaluated passes (stars in \cref{fig:test-time-scaling}), measured against pass 1.

\subsection{Semantic Segmentation Performance}

\begin{table}[ht]
  \centering
  \caption{\textbf{Semantic segmentation on COCO-Object.} $^\ast$COCO-trained (w/o masks).
  $^\diamond$Legend ensembling ($90$ passes).}
  \label{tab:coco-object}
  \scriptsize
  \setlength{\tabcolsep}{2pt}
  \providecommand{\rotlabel}[1]{\rotatebox{90}{\shortstack[l]{#1}}}
  \newlength{\rotheadh}\settowidth{\rotheadh}{OV-Stitcher}%
  \begin{tabularx}{\linewidth}{@{}l*{9}{>{\centering\arraybackslash}X}@{}}
    \toprule
    \raisebox{\dimexpr(\rotheadh-\height)/2\relax}{\textbf{Method}} &
    \rotlabel{DiffSegm.\\\cite{wang2025diffsegmenter}} &
    \rotlabel{iSeg\\\cite{sun2026iseg}} &
    \rotlabel{Talk2DINO$^\ast$\\\cite{barsellotti2025talk2dino}} &
    \rotlabel{RF-CLIP\\\cite{li2026rfclip}} &
    \rotlabel{FSeg-LLaVA\\\cite{zhang2026fsegllava}} &
    \rotlabel{OV-Stitcher\\\cite{moon2026ovstitcher}} &
    \rotlabel{CorrCLIP\\\cite{zhang2025corrclip}} &
    \cellcolor{ovbzeroshot}\rotlabel{\model{} 9B\\(ZS)} &
    \cellcolor{ovbnonzeroshot}\rotlabel{\model{} 9B\\(NZS)} \\
    \midrule
    \textbf{mIoU $\uparrow$} &
    37.9 & 38.4 & 41.0 & 41.8 & 42.0 & 46.5 & 49.4 &
    \cellcolor{ovbzeroshot}\begin{tabular}{@{}c@{}}\second{53.0}$^\diamond$\\(38.7)\end{tabular} &
    \cellcolor{ovbnonzeroshot}\begin{tabular}{@{}c@{}}\best{59.3}$^\diamond$\\(45.0)\end{tabular} \\
    \bottomrule
  \end{tabularx}
\end{table}

We evaluate on Cityscapes~\cite{cordts2016cityscapes} and COCO-Object~\cite{lin2014coco}. All 19 Cityscapes classes fit in one prompt. The 80 categories exceed the 512-token text window that FLUX.2 Klein models are trained with, so each pass splits them into three disjoint groups of 26–27 classes, each prompted in its own sub-pass. Each pass redraws the class partition and the class--color assignments. Legend draws and the generation seed are fixed, so every fine-tuned model sees identical prompts and noise. We compute mIoU at the native evaluation resolution against the GT maps.

\noindent\textbf{Cityscapes.}
With test-time scaling, the zero-shot (ZS) 9B model reaches 73.9 mIoU (\cref{tab:main_perception}), ahead of the strongest zero-shot specialist (SAM~3, 65.2) and of the generalists Vision Banana (69.9) and non-zero-shot (NZS) SenseNova-Vision (71.2). With target-benchmark data, the non-zero-shot 4B model reaches 76.4 mIoU, the best \model{} result. Cityscapes makes up only 0.13\% of the non-zero-shot training samples, and the zero-shot models
already reach 71.3 (4B) and 73.9 (9B) mIoU without any Cityscapes data. 
The non-zero-shot recipe improves these results by 5.1 and 1.8 points, respectively, although these differences also reflect other changes to the training mixture and geometry supervision.
Most of the accuracy the%

\noindent\textbf{COCO-Object.}
\Cref{tab:coco-object} compares against recent task specialists that, like \model{}, receive the class names and are evaluated on the 80 object classes plus background. With test-time scaling, the zero-shot 9B model surpasses all of them, including the CLIP+SAM pipelines CorrCLIP and OV-Stitcher and the LLaVA-based FSeg-LLaVA, while its mean single pass is on par with the diffusion-based DiffSegmenter and iSeg. Test-time scaling helps far more here than on Cityscapes because, with 80 classes, most errors come from
colors bound to the wrong class (\cref{sec:test-time-scaling}). Appendix~\ref{sec:coco-object-filter-analysis} studies pruning the legend to the classes that an MLLM (Qwen3-VL-8B-Instruct or Gemma4 31B) judges present
in the image.\\

\noindent\textbf{Qualitative results.}
\Cref{fig:qual-cs-semseg} show single-pass zero-shot Cityscapes and COCO-Object predictions with the 9B model. Boundaries are sharp and follow object
contours, preserving thin structures such as poles or bicycle frames.

\newlength{\ovbqualw}
\newlength{\ovbqualpairw}
\newcommand{\ovbqualimg}[1]{\includegraphics[width=\ovbqualw,keepaspectratio]{#1}}
\newcommand{\ovbqualswatch}[2]{%
  \raisebox{0.15ex}{\color[RGB]{#1}\rule{0.85ex}{0.85ex}}\hspace{0.15em}#2}
\newcommand{\ovbqualclasses}[1]{%
  \multicolumn{3}{@{}c@{}}{\scriptsize\textit{Prompted foreground:}~#1}}

\begin{figure}[t]
    \centering
    \setlength{\tabcolsep}{1pt}
    \setlength{\ovbqualw}{\dimexpr(\columnwidth-4\tabcolsep)/3\relax}
    \setlength{\ovbqualpairw}{\dimexpr(\columnwidth-8pt)/4\relax}
    \begin{tabular}{@{}ccc@{}}
        \textbf{Input} & \textbf{Ground truth} & \textbf{Prediction} \\
        \ovbqualimg{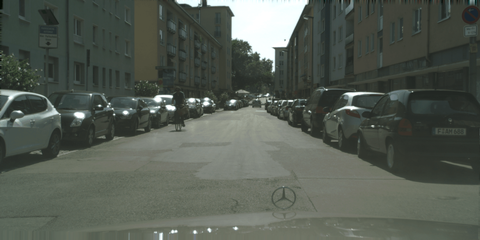} &
        \ovbqualimg{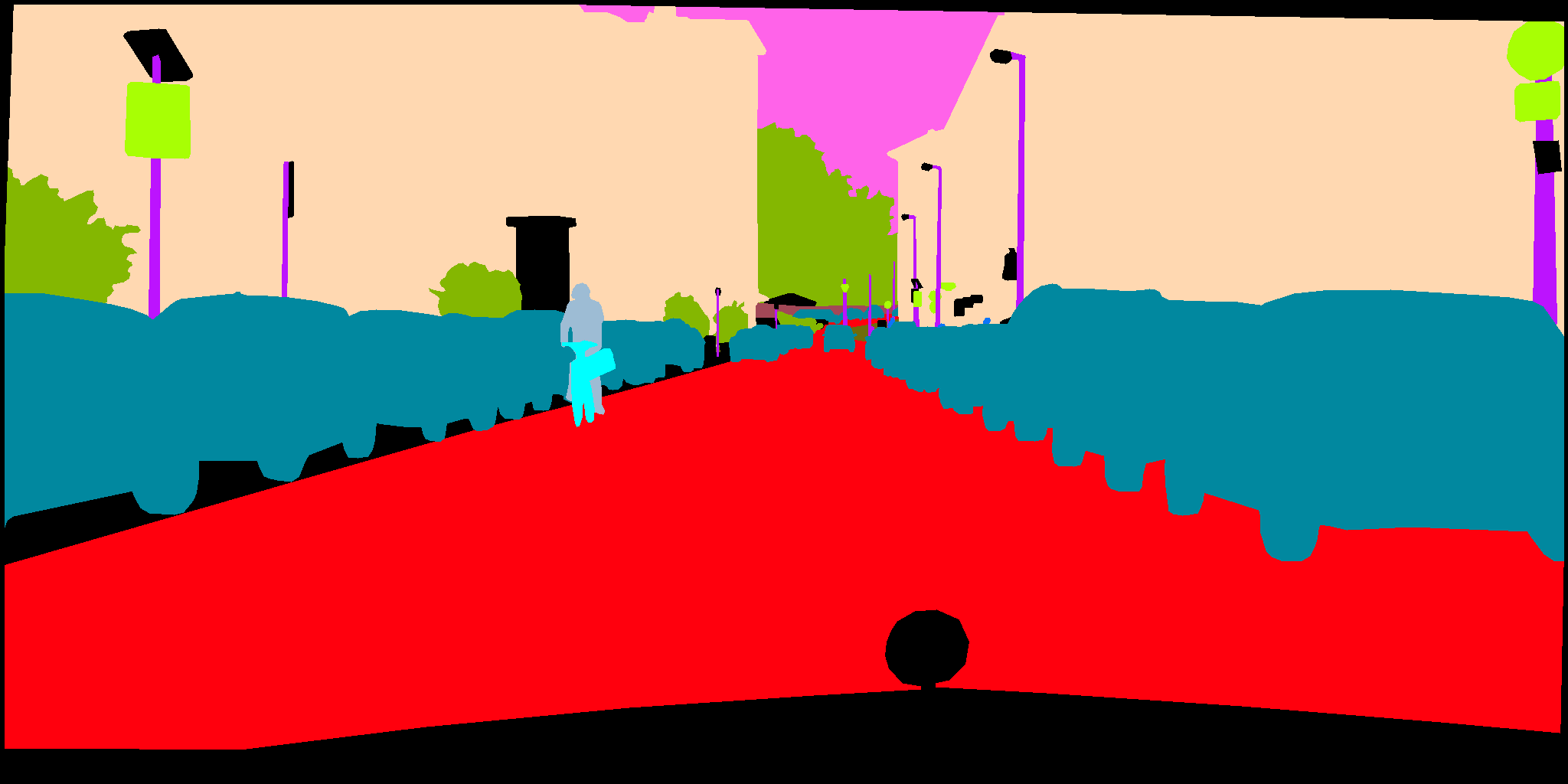} &
        \ovbqualimg{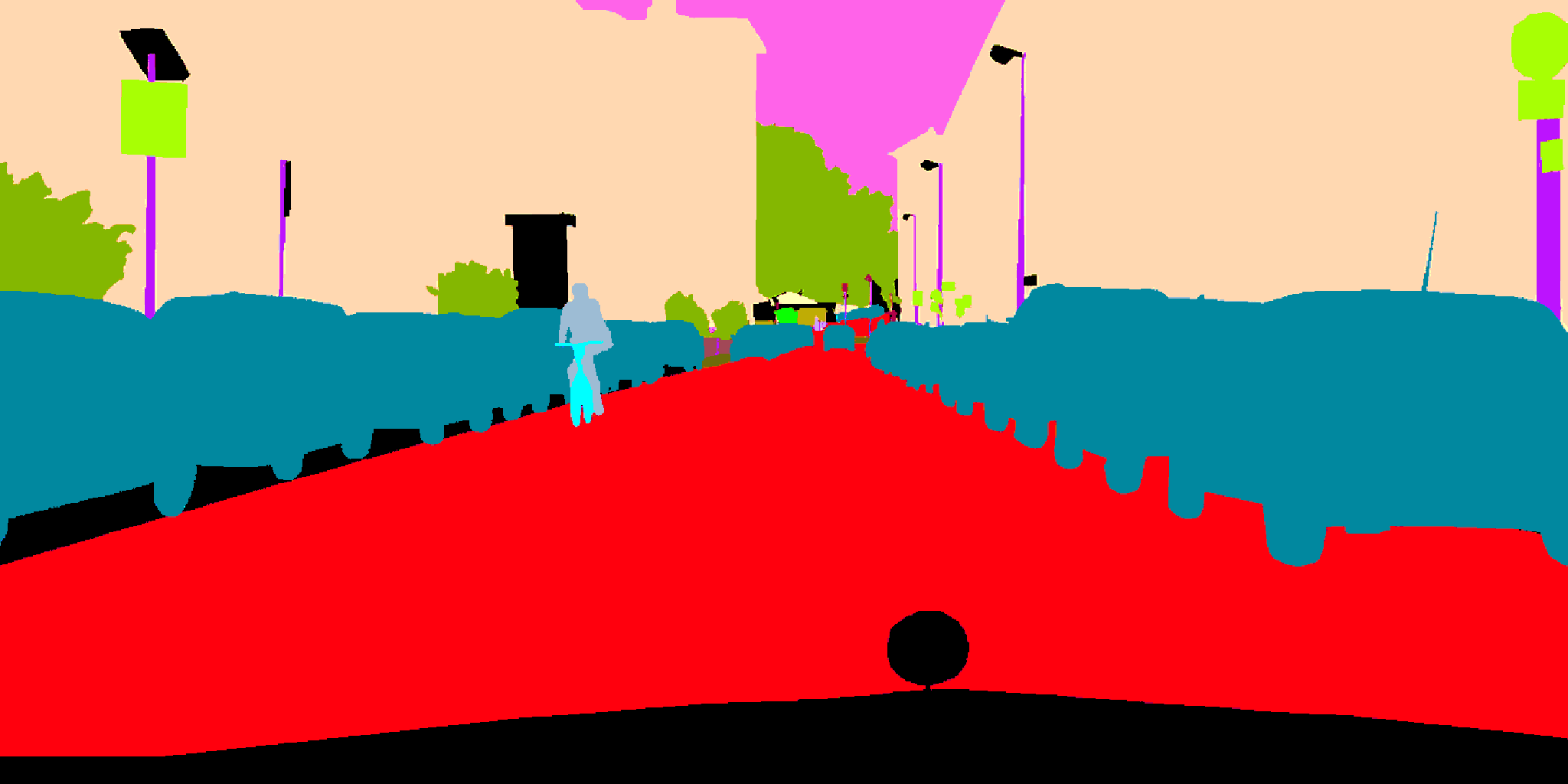} \\
        \ovbqualimg{} &
        \ovbqualimg{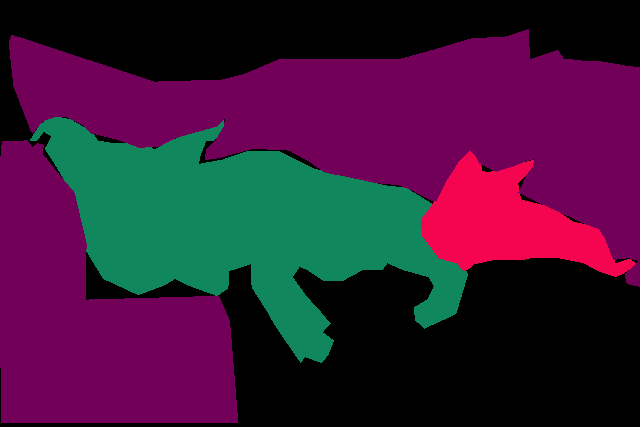} &
        \ovbqualimg{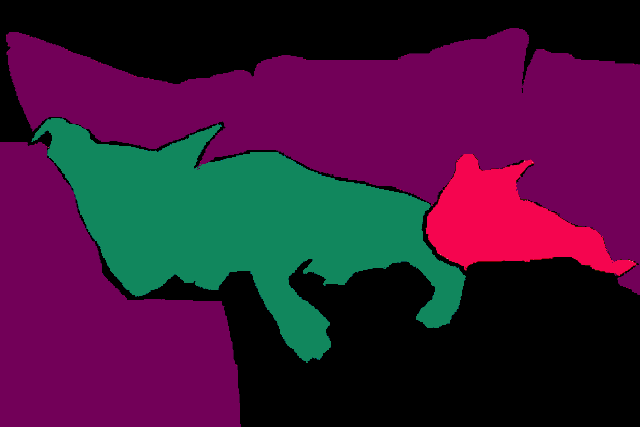} \\[-\dp\strutbox]
        \arrayrulecolor{black!25}\specialrule{0.4pt}{2pt}{2pt}\arrayrulecolor{black}
        \multicolumn{3}{@{}c@{}}{%
            \includegraphics[width=\ovbqualpairw]{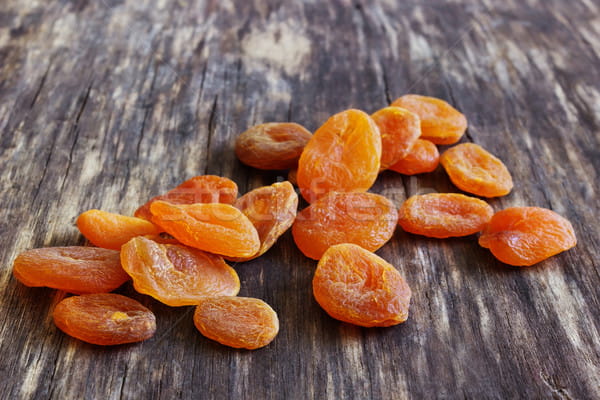}\hspace{2pt}%
            \includegraphics[width=\ovbqualpairw]{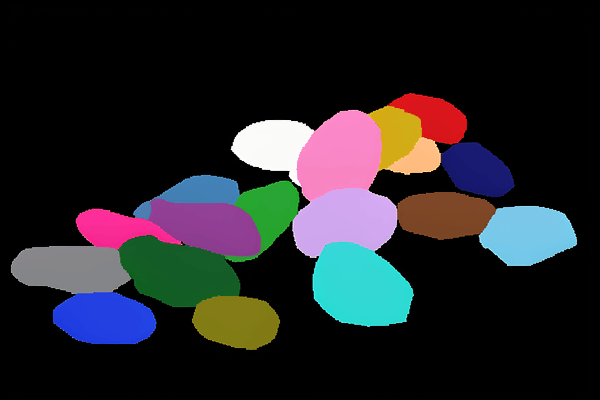}\hspace{4pt}%
            \includegraphics[width=\ovbqualpairw]{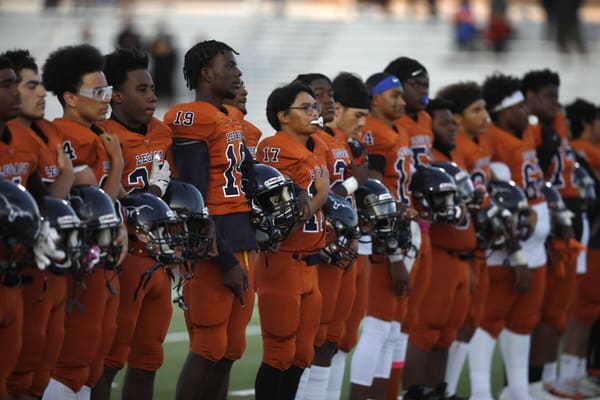}\hspace{2pt}%
            \includegraphics[width=\ovbqualpairw]{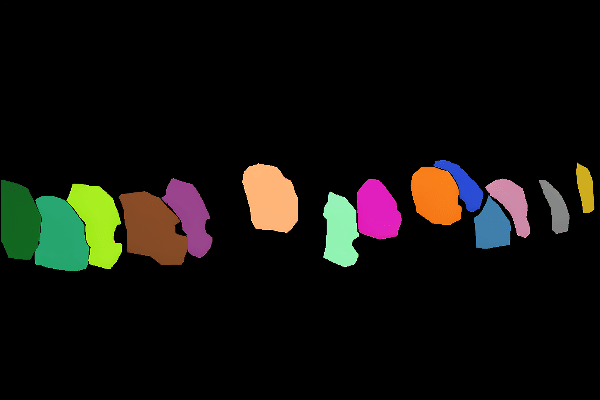}} \\
    \end{tabular}
    \vspace{-4pt}

    \caption{\textbf{Zero-shot \model{} 9B predictions:} semantic segmentation (after readout) on Cityscapes (top) and COCO-Object-80 (middle), and instance segmentation of ``dried apricots'' and ``football helmet'' on SA-Co/Gold (bottom, input--prediction pairs). Prompts are listed in Appendix~\ref{sec:prompt-templates}.}

    \label{fig:qual-cs-semseg}%
    \label{fig:qual-coco-object-80}%
    \label{fig:instanceseg-qualitative-mergedmodel}
    
\end{figure}

\subsection{Referring and Reasoning Segmentation}

\providecommand{\xmark}{\textcolor{red!80!black}{\ding{55}}}

\begin{table}[t]
\centering
\setlength{\tabcolsep}{2pt}
\renewcommand{\arraystretch}{1.05}
\def\wleft{0.51\columnwidth}%
\def\wright{0.45\columnwidth}%
\begin{minipage}[t]{\wleft}
\caption{\textbf{Reasoning segmentation on ReasonSeg val.} Gemini: Gemini 2.5 Pro.}
\label{tab:reasonseg}
\end{minipage}\hfill
\begin{minipage}[t]{\wright}
\caption{\textbf{Referring segmentation on RefCOCOg val (UMD).}}
\label{tab:refcocog}
\end{minipage}\par
\begin{minipage}[t]{\wleft}
\scriptsize
\begin{tabularx}{\linewidth}{@{}Xcc@{}}
\toprule
\textbf{Model} & \textbf{ZS} & \textbf{gIoU}$\uparrow$ \\
\midrule
X-SAM-Phi-3-3.8B {\tiny\citep{wang2026xsam}} & \xmark & 56.6 \\
LISA-13B-LLaVA-1.5 {\tiny\citep{lai2024lisa}} & \xmark & 65.0 \\
SenseNova-Vision {\tiny\citep{han2026visionunifiedmultimodalgeneration}} & \xmark & 63.2 \\
\midrule
SegZero-Qwen2.5-VL-7B {\tiny\citep{liu2025segzero}} & \cmark & 62.6 \\
RSVP-GPT-4o {\tiny\citep{lu2025rsvp}} & \cmark & 64.7 \\
SAM 3 {\tiny\citep{carion2025sam3}} + Gemini & \cmark & \second{77.0} \\
Vision Banana {\tiny\citep{visionbanana2026}} + Gemini & \cmark & \best{79.3} \\
\midrule
\rowcolor{ovbzeroshot} \model{} 4B + Gemma4 & \cmark & 67.9 \\
\rowcolor{ovbnonzeroshot} \model{} 4B + Gemma4 & \xmark & 74.6 \\
\bottomrule
\end{tabularx}
\end{minipage}\hfill
\begin{minipage}[t]{\wright}
\scriptsize
\begin{tabularx}{\linewidth}{@{}Xcc@{}}
\toprule
\textbf{Model} & \textbf{ZS} & \textbf{cIoU}$\uparrow$ \\
\midrule
HyperSeg-Phi2-2.7B {\tiny\citep{wei2024hyperseg}} & \xmark & 79.4 \\
X-SAM-Phi3-3.8B {\tiny\citep{wang2026xsam}} & \xmark & \best{83.8} \\
SenseNova-Vision {\tiny\citep{han2026visionunifiedmultimodalgeneration}} & \xmark & \second{80.3} \\
\midrule
HybridGL {\tiny\citep{liu2025hybrid}} & \cmark & 51.3 \\
LocHeads-13B
{\tiny\citep{kang2025heads}} & \cmark & 67.7 \\
SAM 3 {\tiny\citep{carion2025sam3}} + Gemini & \cmark & 73.4 \\
Vision Banana {\tiny\citep{visionbanana2026}} & \cmark & 73.8 \\
\midrule
\rowcolor{ovbzeroshot} \model{} 9B & \cmark & 57.4 \\
\rowcolor{ovbnonzeroshot} \model{} 9B & \xmark & 69.2 \\
\bottomrule
\end{tabularx}
\end{minipage}
\end{table}

\providecommand{\xmark}{\textcolor{red!80!black}{\ding{55}}}

\begin{table}[t]
\centering
\caption{\textbf{Fine-tuning vs.\ query rewriting on ReasonSeg val (gIoU).}
Raw: original query; Qwen, Gemma4, and Gemini: explicit expressions from
Qwen3-VL-32B-Thinking, Gemma4 31B, and Gemini~3.1~Pro, respectively.}

\label{tab:reasonseg-rewriter}

\setlength{\tabcolsep}{4pt}
\renewcommand{\arraystretch}{1.05}
\scriptsize
\begin{tabular}{@{}lccccc@{}}
\toprule
\textbf{Model} & \textbf{ZS} & \textbf{Raw} & \textbf{Qwen} & \textbf{Gemma4} & \textbf{Gemini} \\
\midrule
\fluxkfourbshort{} (pretrained) & \cmark & 18.5 & 24.2 & 24.1 & 23.6 \\
\midrule
\rowcolor{ovbzeroshot} \model{} 4B & \cmark & \second{41.2} & \second{68.0} & \second{67.9} & \second{69.3} \\
\rowcolor{ovbnonzeroshot} \model{} 4B & \xmark & \best{51.2} & \best{71.9} & \best{74.6} & \best{74.6} \\
\bottomrule
\end{tabular}
\end{table}

We use the referring interface of \cref{sec:recipe_task_interface} and report cIoU on RefCOCOg val (UMD) and gIoU on ReasonSeg val (Appendix~\ref{sec:apdx-evaluation}). RefCOCOg expressions are passed to \model{} unchanged. For ReasonSeg, we use Gemma4 31B to rewrite each query into an explicit referring expression. \cref{tab:reasonseg-rewriter} also reports results on the raw queries and with rewrites by Qwen3-VL-32B-Thinking~\cite{bai2025qwen3} and Gemini~3.1~Pro.

\paragraph{ReasonSeg.}
With query rewriting, the zero-shot 4B model reaches 67.9 gIoU
(\cref{tab:reasonseg}), outperforming zero-shot SegZero and RSVP-GPT-4o, as well as the target-trained LISA and SenseNova-Vision. 
SAM~3 and Vision Banana score higher, but both use Gemini~2.5~Pro as their reasoner and SAM~3 also uses it to refine its masks iteratively.
\Cref{tab:reasonseg-rewriter} disentangles the contributions of the base model, fine-tuning, and query rewriting. Even with Gemma4 rewriting, \fluxkfourbshort{} reaches only 24.1 gIoU, whereas frozen Nano Banana Pro, the base of Vision Banana, already reaches 70.2 on raw queries
\cite{di2026probegen}, close to zero-shot \model{}. Rewriting adds 26.7 gIoU
to the fine-tuned model but only 5.6 to the pretrained one,
so explicit expressions pay off only once fine-tuning has
taught the model to segment. The non-zero-shot
recipe improves the 4B model more on raw queries (41.2 to 51.2) than with
Gemma4 rewriting (67.9 to 74.6), suggesting that it partly learns to resolve implicit queries itself. Replacing Gemma4 with Gemini 3.1 Pro leaves
the non-zero-shot 4B model unchanged at 74.6 gIoU. The
choice of reasoner therefore matters much less than the perception ability of the base generator.

\paragraph{RefCOCOg.}
The zero-shot 9B model reaches 57.4 cIoU (\cref{tab:refcocog}), ahead of zero-shot HybridGL but behind LocalizationHeads-LLaVA, SAM~3 with Gemini~2.5~Pro, and Vision Banana. \model{} resolves long compositional expressions without a dedicated grounding module, whereas the stronger entries use an MLLM to interpret the expression or, in the case of Vision Banana, build on Nano Banana Pro, a proprietary generator of undisclosed but probably larger size. With the target training splits (non-zero-shot regime), the 9B model reaches 69.2 cIoU.

\noindent\textbf{Qualitative results.}
Even on raw ReasonSeg
queries, OVB often segments the region that the query implies (\cref{fig:qual-reasonseg}). On RefCOCOg, \model{} follows object outlines more tightly than the ground truth
(\cref{fig:qual-refcocog-boundaries} in the appendix).

\begin{figure}[t]
\centering
\renewcommand{\arraystretch}{1}
\setlength{\tabcolsep}{2pt}

\newlength{\rscolwtwo}
\setlength{\rscolwtwo}{\dimexpr(\linewidth-2\tabcolsep)/2\relax}

\newcommand{\rsimgtwo}[1]{\includegraphics[width=\rscolwtwo]{#1}}

\begin{tabular}{@{}
  >{\centering\arraybackslash}p{\rscolwtwo}
  >{\centering\arraybackslash}p{\rscolwtwo}
@{}}

\rsimgtwo{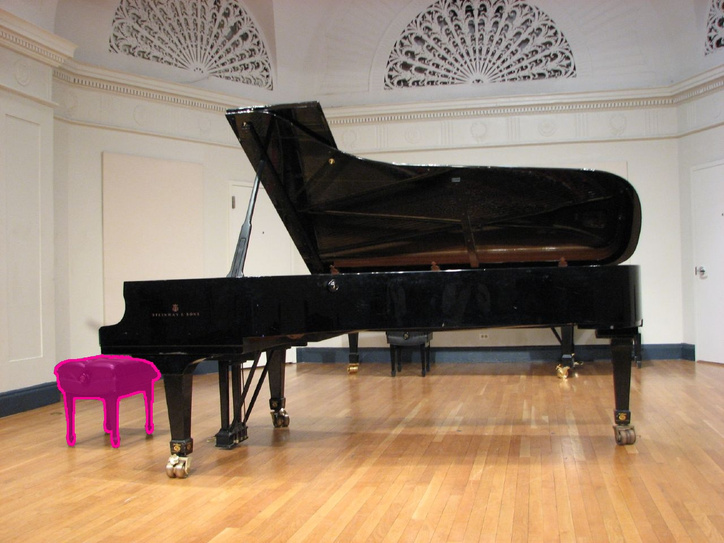} &
\rsimgtwo{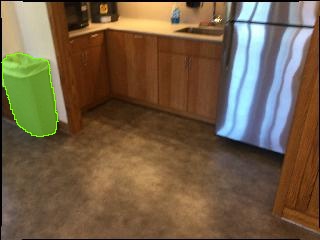} \\[1pt]

\promptbox{\parbox[t]{0.85\rscolwtwo}{\scriptsize\textcolor{promptaccent}{\textbf{Expression:}}\ \itshape the place where piano players should sit}} &
\promptbox{\parbox[t]{0.85\rscolwtwo}{\scriptsize\textcolor{promptaccent}{\textbf{Expression:}}\ \itshape After cooking, consuming food, and preparing for food, where can we throw away the rest of the food and scraps?}} \\

\end{tabular}
\caption{\textbf{Zero-shot reasoning segmentation on ReasonSeg with \model{} 9B.}
The predicted mask is overlaid on the input. Each expression is the \textbf{raw ReasonSeg query} without any rewriting and fills \texttt{\{expression\}} in the base prompt template of Sec.~\ref{sec:prompt-templates}.}
\label{fig:qual-reasonseg}
\end{figure}

\subsection{Instance Segmentation Performance}
\label{sec:instance_seg_perf}
\providecommand{\xmark}{\textcolor{red!80!black}{\ding{55}}}
\providecommand{\best}[1]{\textbf{#1}}
\providecommand{\second}[1]{\underline{#1}}

\begin{table}[ht]
\centering
\caption{\textbf{Instance segmentation on SA-Co/Gold.} $^\diamond$ denotes legend ensembling score (\cref{sec:test-time-scaling}), with the mean single-pass score over $R{=}11$ passes in parentheses. $^\dagger$ we evaluate the released checkpoint.}
\label{tab:sa-cogold}

\setlength{\tabcolsep}{4pt}
\renewcommand{\arraystretch}{1.05}
\scriptsize
\begin{tabular}{@{}p{0.39\columnwidth}cccc@{}}
\toprule
\textbf{Model} &
\textbf{ZS} &
\textbf{cgF1} $\uparrow$ &
\textbf{MCC} $\uparrow$ &
\textbf{pmF1} $\uparrow$ \\
\midrule
SenseNova-Vision$^\dagger$ {\tiny\citep{han2026visionunifiedmultimodalgeneration}} & \xmark & 30.0 & 0.72 & 39.8 \\
SAM 3 {\tiny\citep{carion2025sam3}} & \xmark & \second{54.1} & 0.82 & \second{66.1} \\
SAM 3 + Llama 3.2 (ft) & \xmark & \best{61.2} & \best{0.86} & \best{70.8} \\
\midrule
gDINO-T {\tiny\citep{liu2024grounding_dino}} & \cmark & 3.3 & 0.15 & 16.2 \\
LLMDet-L {\tiny\citep{fu2025llmdet}} & \cmark & 6.5 & 0.21 & 27.3 \\
Gemini 2.5 {\tiny\citep{gemini25}} & \cmark & 13.0 & 0.29 & 46.1 \\
APE-D {\tiny\citep{shen2024_aped}} & \cmark & 16.4 & 0.40 & 36.9 \\
DINO-X {\tiny\citep{ren2024dinox}} & \cmark & 21.3 & 0.38 & 55.2 \\
OWLv2 {\tiny\citep{minderer2023gowol}} & \cmark & 24.6 & 0.57 & 42.0 \\
VB + Gemini 3.1 Flash-Lite & \cmark & 47.5 & \second{0.84} & 56.0 \\
\midrule
\rowcolor{ovbzeroshot} \model{} 9B + Qwen3-VL-8B-Inst & \cmark & 37.4$^\diamond$(36.0) & 0.81(0.81) & 46.1$^\diamond$(44.2)  \\
\rowcolor{ovbnonzeroshot} \model{} 9B + Qwen3-VL-8B-Inst & \xmark & 39.6$^\diamond$(38.1) & 0.81(0.82) & 48.2$^\diamond$(46.5) \\	
\arrayrulecolor{black!20}
\specialrule{0.25pt}{2pt}{2pt}
\arrayrulecolor{black}
\rowcolor{ovbzeroshot} \model{} 9B + Qwen3-VL-30B-A3B & \cmark & 37.9$^\diamond$(36.4) & 0.80(0.80) & 47.3$^\diamond$(45.5)  \\
\rowcolor{ovbnonzeroshot} \model{} 9B + Qwen3-VL-30B-A3B & \xmark & 40.1$^\diamond$(38.7) & 0.80(0.80) & 49.6$^\diamond$(47.8)  \\
\bottomrule
\end{tabular}
\end{table}

We use the instance interface of \cref{sec:recipe_task_interface} and report
cgF1, IL-MCC, and pmF1 on SA-Co/Gold (Appendix~\ref{sec:apdx-evaluation}). A
Qwen3-VL-8B-Instruct presence gate decides whether the queried category is present
before \model{} paints its instances. The readout is described in
\cref{sec:instance-readout}. SenseNova-Vision evaluation uses the same gate and a panoptic prompt naming only
the query, dropping instances with other labels as in its benchmark parsing. The
model is non-zero-shot, since it trains on SA-1B images used by one SA-Co/Gold
subset.

\paragraph{Results.}
With legend ensembling over 11 passes, the zero-shot 9B model reaches 37.4 cgF1 (\cref{tab:sa-cogold}, mean single pass 36.0), ahead of OWLv2, DINO-X, and APE-D and almost 3 times the score of Gemini~2.5. The non-zero-shot
SAM 3 and Vision Banana remain ahead (54.1 and 47.5). Ensembling raises the non-zero-shot 9B model from 38.1 to 39.6 cgF1, and the Qwen3-VL-30B-A3B gate lifts the two ensembled scores to 37.9 and 40.1.
\Cref{fig:instanceseg-qualitative-mergedmodel} shows zero-shot predictions on SA-Co/Gold, where instances of the same category receive distinct colors even
when they overlap or are partially occluded.

\subsection{Test-time Scaling through Legend Ensembling}
\label{sec:test-time-scaling}

Legend ensembling varies the class–color legend in the prompt across $R$ passes over the same image while keeping the generation seed fixed. Each pass is read out with its own legend, and the resulting class maps are fused
by per-pixel majority vote (see \cref{fig:cityscapes-scaling-process}).
A black pixel is a vote for the background class on COCO-Object and no vote on Cityscapes, which has no background class. For instance segmentation, each pass permutes the 32 instance colors. Since these colors carry no class identity, voting operates on the refined binary masks of each pass (Appendix~\ref{sec:instance-readout}). Masks from different passes that overlap with $\mathrm{IoU} \geq 0.5$ are grouped into one instance. An instance is kept if it appears in at least half of the passes, and its final
mask contains the pixels covered by at least half of them. No ground-truth is used.

\begin{figure}[ht]
    \centering
    \includegraphics[width=\columnwidth]{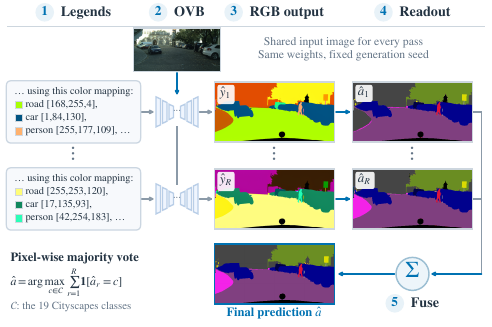}
    \caption{\textbf{Test-time scaling via legend ensembling.}
    Each pass's RGB output \(\hat{y}_r\) is read out with its own class--color legend into a class map \(\hat{a}_r\), and the \(R\) class maps are fused by per-pixel majority vote into \(\hat{a}\). Here, the readout maps are shown in the standard Cityscapes palette.}
    \label{fig:cityscapes-scaling-process}
\end{figure}

\begin{figure}[t]
    \centering
    \includegraphics[width=\columnwidth]
        {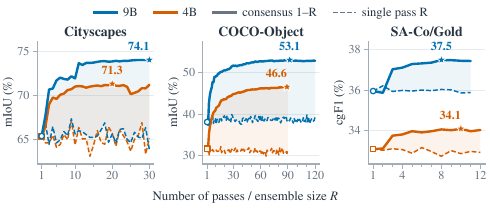}
       \caption{\textbf{Test-time scaling across segmentation benchmarks.}
    Legend-ensembling improves zero-shot \model{} 4B and 9B predictions
    as passes \(R\) increase. Stars mark peak consensus. Per
    pass, semantic legends are redrawn and instance colors are
    permuted.}
    \label{fig:test-time-scaling}
\end{figure}

\paragraph{Scaling behavior.} 
Under legend ensembling, majority-vote consensus improves as passes increase (\cref{fig:test-time-scaling}, dashed lines show single passes). On Cityscapes, the 9B model climbs quickly from 65.4 mIoU (pass 1) and saturates at about 74 mIoU after 10–15 passes, whereas COCO-Object, with 80 classes, keeps improving over many more passes. On SA-Co/Gold, each prompt names
one category, so there are fewer binding errors to correct
and the gain is smaller but ensembling still improves the 9B model from 36.0 to 37.5 cgF1. \Cref{fig:cityscapes-scaling-progression} shows how the predictions evolve as the number of passes grows.

\begin{table}[t]
\centering
\setlength{\tabcolsep}{3pt}
\renewcommand{\arraystretch}{1.0}
{\fontsize{7pt}{9pt}\selectfont
\caption{\textbf{Color binding vs.\ region segmentation quality}
of zero-shot \model{} 9B on Cityscapes.
\emph{Standard readout} decodes predictions based on the
prompted class--color legend.
\emph{Oracle relabeling} assigns the majority ground-truth class to predicted connected regions, removing
color-binding errors and isolating region errors.}
\label{tab:tts-binding}

\begin{tabular}[b]{@{}lccccc@{}}
\toprule
IoU \(\uparrow\) & \multicolumn{5}{c}{\textbf{(a) Training progression ($R{=}1$)}} \\
\midrule
& 12.5K & 22.5K & 56.3K & 74.3K & 93K \\
\midrule
Standard readout
& 51.1 & 56.7 & 61.2 & 65.5 & 68.1 \\
Oracle relabeling
& 83.6 & 84.7 & 85.9 & 85.2 & 85.9 \\
\bottomrule
\end{tabular}%
\hspace{0.5em}
\begin{tabular}[b]{@{}lcc@{}}
\toprule
\multicolumn{3}{c}{\textbf{(b) Final Model}} \\
\midrule
& $R{=}1$ & $R{=}20$ \\
\midrule
& 65.4 & \best{73.9} \\
& \best{85.9} & 85.6 \\
\bottomrule
\end{tabular}
}
\end{table}

\paragraph{Color binding and region segmentation errors.}

As a diagnostic, we relabel each predicted connected region with its majority ground-truth class (\emph{oracle} in \cref{tab:tts-binding}). The oracle score is high from early in training while the standard score keeps rising, so OVB learns to delineate regions well before it reliably paints them in the prompted colors. For the final 9B model, the oracle score far exceeds
the single-pass score on Cityscapes (85.9 vs. 65.4 mIoU). Ensembling raises the standard score to 73.9 while the oracle score stays at 85.6, suggesting that much of the gain comes from correcting class–color bindings and the regions themselves barely change.
Single-pass errors are commonly of three kinds: whole regions bound to a wrong class, unassigned pixels left black, and small speckles along object contours, where blended colors are read out as a wrong class. Legend ensembling reduces all three (see \cref{fig:cityscapes-scaling-progression}).

\paragraph{Class-wise gains and uncertainty.}
\Cref{fig:cityscapes-perclass-test-time-scaling} shows that the benefit is largest for infrequent and thin classes, which have the least stable color assignments in a single pass (e.g., ``train'' gains \(+34.4\) IoU), while large, frequent classes improve little.
Moreover, the agreement across passes provides empirical \textit{per-pixel confidence estimates}. The 6\% of Cityscapes pixels with agreement below 0.9 account for 58\% of all errors and are concentrated near object boundaries (Appendix~\ref{sec:cityscapes-vote-agreement}, \cref{fig:cityscapes-vote-agreement}).

\begin{figure}[t]
\centering
\includegraphics[width=\columnwidth]{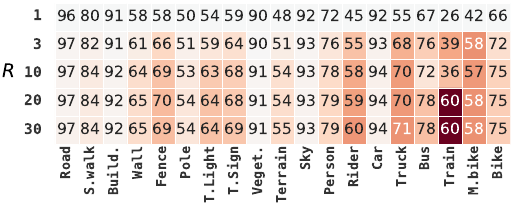}
\caption{\textbf{Per-class test-time scaling on Cityscapes.}
Per-class IoU of the zero-shot OVB 9B on the Cityscapes validation set for legend ensembles of size R. Color encodes
the gain over the \(R{=}1\) baseline, from no change (white) to the
largest gain, \(+34.4\) points on Train (dark red).}
\label{fig:cityscapes-perclass-test-time-scaling}
\end{figure}

\begin{figure}[t]
    \centering
    \includegraphics[width=\columnwidth]{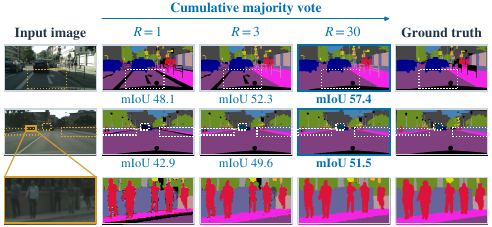}
    \caption{\textbf{Legend ensembling corrects missing and mislabeled regions.}
    \model{} 9B on two Cityscapes images with per-image mIoU. Dotted boxes mark the recovered road (top) and the recovered sidewalk and corrected bus (middle). \emph{Bottom:} zoom into the solid box. Single-pass speckles from blended colors at object outlines vanish with more passes.} 
    \label{fig:cityscapes-scaling-progression}
\end{figure}

\subsection{Performance on Geometry Tasks}
\label{sec:final-geometry}

\begin{table}[t]
\centering
\caption{\textbf{Depth and surface-normal estimation.} OVB depth uses per-image disparity-affine alignment. Full results and further baselines are in Appendix Tables~\ref{tab:final-depth-full} and~\ref{tab:final-normal-full}.}

\label{tab:final-geometry}
\scriptsize
\setlength{\tabcolsep}{4.4pt}
\renewcommand{\arraystretch}{1.02}

\begin{tabular}{@{}lcccccc@{}}
\toprule
\multicolumn{7}{@{}l}{\textbf{Aligned depth} $\delta_1\uparrow$ (\%)} \\
Method & ZS & NYU & KITTI & ETH3D & ScanNet & DIODE \\
\midrule
Marigold V2$^{\dagger}$ & \cmark & 98.0 & \best{97.4} & \best{99.2} & \best{97.9} & \best{97.1} \\
Lotus-2$^{\dagger}$ & \cmark & 97.6 & 94.1 & 98.6 & 97.2 & 95.5 \\
FE2E$^{\dagger}$ & \cmark & 97.6 & \second{96.0} & \second{98.7} & 97.1 & \second{96.4} \\
Vision Banana & \cmark & 94.8$^{\ddagger}$ & 91.5$^{\ddagger}$ & 93.5$^{\ddagger}$ & -- & -- \\
\midrule
\rowcolor{ovbzeroshot}
OVB 9B & \cmark & 97.4 & 95.3 & 98.2 & 93.4 & 95.4 \\
\rowcolor{ovbnonzeroshot}
OVB 9B & \xmark & \best{99.0} & 83.2 & 98.3 & \second{97.8} & 94.1 \\
\rowcolor{ovbzeroshot}
OVB 4B & \cmark & 97.5 & 92.5 & 98.2 & 95.8 & 94.1 \\
\rowcolor{ovbnonzeroshot}
OVB 4B & \xmark & \second{98.7} & 85.9 & 98.4 & 97.5 & 93.6 \\
\bottomrule
\end{tabular}

\vspace{3pt}

{
\setlength{\tabcolsep}{8.3pt}
\begin{tabular}{@{}lccccc@{}}
\toprule
\multicolumn{6}{@{}l}{\textbf{Surface normals} mean error $\downarrow$ (degrees)} \\
Method & ZS & NYU & ScanNet & iBims-1 & Sintel \\
\midrule
Marigold V2$^{\dagger}$ & \cmark & 16.6 & \best{14.1} & \second{15.9} & \second{28.7} \\
DSINE$^{\dagger}$ & \cmark & 16.4 & 16.2 & 17.1 & 34.9 \\
Lotus-2$^{\dagger}$ & \cmark & 16.9 & \second{14.2} & \best{15.4} & 30.3 \\
Vision Banana & \cmark & 17.8 & 15.1 & -- & -- \\
\midrule
\rowcolor{ovbzeroshot}
OVB 9B & \cmark & 19.85 & 19.07 & 16.66 & 29.88 \\
\rowcolor{ovbnonzeroshot}
OVB 9B & \xmark & \second{13.51} & 19.86 & 17.18 & \best{28.63} \\
\rowcolor{ovbzeroshot}
OVB 4B & \cmark & 19.87 & 18.81 & 16.37 & 34.18 \\
\rowcolor{ovbnonzeroshot}
OVB 4B & \xmark & \best{13.21} & 20.26 & 17.45 & 29.42 \\
\bottomrule
\end{tabular}
}
\vspace{2pt}
\parbox{\columnwidth}{\scriptsize
$^{\dagger}$Published scores from~\cite{pavlovic2026marigoldv2}. $^{\ddagger}$Vision Banana~\cite{visionbanana2026} predicts metric depth, whereas OVB is aligned depth per image. 
}
\end{table}

\begin{figure}[t]
\centering
\includegraphics[width=\columnwidth]{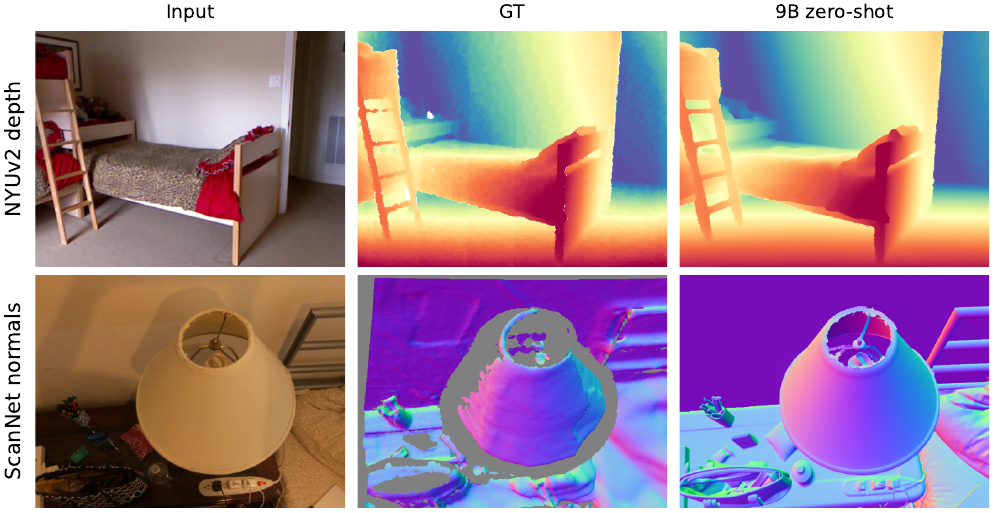}
\caption{\textbf{Selected geometry predictions of the 9B zero-shot model.} NYUv2 depth (top) uses disparity-affine alignment and a shared Spectral scale (red near, blue far). ScanNet normals (bottom) use camera-facing RGB. White and gray regions mark invalid ground truth. Paired ZS/NZS examples are shown in \cref{fig:geometry-nzs-comparison}.}
\label{fig:final-geometry-qualitative}
\end{figure}

Table~\ref{tab:final-geometry} compares the same four selected models on full depth and normal benchmarks. 
We evaluate depth on the five benchmark splits in Appendix~\ref{sec:final-geometry-results}, fitting a scale and shift to the ground-truth disparity for each image. Marigold V2 aligns in log-depth. Normals use the fixed camera-facing RGB convention, without test-image axis fitting. Published specialist scores retain their original training data and preprocessing and are reference values; Vision Banana's metric-depth scores are not directly comparable with our aligned relative-depth scores.

\paragraph{Depth.}
The non-zero-shot models improve depth on NYUv2 and ScanNet for both model sizes, and on ScanNet come close to the published specialist scores. Both zero-shot models remain stronger on KITTI. DIODE shows a mixed result: non-zero-shot training lowers 4B AbsRel but reduces its threshold accuracy, while the 9B model loses on both metrics (Appendix Table~\ref{tab:final-depth-full}). The 9B zero-shot example in \cref{fig:final-geometry-qualitative} recovers the room layout, although fine boundaries remain softer than the ground truth.

\paragraph{Surface normals.}
The non-zero-shot models lower mean angular error on NYUv2 and Sintel, but increase it on ScanNet and iBims-1 (\cref{tab:final-geometry}). For 9B, mean error falls from $19.85^\circ$ to $13.51^\circ$ on NYUv2 and from $20.53^\circ$ to $11.43^\circ$ on DIODE indoor (Appendix Table~\ref{tab:final-normal-full}).

\subsection{Retention of Generation Abilities} 
In terms of image generation, no training mixture contains data from the generation benchmarks, so all four models are zero-shot on them.
\paragraph{Image editing.} The examples in Appendix~\cref{fig:qual-imgedit} show instruction-following edits that preserve unrelated image regions. \model{} prefixes editing prompts with \texttt{[TASK: IMAGE\_EDITING]}. Adding the same prefix to the base models barely changes their ImgEdit-Bench~\cite{ye2025imgedit} scores (\cref{tab:imgedit_summary}), supporting the attribution of score differences to fine-tuning. Perception fine-tuning improves the 4B Basic average over its base model (3.66 $\to$ 3.78 and 3.82 for the zero-shot and non-zero-shot mixes), mainly through the \emph{remove} edit type (Appendix Table~\ref{tab:imgedit_results_full}), and retains the 9B Basic average (3.89 vs.\ 3.86 and 3.89). Retention is less consistent on the 50-prompt UGE suite: the 9B non-zero-shot model drops from the base model's 4.03 to 3.68, despite matching its Basic score.

\begin{table}[t]
\centering
\caption{\textbf{ImgEdit-Bench results} (mean over 3 seeds). 
Scores in parentheses use the image-editing prefix in the prompts.
Full results are in Appendix Table~\ref{tab:imgedit_results_full}. 
}
\label{tab:imgedit_summary}

\setlength{\tabcolsep}{3pt}
\renewcommand{\arraystretch}{1.05}
\footnotesize
\begin{tabular}{@{}lcc@{}}
\toprule
\textbf{Model} &
\textbf{Basic avg.} $\uparrow$ & \textbf{UGE avg.} $\uparrow$ \\
\midrule
\fluxkfourbshort{} & 3.66 (3.65) & 3.69 (3.67) \\
\rowcolor{ovbzeroshot}  \quad \model{} 4B (zero-shot mix) & \second{3.78} & \second{3.75} \\
\rowcolor{ovbnonzeroshot} \quad \model{} 4B (non-zero-shot mix) & \best{3.82} & \best{3.79} \\
\midrule
\fluxkninebshort{} & 3.89 (3.89) & \best{4.03} (3.99) \\
\rowcolor{ovbzeroshot}  \quad \model{} 9B (zero-shot mix) & \second{3.86} & \second{3.95} \\
\rowcolor{ovbnonzeroshot} \quad \model{} 9B (non-zero-shot mix) & \best{3.89} & 3.68 \\
\bottomrule
\end{tabular}
\end{table}

\paragraph{Text-to-image generation.} On GenAI-Bench (\cref{tab:genai_results}), we score all 1600 prompts with VQAScore~\citep{li2024genaibench}, using CLIP--FlanT5 as the judge over three generation seeds, with base and fine-tuned models sharing the same per-prompt noise. Our perception fine-tuning lowers VQAScore for all four models. The drop is smaller for 9B than for 4B (−0.030 vs. −0.050 with the zero-shot mixture) and slightly smaller with the non-zero-shot mixture (−0.027 and −0.049). The selected examples in \cref{fig:genai-retention} in the Appendix show successful text rendering and object counts.

\providecommand{\xmark}{\textcolor{red!80!black}{\ding{55}}}
\providecommand{\std}[1]{{\scriptsize$\pm$#1}}

\begin{table}[t]
\centering
\caption{\textbf{GenAI-Bench results} (mean over 3 seeds). $\Delta$ is the average per-prompt difference to the base model.
}
\label{tab:genai_results}

\setlength{\tabcolsep}{4pt}
\renewcommand{\arraystretch}{1.05}
\scriptsize
\begin{tabular}{@{}lcc@{}}
\toprule
\textbf{Model} & \textbf{VQAScore} $\uparrow$ & $\Delta$ \\
\midrule
\fluxkfourbshort{} & \best{0.7911} & \na \\
\rowcolor{ovbzeroshot}  \quad \model{} 4B (zero-shot mix) & 0.7412 & $-0.0499$ \\
\rowcolor{ovbnonzeroshot} \quad \model{} 4B (non-zero-shot mix) & \second{0.7418} & \best{$-0.0493$} \\
\midrule
\fluxkninebshort{} & \best{0.8123} & \na \\
\rowcolor{ovbzeroshot}  \quad \model{} 9B (zero-shot mix) & 0.7823 & $-0.0300$ \\
\rowcolor{ovbnonzeroshot} \quad \model{} 9B (non-zero-shot mix) & \second{0.7857} & \best{$-0.0266$} \\
\bottomrule
\end{tabular}
\end{table}

\subsection{Feature Quality}
We also test whether generative perception tuning improves the backbone's frozen features for downstream vision tasks. We probe the features of single-stream block 6 (4B) and 3 (9B), extracted with an empty prompt at noise level 0.7, and, as in DINOv3~\cite{simeoni2025dinov3}, keep the better of the pre- and post-normalization features. We evaluate unsupervised object discovery with TokenCut~\cite{wang2022tokencut} on COCO20k~\cite{vo2020toward}, following the standard Correct Localization (CorLoc) protocol, and geometric and semantic correspondence on ScanNet and SPair-71k~\cite{min2019spair} with Probe3D~\cite{elbanani2024probe3d} (\cref{tab:feature-quality}). Fine-tuning improves all three for both model sizes: object discovery by $+8.3$ (4B) and $+3.8$ (9B) CorLoc, geometric correspondence by up to $+13.1$ %
R@5, and semantic correspondence by up to $+2.6$ R@0.1. \model{} 9B even surpasses DINOv3 7B in object discovery (56.3 vs.\ 54.8) and semantic correspondence (61.9 vs.\ 56.1). %
Learning to output dense perception maps thus sharpens how well the backbone's features localize objects and match points across images. Appendix~\cref{fig:feature-quality-qual} shows that the fine-tuned features localize objects that the pretrained features miss or only partly cover, and their similarity maps are less noisy and better follow the object outline.

\begin{table}[t]
\centering
\caption{\textbf{Quality of frozen features} (empty prompt, noise
level 0.7) with zero-shot \model{} models. Object discovery: TokenCut CorLoc on COCO20k, best over keep fractions 0.1--0.8 for every model. Correspondence:
geometric (ScanNet R@5) and semantic (SPair-71k R@0.1). $^\dagger$Image
size 1024; our runs use 800.}
\label{tab:feature-quality}

\setlength{\tabcolsep}{4pt}
\renewcommand{\arraystretch}{1.05}
\scriptsize
\begin{tabular}{@{}lccc@{}}
\toprule
\textbf{Model} & \textbf{COCO20k} & \textbf{ScanNet} & \textbf{SPair} \\
\midrule
DINOv3 7B/16 (published)   & 55.1 & --            & 58.7$^\dagger$ \\
DINOv3 7B/16 (our run)     & \second{54.8} & \textbf{40.7} & 56.1 \\
\midrule
\fluxkfourbshort{} (pretrained)& 39.3 & 27.6          & 57.8 \\
\rowcolor{ovbzeroshot} \model{} 4B & 47.6 & \textbf{40.7} & \second{59.4} \\
\midrule
\fluxkninebshort{} (pretrained)& 52.5 & 25.6          & 59.3 \\
\rowcolor{ovbzeroshot} \model{} 9B & \best{56.3} & \second{30.5} & \best{61.9} \\
\bottomrule
\end{tabular}
\end{table}

\section{Analysis of Design Choices}
\label{sec:ablations}

In this section, we analyze critical design choices during task definition, data acquisition, and model training to derive the final recipe for generative visual perception (\cref{fig:analysis_overview}). 
Each analysis varies one factor and keeps other settings as described below.

\paragraph{Training.}
The analyses use two reduced setups, one for segmentation and one for geometry. Each variant fine-tunes \fluxkninebshort{} with the full-scale 9B setup (\cref{sec:recipe_training}) for 37.5K steps, at a global batch size of 64 (segmentation) or 8 (geometry). A segmentation run takes about
370 GPU-hours on GH200 GPUs (23\,h on 16 GPUs), and a geometry run about 80 GPU-hours on
H200 GPUs (20\,h on 4 GPUs).

\paragraph{Data.}
Training examples follow the segmentation and geometry mixtures in Appendix Table~\ref{tab:task-sampling-weights}. Each epoch draws about 250K (segmentation) or 160K (geometry) samples, so the 2.4M and 300K samples seen during training amount to 9.6 and 1.9 epochs. Segmentation sampling is without replacement; geometry sampling follows source quotas and uses replacement when a source is too small (Appendix~\ref{sec:data-protocol}).

\paragraph{Evaluation.}
\label{sec:ablation-evaluation}
Segmentation variants are evaluated on the segmentation benchmarks in Appendix Table~\ref{tab:datasets}. For the geometry ablations, we additionally report five evaluations of other tasks to check whether a gain in depth or normals comes at the cost of the model's other capabilities: Cityscapes mIoU (CS), COCO-Stuff mIoU (CO), RefCOCOg cIoU (RS), SA-Co/Gold-14K cgF1 (IS), and the ImgEdit UGE score (IE).
Cityscapes and RefCOCOg use the full validation splits, and COCO-Stuff and SA-Co/Gold use fixed subsets (\ref{sec:subset-construction}). Appendix~\ref{sec:apdx-evaluation} gives the full evaluation protocol.

\subsection{Task Definition}
\setrqprefix{S}{violet}
\label{sec:task_definition}

We first study how to express the tasks as RGB images. Additional analysis can be found in Appendix~\ref{sec:design-analysis}.

\researchquestion{color-schema}{
Which class--color mappings let the model generalize and follow the prompt?
}

\paragraph{Setup.}
To test whether a model memorizes a fixed class--color mapping (\textit{e.g.}, `car'$\rightarrow$blue) or learns to follow the colors given in the prompt, we vary the mapping at training and evaluation time, using the 19 Cityscapes classes shared by all datasets in this experiment.
\emph{Fixed colors} (\#1) always lists all 19 classes with fixed color--class correspondences, \emph{random colors} (\#2) lists all 19 classes with new colors for every sample, and \emph{random subset} (\#3) lists only a random subset of present classes, leaving unlisted regions black (\cref{fig:color_mapping_example}). Random colors are drawn from the 100 most-separated colors of our palette with a minimum pairwise color difference of $\Delta E=25$ (Euclidean distance in CIELAB, CIE76), relaxed when too few colors satisfy it, or from the 37-color named palette when a prompt uses color names (\cref{rq:color-segmentation}). Distance requirement is relaxed when fewer than required colors satisfy the filtering criteria. Each variant fine-tunes \fluxkninebshort{} for 6K steps at batch size 64 on a 40K-sample pool of 25\% semantic segmentation (BDD100K and ACDC), 25\% identity reconstruction and 50\% image editing.

\paragraph{Analysis.}
As shown in \cref{tab:color_randomization}, training with \emph{fixed colors} performs well under the same evaluation protocol (53.9 mIoU), but collapses when the colors are varied at inference (7.1), indicating rigid memorization of the mapping. Training with random colors (\#2) avoids this collapse and reaches 51.0 mIoU with unseen color assignments. In the \emph{set} protocol, the prompt lists only a subset of the classes and unrequested regions are not scored (Appendix~\ref{sec:apdx-evaluation}), so it tests whether the model follows the legend. \emph{Random subset} performs best here (66.5 mIoU) but loses 15.0 mIoU under the random-draw (\emph{rd}) protocol, where all Cityscapes classes are requested.

\begin{table}[ht]
    \scriptsize
    \centering
    \rqcaption[color-schema]{\textbf{Effects of class--color mapping}  during training (rows) and evaluation (columns).
    All values are mIoU and gray indicates the default setup used across experiments.
    }
    \label{tab:color_randomization}
    \setlength{\tabcolsep}{2pt}
    \renewcommand{\arraystretch}{1.05}
    \begin{tabular}{@{}l*{5}{c}@{}}
        \toprule
        \textbf{Considered classes \& class--color mapping} &
        \multicolumn{3}{c}{\textbf{Cityscapes}} &
        \makecell{\textbf{COCO-}\\\textbf{Stuff}} &
        \makecell{\textbf{COCO-}\\\textbf{Things}} \\
        \cmidrule(lr){2-4}
        \cmidrule(lr){5-5}
        \cmidrule(l){6-6}
        & \textbf{fixed} & \textbf{rd} & \textbf{set} & \textbf{rd} & \textbf{rd} \\
        \midrule
        (\#1) all classes \& canonical colors (\textbf{fixed}) & \textbf{53.9} & 7.1 & 21.2 & 4.4 & 3.5 \\
        (\#2) all classes \& random colors (\textbf{rd}) & \underline{31.5} & \textbf{51.0} & \underline{53.6} & \textbf{19.8} & \textbf{13.0} \\
        \rowcolor{segmentationreferencecolor} (\#3) random subset \& random colors (\textbf{set}) & 24.1 & \underline{36.0} & \textbf{66.5} & \underline{14.7} & \underline{10.5} \\
        \bottomrule
    \end{tabular}
\end{table}

\takeaway{color-schema}{
Models memorize fixed class--color mappings. Random colors let the model follow new color assignments, subset prompts teach it to paint only the listed classes, and full-set prompts keep accuracy high when every class is requested. The recipe therefore combines random colors with a mixture of subset, full-set and negative-class prompts (adding absent classes).

}

\researchquestion{color-segmentation}{
What color palette best supports unambiguous color generation for semantic segmentation?
}

\paragraph{Setup.}
We construct a 200-color palette from Matplotlib's XKCD colors using farthest-point sampling in CIELAB space and order colors from most to least separated. We compare the full palette with its first 100, 64 and 32 colors and a \emph{per-image} variant that uses the $N$ most-separated colors, where $N$ is the number of classes present in the image, or the number of listed classes if the prompt adds absent ones. 
For each training and evaluation sample, colors are drawn from the corresponding palette and randomly assigned to the requested classes. Smaller palettes offer fewer colors but larger distances between them.

\paragraph{Analysis.}
Restricting the palette to more widely separated colors generally improves segmentation over using all 200 colors, particularly on COCO-Stuff and COCO-Things. 
Shrinking the palette further does not always help: \emph{Top 32} improves Cityscapes but lowers both COCO benchmarks. \emph{Top 64} achieves the highest mean over the six settings (54.5 mIoU) and becomes the default. Appendix~\ref{sec:supp-set-segmentation} evaluates this model with a varying requested classes per image and finds that mIoU drops as more classes are requested.

\begin{table}[h]
    \footnotesize
    \centering
    \rqcaption[color-segmentation]{\textbf{Role of semantic segmentation palette.}
Rows change the palette from which colors are sampled and assigned to classes for every image.
Palette and sampling rule are shared for training and evaluation (mIoU under set and random-draw (rd)).}
    \label{tab:palette_sampling}

    \HeatRange{cs-set}{73.3}{75.8}
    \HeatRange{cs-rd}{57.9}{60.8}
    \HeatRange{stuff-set}{51.3}{60.5}
    \HeatRange{stuff-rd}{32.6}{38.7}
    \HeatRange{things-set}{43.1}{59.4}
    \HeatRange{things-rd}{18.8}{35.4}
    
    \setlength{\tabcolsep}{3pt}
    
    \renewcommand{\arraystretch}{1.08}
    
    \scriptsize
    
    \begin{tabular}{l*{6}{c}}
    
    \toprule
    
    \textbf{Palette} &
    \multicolumn{2}{c}{\textbf{Cityscapes}} &
    \multicolumn{2}{c}{\textbf{COCO-Stuff}} &
    \multicolumn{2}{c}{\textbf{COCO-Things}} \\
    
    \cmidrule(lr){1-1}
    \cmidrule(lr){2-3}
    \cmidrule(lr){4-5}
    \cmidrule(lr){6-7}
    
    & \textbf{set} & \textbf{rd} &
    \textbf{set} & \textbf{rd} &
    \textbf{set} & \textbf{rd} \\
    
    \midrule
    
    All 200 colors
    & \Heat{cs-set}{75.1}
    & \Heat{cs-rd}{57.9}
    & \Heat{stuff-set}{54.8}
    & \Heat{stuff-rd}{32.6}
    & \Heat{things-set}{43.1}
    & \Heat{things-rd}{18.8} \\

    Top 100
    & \Heat{cs-set}{73.3}
    & \Heat{cs-rd}{59.6}
    & \Heat{stuff-set}{57.1}
    & \Heat{stuff-rd}{36.6}
    & \Heat{things-set}{55.4}
    & \textbf{\Heat{things-rd}{35.4}} \\

    \cellcolor{segmentationreferencecolor} Top 64
    & \Heat{cs-set}{74.4}
    & \Heat{cs-rd}{60.3}
    & \textbf{\Heat{stuff-set}{60.5}}
    & \textbf{\Heat{stuff-rd}{38.7}}
    & \textbf{\Heat{things-set}{59.4}}
    & \Heat{things-rd}{33.6} \\

    Top 32
    & \textbf{\Heat{cs-set}{75.8}}
    & \textbf{\Heat{cs-rd}{60.8}}
    & \Heat{stuff-set}{51.3}
    & \Heat{stuff-rd}{36.5}
    & \Heat{things-set}{43.9}
    & \Heat{things-rd}{31.4} \\

    Top $N_i$ (\# classes per img.)
    & \Heat{cs-set}{75.2}
    & \Heat{cs-rd}{57.9}
    & \Heat{stuff-set}{56.1}
    & \Heat{stuff-rd}{38.2}
    & \Heat{things-set}{53.9}
    & \Heat{things-rd}{35.0} \\
    
    \bottomrule
    \end{tabular}
    \vspace{-0.5em}
\end{table}

\takeaway{color-segmentation}{
Well-separated colors outperform the full palette, with 64 colors providing the best overall trade-off.
}

\subsection{Training Data}
\setrqprefix{D}{green}

\researchquestion{D3}{
How do image-generation and image-editing data affect geometry and retained capabilities?
}

\paragraph{Setup.}
We vary the share of image-generation examples and, separately, editing and identity-reconstruction examples. Each share is set to zero, half its default, its default, or twice its default, with the remaining sampling probability redistributed across the other tasks. The defaults are 15\% and 25\%, respectively. All variants train for 50K steps.
\paragraph{Analysis.}
No setting is best for all tasks, and trends are not monotonic (\cref{tab:main-creative-data}).
Removing generation data gives the lowest normal error but worsens depth. Doubling it to 30\% raises depth AbsRel from 0.061 to 0.063, slightly increases $\delta_1$ from 96.5\% to 96.6\%, and worsens both normal errors.
Removing editing/identity data worsens both geometry tasks and sharply reduces editing retention.
The variant without generation examples also gives the highest GenAI-Bench CLIP--FlanT5 score in this sweep (0.8074 versus 0.7871 at the 15\% default; Appendix Table~\ref{tab:geometry-generation-retention}(d)). Editing and identity-reconstruction examples remain in its training mixture. Increasing the generation share to 30\% provides no consistent improvement over the 15\% default.
The 15\% generation / 25\% editing/identity control ranks first or second on seven of nine metrics in each panel.

\takeaway{D3}{
We retain 15\% generation data and 25\% editing and identity-reconstruction data to balance depth, segmentation, and editing performance.
}

\begin{table}[t]
  \centering
  \rqcaption[D3]{\textbf{Effect of image generation and editing data.}
We vary generation or the editing/identity group at fixed 50K compute. Geometry averages five complete splits per task (Sec.~\ref{sec:geometry-full-splits}); other-task evaluations follow Sec.~\ref{sec:ablation-evaluation}. Appendix Table~\ref{tab:geometry-generation-retention} adds the generation-retention scores.}
  \label{tab:main-creative-data}
   \setlength{\tabcolsep}{3pt}   %
  \renewcommand{\arraystretch}{1.08}
  \resizebox{\columnwidth}{!}{%
  \begin{tabular}{@{}lcccccccccc@{}}
    \toprule
    \multirow{2}{*}{\shortstack[l]{Training\\fraction}}
    & \multicolumn{2}{c}{Depth}
    & \multicolumn{2}{c}{Normals}
    & \multicolumn{5}{c}{Guardrails} \\
    \cmidrule(lr{0.2em}){2-3}\cmidrule(lr{0.2em}){4-5}\cmidrule(l{0.2em}){6-10}
    & AbsRel$\downarrow$ & $\delta_1\uparrow$ (\%)
    & Mean$\downarrow$ & Med.$\downarrow$
    & CS$\uparrow$ & CO$\uparrow$ & RS$\uparrow$ & IS$\uparrow$ & IE$\uparrow$ \\
    \midrule
    \multicolumn{10}{@{}l}{\textit{(a) Image generation}} \\
    0\%   & 0.064 & 96.2 & \best{22.2} & \best{12.9} & 25.9 & 15.1 & 40.1 & \second{17.6} & \best{4.1} \\
    7.5\% & 0.074 & 93.5 & 26.0 & 17.0 & \second{30.6} & \second{19.5} & 43.9 & \best{19.0} & 3.8 \\
    15\% default & \best{0.061} & \second{96.5} & \second{24.9} & \second{16.5} & 27.4 & \best{21.2} & \best{45.8} & 17.5 & \second{4.0} \\
    30\%  & \second{0.063} & \best{96.6} & 26.7 & 18.2 & \best{30.8} & 14.6 & \second{45.7} & 15.0 & 3.7 \\
    \addlinespace[3pt]
    \multicolumn{10}{@{}l}{\textit{(b) Editing and identity reconstruction}} \\
    0\%    & \second{0.062} & \second{96.3} & 31.5 & 23.9 & \second{30.8} & \second{21.2} & 40.3 & 16.6 & 2.3 \\
    12.5\% & 0.064 & 96.0 & \best{24.1} & \best{15.1} & \best{33.8} & 15.6 & \second{45.3} & \best{17.6} & \second{3.9} \\
    25\% default & \best{0.061} & \best{96.5} & \second{24.9} & 16.5 & 27.4 & \best{21.2} & \best{45.8} & \second{17.5} & \best{4.0} \\
    50\%   & 0.063 & 96.3 & 25.0 & \second{16.0} & 28.6 & 17.2 & 39.7 & 15.2 & 3.6 \\
    \bottomrule
  \end{tabular}}
\end{table}

\subsection{Architecture}
\label{sec:ablation_architecture}
\setrqprefix{A}{red}

\researchquestion{trainable}{
Which model components should be tuned?
}

\begin{table}[t]
  \centering
\rqcaption[trainable]{\textbf{Effect of adaptation scope.}
Depth averages NYUv2 and KITTI Eigen; normals average NYUv2 and DIODE indoor. ``Last $k$'' and ``All 24'' include output layers. Shading marks the geometry-ablation reference; $^\dagger$ marks the final-model scope. Evaluations follow Sec.~\ref{sec:ablation-evaluation}.}
  \label{tab:main-architecture-scope}
  \setlength{\tabcolsep}{0.75pt}
  \renewcommand{\arraystretch}{1.03}
  \resizebox{\columnwidth}{!}{%
  \begin{tabular}{@{}lcccccccccc@{}}
    \toprule
    & \multicolumn{2}{c}{Depth} & \multicolumn{2}{c}{Normals} & \multicolumn{5}{c}{Other tasks} \\
    \cmidrule(lr){2-3}\cmidrule(lr){4-5}\cmidrule(l){6-10}
    Updated scope & AbsRel$\downarrow$ & $\delta_1\uparrow$ (\%) & Mean ($^\circ$)$\downarrow$ & Med. ($^\circ$)$\downarrow$ & CS$\uparrow$ & CO$\uparrow$ & RS$\uparrow$ & IS$\uparrow$ & IE$\uparrow$ \\
    \midrule
    Last 4  & 0.120 & 85.3 & 35.2 & 30.6 & 4.5 & 3.5 & 12.5 & 3.2 & 3.60 \\
    Last 12 & 0.090 & 91.8 & 22.8 & 17.9 & 6.4 & 4.1 & 25.6 & 11.5 & \best{3.96} \\
    All 24  & 0.069 & 95.1 & 20.2 & 14.7 & 12.2 & 5.0 & 32.0 & 16.7 & 3.56 \\
    \makecell[l]{Full except\\embeddings} & \best{0.068} & \best{95.3} & \second{19.3} & \second{13.4} & \second{21.6} & \second{12.8} & \second{41.1} & \best{18.3} & 3.66 \\
    \rowcolor{segmentationreferencecolor}
    Full transformer$^\dagger$ & \second{0.068} & \second{95.1} & \best{18.9} & \best{13.0} & \best{26.0} & \best{15.1} & \best{44.4} & \second{17.5} & 3.74 \\
    \bottomrule
  \end{tabular}}
\end{table}

\paragraph{Setup.}
We progressively increase the fraction of the \fluxkninebshort{} transformer updated during fine-tuning, from only the last few single-stream blocks to all transformer parameters. \emph{Full except embeddings} updates all transformer parameters except the input and context embeddings, while \emph{full transformer} additionally adapts those embeddings.

\paragraph{Analysis.}
Tuning only the last few blocks falls behind on every task except editing (\cref{tab:main-architecture-scope}). Freezing only the embeddings gives the lowest depth error and the best instance score, whereas full-transformer adaptation gives the lowest normal error. The final models update the full transformer, which is best or second best on every task except editing.

Appendix~\ref{sec:supp-lora} studies low-rank adaptation of the 4B model as a low-compute alternative: adapters of rank 64--128 match full fine-tuning on editing, depth and normals, but trail it by about 12 mIoU on semantic segmentation.

\takeaway{trainable}{
Updating more transformer parameters substantially improves segmentation and normal estimation. Depth accuracy is highest when only the embeddings remain frozen.
}

\section{Related Work}
\label{sec:related-works}

Pretrained image generators are increasingly used for visual
perception, either by extracting their semantic and geometric features~\citep{zhao2023unleashingtexttoimagediffusionmodels,
xu2023openvocabularypanopticsegmentationtexttoimage} or by framing
vision tasks directly as image prediction~\citep{bar2022visualprompting,
wang2023images,wang2023seggptsegmentingcontext,
geng2023instructdiffusiongeneralistmodelinginterface}.
Specialized models such as Marigold and Lotus further exploit
generative priors for dense geometric prediction~\citep{
ke2024repurposingdiffusionbasedimagegenerators,
he2025lotusdiffusionbasedvisualfoundation}, but do not aim at preserving the general-purpose generation abilities.

More closely related are generalist models that unify generation and
perception. DICEPTION~\cite{zhao2025diceptiongeneralistdiffusionmodel}, Jodi~\cite{xu2025jodiunificationvisualgeneration}, OneDiffusion~\cite{le2025diffusiongenerate}, and PixWizard~\cite{lin2025pixwizardversatileimagetoimagevisual} adapt
diffusion-based generators to multiple vision tasks, while SenseNova-Vision~\cite{han2026visionunifiedmultimodalgeneration}
and MODUS~\cite{ye2026modus} build on unified multimodal architectures.
Vision Banana~\citep{visionbanana2026} demonstrates strong multi-task
perception through instruction-tuning a proprietary image generator
while retaining its generative capabilities. Other studies find that frozen generators can already solve some perception tasks out-of-the-box~\citep{yang2026letrgblanguagevision,
liu2026opensourceimageeditingmodels}.
\model{} differs from these works in that every component is public and in its controlled analysis of critical design choices. An extended discussion is provided in Appendix~\ref{sec:appdx-related-work}.

\section{Conclusion}

\modellong{} is a fully open recipe for turning pretrained image generators into unified segmentation and geometry models while largely preserving their editing abilities.
Controlled experiments favor randomized, well-separated color palettes, relative inverse depth, full transformer adaptation, and highlight a trade-off between perception accuracy and retained generation quality.
The resulting models are competitive across tasks, improve further with training-free legend ensembling, and yield stronger frozen features for object discovery and correspondence.
Their main limitation is the weaker perception of open base models relative to proprietary generators such as Nano Banana Pro.
We release data pipelines, code, and weights to promote further research.

\iftoggle{cvprfinal}{%
\clearpage

\section*{Acknowledgements}
The authors thank Imanol G. Estepa for help with experimental evaluation. The authors also gratefully acknowledge the HPC resources provided by the Erlangen National High Performance Computing Center (NHR@FAU) of the Friedrich-Alexander Universität Erlangen-Nürnberg (FAU) under the BayernKI project v115be. BayernKI funding is provided by Bavarian state authorities. This work was in part supported by the BMFTR (FKZ: 16IS24060), the DFG (SFB 1233, project number: 276693517), and the BMWE (nxtAIM project grant no. 19A23014G).  This work was in part supported by the European Union’s
Horizon Europe research and innovation programme under grant agreement No 101214398 (ELLIOT). The authors also gratefully acknowledge the Gauss Centre for Supercomputing e.V. (GCS) for providing computing time on the Supercomputer JUPITER at the Jülich Supercomputing Centre. JUPITER is supported by the EuroHPC JU and GCS through funding by the European Commission, the German Federal Ministry of Research, Technology and Space, and the Ministry of Culture and Science of the State of North Rhine-Westphalia. The authors gratefully acknowledge the computational and data resources provided by the Leibniz Supercomputing Centre (www.lrz.de). Authors further acknowledge support from the Helmholtz Association Initiative and Networking Fund (HAICORE@HZDR).
}{}

{
    \small
    \bibliographystyle{ieeenat_fullname}
    \bibliography{main}
}

\clearpage
\appendix
\maketitlesupplementary

\section*{Supplementary overview}
This supplement provides the implementation details, experimental protocols,
and further analyses needed to reproduce and interpret \model{}, together with
the exact prompts and additional quantitative and qualitative results.

\subsection*{Appendix contents}
{\footnotesize
\begin{description}[leftmargin=6.2em,labelwidth=5.5em,labelsep=0.7em,
                    itemsep=0.35em,topsep=0.2em]
  \item[Appendix~\ref{sec:appdx-related-work}] \textbf{Extended Related Work}
  \item[Appendix~\ref{sec:supp-recipe}] \textbf{RGB Task Encoding and Readout Details}
    \begin{description}[leftmargin=3.2em,labelwidth=2.6em,labelsep=0.6em,
                        itemsep=0.05em,topsep=0.1em]
      \item[\ref{sec:supp-recipe-additional-taskrgb-details}] Semantic-Seg, Depth, and Normal Encoding
      \item[\ref{sec:instance-readout}] Instance-segmentation inference and readout
    \end{description}

  \item[Appendix~\ref{sec:experimental-protocol}] \textbf{Implementation
    Details}
    \begin{description}[leftmargin=3.2em,labelwidth=2.6em,labelsep=0.6em,
                        itemsep=0.05em,topsep=0.1em]
      \item[\ref{sec:supp-default-optimization}] Training configurations
      \item[\ref{sec:data-protocol}] Training data
      \item[\ref{sec:apdx-evaluation}] Evaluation details and metrics
      \item[\ref{sec:subset-construction}] Evaluation sets 
    \end{description}

  \item[Appendix~\ref{sec:consolidated-benchmark-results}]
    \textbf{Additional Results}
    \begin{description}[leftmargin=3.2em,labelwidth=2.6em,labelsep=0.6em,
                        itemsep=0.05em,topsep=0.1em]
      \item[\ref{sec:coco-object-filter-analysis}] Legend-pruning on COCO-Object
      \item[\ref{sec:cityscapes-vote-agreement}] Confidence estimates from vote agreement
      \item[\ref{sec:final-geometry-results}] Full final-model geometry results
      \item[\ref{sec:extended-imgeditbench-results}] Extended image editing results
    \end{description}

  \item[Appendix~\ref{sec:design-analysis}] \textbf{Extended Analysis of Design Choices}
    \begin{description}[leftmargin=3.2em,labelwidth=2.6em,labelsep=0.6em,
                        itemsep=0.05em,topsep=0.1em]
      \item[\ref{sec:supp-2d-analysis}] Segmentation
        \begin{description}[leftmargin=3.6em,labelwidth=3.0em,labelsep=0.6em,
                            itemsep=0.05em,topsep=0.05em]
          \item[\ref{sec:supp-set-segmentation}] Number of requested classes
          \item[\ref{sec:supp-instance-prompting}] Instance colors and prompts
        \end{description}
      \item[\ref{sec:geometry-ablations}] Geometry
        \begin{description}[leftmargin=3.6em,labelwidth=3.0em,labelsep=0.6em,
                            itemsep=0.05em,topsep=0.05em]
          \item[\ref{sec:prompt-ablation}] Prompt and representation
          \item[\ref{sec:training-supervision-composition}] Supervision and task mixture
          \item[\ref{sec:length-ablation}] Training duration
          \item[\ref{sec:phase2-validation}] Combined choices and seed variation
          \item[\ref{sec:phase2-scope-data}] Replay and perception-data diversity
        \end{description}
      \item[\ref{sec:supp-lora}] Parameter-efficient fine-tuning
    \end{description}

  \item[Appendix~\ref{sec:prompt-definitions}] \textbf{Prompt Definitions}
    \begin{description}[leftmargin=3.2em,labelwidth=2.6em,labelsep=0.6em,
                        itemsep=0.05em,topsep=0.1em]
      \item[\ref{sec:prompt-templates}] Prompts for 2D tasks
      \item[\ref{sec:exact-geometry-prompts}] Exact geometry prompts
    \end{description}

  \item[Appendix~\ref{sec:qualitative-results}] \textbf{Additional Qualitative Results}
\end{description}
}

\renewcommand{\cmark}{\textcolor{green!55!black}{\ding{51}}}
\newcommand{\fmark}{\textcolor{blue!70!black}{\ding{100}}} %
\renewcommand{\xmark}{\textcolor{red!80!black}{\ding{55}}}
\newcommand{\pmark}{\textcolor{orange!90!black}{$\sim$}}
\newcommand{\qmark}{\textcolor{black!60}{?}}
\renewcommand{\nmark}{\textcolor{black!50}{\textemdash}}
\newcommand{\rot}[1]{\rotatebox[origin=c]{60}{\scriptsize #1}}

\begin{table*}[ht]
\centering
\scriptsize
\setlength{\tabcolsep}{2.5pt}
\renewcommand{\arraystretch}{1.12}
\caption{
\textbf{Task adaptation/evaluation coverage and openness of generator-based perception models.}
\cmark{} indicates task-specific adaptation, \fmark{} quantitative evaluation without parameter updates, \pmark{} partial/indirect/qualitative evidence, and \xmark{} that the task is not addressed.
``Gen./edit kept'' marks methods that aim to preserve generation and editing.
\public{} denotes public, \pmark{} partial, and \proprietary{} proprietary access.
``Ref. seg.'' denotes referring-expression and ``Prompt seg.'' point/box/visual-prompt segmentation.
Release status checked October 2026.
\label{tab:comparison}
}
\resizebox{\textwidth}{!}{
\setlength{\tabcolsep}{1pt}
\begin{tabular}{llcccccccccccccc}
\toprule
 & &
\multicolumn{9}{c}{\textbf{Task adaptation / evaluation}} &
\multicolumn{5}{c}{\textbf{Reproducibility}} \\
\cmidrule(lr){3-11}
\cmidrule(lr){12-16}
\textbf{Model} & \textbf{Base Model / Paradigm}
& \rot{Sem. seg.}
& \rot{Ref. seg.}
& \rot{Inst. seg.}
& \rot{Prompt seg.}
& \rot{Depth}
& \rot{Normals}
& \rot{Class.}
& \rot{Det./pose}
& \rot{Gen./edit kept}
& \rot{Arch.}
& \rot{Weights}
& \rot{Train code}
& \rot{Train data}
& \rot{Data pipeline} \\
\midrule

\rowcolor{ovbzeroshot} \modellong{}
& open I2I diffusion / flow model
& \cmark & \cmark & \cmark & \xmark
& \cmark & \cmark & \cmark & \xmark & \cmark
& \public & \public & \public & \public & \public \\

\midrule
\multicolumn{16}{l}{\textbf{\textit{Generalist / multi-task generative perception models}}} \\
Vision Banana~\cite{visionbanana2026}
& Nano Banana Pro
& \cmark & \cmark & \cmark & \xmark
& \cmark & \cmark & \xmark & \xmark & \cmark
& \proprietary & \proprietary & \proprietary & \proprietary & \proprietary \\

Z.S. Vision Learners~\cite{liu2026opensourceimageeditingmodels}
& frozen open image editors
& \fmark & \xmark & \xmark & \xmark
& \fmark & \fmark & \xmark & \xmark & \pmark
& \public & \public & n/a & n/a & \public \\

RINO-Zero~\cite{yang2026letrgblanguagevision}
& frozen open image editors / RGB interface
& \fmark & \fmark & \fmark & \xmark
& \fmark & \fmark & \xmark & \fmark & \fmark
& \public & \public & n/a & n/a & \public \\

DICEPTION~\cite{zhao2025diceptiongeneralistdiffusionmodel}
& text-to-image diffusion
& \cmark & \pmark & \cmark & \cmark
& \cmark & \cmark & \xmark & \cmark & \xmark
& \public & \public & \proprietary & \pmark & \pmark \\

SenseNova-Vision~\cite{han2026visionunifiedmultimodalgeneration}
& unified multimodal generation
& \cmark & \cmark & \pmark & \cmark
& \cmark & \cmark & \xmark & \cmark & \cmark
& \public & \public & \public & \public & \pmark \\

Jodi~\cite{xu2025jodiunificationvisualgeneration}
& joint diffusion over image/label domains
& \cmark & \xmark & \xmark & \xmark
& \cmark & \cmark & \xmark & \cmark & \cmark
& \public & \public & \public & \public & \pmark \\

OneDiffusion~\cite{le2025diffusiongenerate}
& bidirectional diffusion model
& \cmark & \xmark & \xmark & \xmark
& \cmark & \xmark & \xmark & \cmark & \cmark
& \public & \public & \proprietary & \pmark & \pmark \\

PixWizard~\cite{lin2025pixwizardversatileimagetoimagevisual}
& open-language I2I assistant
& \cmark & \cmark & \xmark & \xmark
& \cmark & \cmark & \xmark & \cmark & \cmark
& \public & \public & \public & \pmark & \pmark \\

Prompt Diffusion~\cite{wang2023incontextlearningunlockeddiffusion}
& diffusion with visual-language prompts
& \cmark & \xmark & \xmark & \xmark
& \cmark & \xmark & \xmark & \xmark & \cmark
& \public & \public & \public & \pmark & \pmark \\

InstructDiffusion~\cite{geng2023instructdiffusiongeneralistmodelinginterface}
& instruction-tuned diffusion editor
& \cmark & \cmark & \xmark & \xmark
& \xmark & \xmark & \xmark & \cmark & \cmark
& \public & \public & \public & \pmark & \pmark \\

InstructCV~\cite{gan2024instructcvinstructiontunedtexttoimagediffusion}
& instruction-tuned T2I diffusion
& \cmark & \xmark & \xmark & \xmark
& \cmark & \xmark & \cmark & \cmark & \pmark
& \public & \public & \public & \pmark & \pmark \\

\midrule
\multicolumn{16}{l}{\textbf{\textit{Task-specialized generative perception models}}} \\

Marigold (v2)~\cite{ke2024repurposingdiffusionbasedimagegenerators,
ke2025marigoldaffordableadaptationdiffusionbased, pavlovic2026marigoldv2}
& Stable Diffusion dense-prediction adapter
& \xmark & \xmark & \xmark & \xmark
& \cmark & \cmark & \xmark & \xmark & \xmark
& \public & \public & \public & \public & \public \\

Lotus~\cite{he2025lotusdiffusionbasedvisualfoundation,
he2026lotus2advancinggeometricdense}
& diffusion-based dense-prediction VFM
& \xmark & \xmark & \xmark & \xmark
& \cmark & \cmark & \xmark & \xmark & \xmark
& \public & \public & \public & \public & \public \\

FE2E~\cite{wang2026editordensegeometryestimator}
& pretrained image editor
& \xmark & \xmark & \xmark & \xmark
& \cmark & \cmark & \xmark & \xmark & \xmark
& \public & \public & \proprietary & \pmark & \pmark \\

DiGSeg~\cite{wang2026diffusionmodelgeneralistsegmentation}
& diffusion segmentation learner
& \cmark & \pmark & \xmark & \xmark
& \xmark & \xmark & \xmark & \xmark & \xmark
& \public & \proprietary & \proprietary & \pmark & \proprietary \\

Prompt2Seg~\cite{alagoz2026promptingdiffusionmodelszeroshot}
& promptable diffusion segmentation
& \xmark & \xmark & \cmark & \cmark
& \xmark & \xmark & \xmark & \xmark & \xmark
& \public & \proprietary & \proprietary & \pmark & \proprietary \\

gen2seg~\cite{khangaonkar2026gen2seggenerativemodelsenable}
& generative instance segmentation
& \xmark & \xmark & \cmark & \pmark
& \xmark & \xmark & \xmark & \xmark & \xmark
& \public & \public & \public & \public & \public \\

DiffDIS~\cite{yu2025highprecisiondichotomousimagesegmentation}
& diffusion-based dichotomous segmentation
& \pmark & \xmark & \xmark & \xmark
& \xmark & \xmark & \xmark & \xmark & \xmark
& \public & \public & \public & \public & \public \\

\bottomrule
\end{tabular}
}
\end{table*}

\section{Extended Related Work}
\label{sec:appdx-related-work}

\paragraph{Generative image models as visual priors.}
Image generators have evolved from text-to-image synthesis systems into flexible image-to-image engines through latent diffusion, transformer-based diffusion backbones, instruction tuning, and controllable conditioning~\citep{rombach2022highresolutionimagesynthesislatent,peebles2023scalablediffusionmodelstransformers,brooks2023instructpix2pixlearningfollowimage,zhang2023addingconditionalcontroltexttoimage,xiao2024omnigenunifiedimagegeneration}. 
This has motivated a growing line of work that reuses generative priors to solve perception tasks instead of training separate discriminative models. 
Early diffusion-perception methods such as VPD~\citep{zhao2023unleashingtexttoimagediffusionmodels}, ODISE~\citep{xu2023openvocabularypanopticsegmentationtexttoimage}, and TADP~\citep{kondapaneni2024textimagealignmentdiffusionbasedperception} showed that pretrained diffusion features and denoising pipelines contain spatially aligned semantic and geometric information useful for segmentation, depth, and dense prediction. These methods typically train perception heads on the extracted features or modify the architecture, so the prediction is not produced by the generator itself.

\paragraph{Vision tasks as image generation.}
A second line of work frames vision tasks directly as image generation. Early methods show that masks, depth maps, and other perception outputs can be represented as RGB images~\citep{wang2023seggptsegmentingcontext,wang2023incontextlearningunlockeddiffusion,geng2023instructdiffusiongeneralistmodelinginterface,gan2024instructcvinstructiontunedtexttoimagediffusion}.
More recent contributions exploit strong visual priors acquired by generative foundation models during large-scale pretraining for dense prediction, mostly depth and surface normals~\citep{ke2024repurposingdiffusionbasedimagegenerators,ke2025marigoldaffordableadaptationdiffusionbased,gui2024depthfmfastmonoculardepth,fu2024geowizardunleashingdiffusionpriors,ye2024stablenormalreducingdiffusionvariance,xu2024mattersrepurposingdiffusionmodels,he2025lotusdiffusionbasedvisualfoundation,he2026lotus2advancinggeometricdense,wang2026editordensegeometryestimator}.
Several concurrent works study segmentation-specific uses of generative models~\citep{wang2026diffusionmodelgeneralistsegmentation,alagoz2026promptingdiffusionmodelszeroshot,khangaonkar2026gen2seggenerativemodelsenable,yu2025highprecisiondichotomousimagesegmentation}. 
These models usually address one or a few tasks and do not try to preserve or evaluate general image generation. 

\paragraph{Unified generation and understanding.}
The idea of unifying heterogeneous vision tasks through a shared image-space prediction interface predates the recent wave of generative foundation models. Early works such as Visual Prompting via Image Inpainting~\citep{bar2022visualprompting} and Painter~\citep{wang2023images} formulated diverse computer-vision tasks as image prediction. More recently, this idea has been revisited in the context of pretrained generative models that unify generation and understanding. 
DICEPTION~\citep{zhao2025diceptiongeneralistdiffusionmodel} is one of the closest open baselines, using color-encoded perception targets in a diffusion model. Vision Banana~\citep{visionbanana2026} demonstrates that instruction-tuning Nano Banana Pro on vision-task data can turn a generator into a strong generalist perception model without sacrificing its original generation abilities. Generally, unified multimodal systems differ mainly in the generative backbone they adapt. Jodi~\citep{xu2025jodiunificationvisualgeneration}, OneDiffusion~\citep{le2025diffusiongenerate}, PixWizard~\citep{lin2025pixwizardversatileimagetoimagevisual}, and UniPercept~\citep{Zhao_2026_CVPR} build on diffusion- or flow-based image generators and model perception outputs as images, labels, or task-specific visual representations. SenseNova-Vision~\citep{han2026visionunifiedmultimodalgeneration} and MODUS~\cite{ye2026modus} instead adapt BAGEL-style~\citep{deng2025emergingpropertiesunifiedmultimodal} unified multimodal decoders, combining autoregressive text generation with a shared flow-matching visual generation pathway for producing dense RGB outputs through VAE latents. Relatedly, RINO~\citep{yang2026letrgblanguagevision} and other recent works~\citep{liu2026opensourceimageeditingmodels,di2026probegen} show that off-the-shelf image generation or editing models already possess substantial dense-prediction capabilities without perception-specific adaptation or retraining, though they typically underperform tuned approaches and rely on adjusted evaluation protocols to become competitive. \model{} adapts pretrained flow-matching generators using only public data, releases its training code and weights, and ablates the design choices that proprietary systems leave unspecified. An overview of related works is provided in \cref{tab:comparison}.

\section{RGB Task Encoding and Readout Details}
\label{sec:supp-recipe}

\subsection{Semantic-Seg, Depth, and Normal Encoding}
\label{sec:supp-recipe-additional-taskrgb-details}

\paragraph{Semantic segmentation.}
\looseness=1 \Cref{fig:color_mapping_example} shows one training image under the three class--color mappings compared in \cref{rq:color-schema}. The released recipe uses randomized colors and mixes subset, full-set (all classes present in the image), and negative-class prompts (\cref{sec:color_map_seg}).

\providecommand{\cmswatch}[1]{\textcolor[HTML]{#1}{\rule{0.6em}{0.6em}}}
\begin{figure}[t]
    \centering
    \scriptsize
    \setlength{\tabcolsep}{1pt}
    \renewcommand{\arraystretch}{0.9}
    \begin{tabular}{@{}p{0.495\columnwidth}p{0.495\columnwidth}@{}}
        \includegraphics[width=\linewidth]{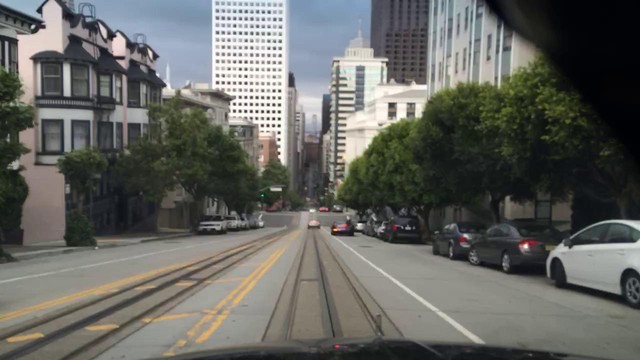} &
        \includegraphics[width=\linewidth]{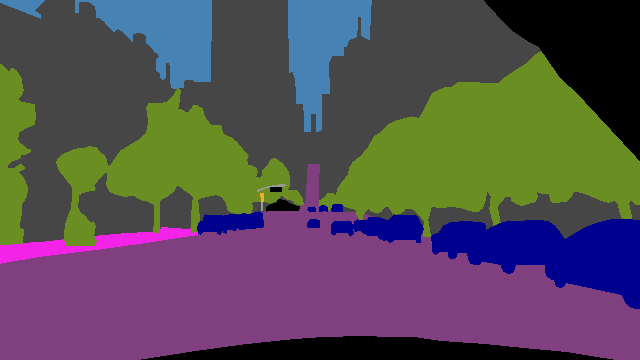} \\
        \centering\textbf{Input image} &
        \centering\arraybackslash\textbf{Fixed colors (\#1)} \\
        &
        \texttt{road}~\cmswatch{804080}, \texttt{sidewalk}~\cmswatch{F423E8}, \texttt{car}~\cmswatch{00008e}, \dots{}\newline all 19 classes, same colors across images \\[3pt]
        \includegraphics[width=\linewidth]{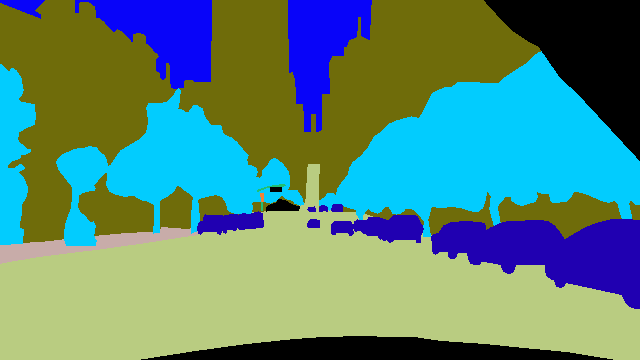} &
        \includegraphics[width=\linewidth]{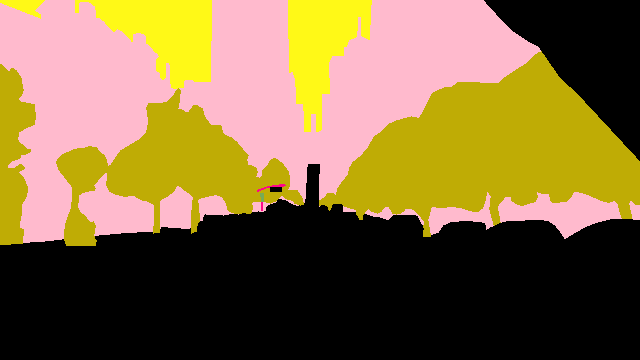} \\
        \centering\textbf{Random colors (\#2)} &
        \centering\arraybackslash\textbf{Random subset (\#3)} \\[3pt]
        \texttt{road}~\cmswatch{B9CC81}, \texttt{sidewalk}~\cmswatch{C8ACA9}, \texttt{car}~\cmswatch{2100b2}, \dots{}\newline all 19 classes, new colors per image &
        \texttt{vegetation}~\cmswatch{BFAC05}, \texttt{traffic light}~\cmswatch{77AB56}, \texttt{building}~\cmswatch{FFBACD}, \texttt{pole}~\cmswatch{FF0789}, \texttt{sky}~\cmswatch{FFF917}; others~\cmswatch{000000}
    \end{tabular}
    \caption{\textbf{Training targets under the three class--color mappings of \cref{tab:color_randomization}.}
    The same BDD100K image with its target and an excerpt of its class--color legend for each variant. \#1 and \#2 always list all 19 Cityscapes classes; \#3 lists a random subset of the classes present in the image.}
    \label{fig:color_mapping_example}
\end{figure}

\paragraph{Depth.}

\looseness=1 We first convert each source to camera-$z$ depth $z_i$ in meters.
The default target is inverse depth $d_i=1/z_i$, normalized using the
minimum and maximum valid inverse depths in each image:
$p_i=(d_i-d_{\min})/(d_{\max}-d_{\min})$.
We replicate $p_i$ over RGB, so nearer points are brighter. Constant-depth
images receive $p_i=0.5$. This encoding discards metric scale and evaluation
uses disparity-affine alignment (Sec.~\ref{sec:metrics}).

The metric-RGB ablation instead maps camera-$z$ depth in meters to
$t(z)=1-(1+z/10)^{-2}$ and then to the fixed RGB-cube path of
Vision Banana~\cite{visionbanana2026}. The decoder projects each generated
color onto the nearest path segment and inverts $t$.

\paragraph{Surface normals.}
We convert every source to unit camera-space vectors with $+x$ image-right,
$+y$ image-up, and $+z$ toward the camera.  The canonical RGB encoding is
\begin{equation}
 (R,G,B)=((1-x)/2,(1+y)/2,(1+z)/2).
 \label{eq:normal-encoding}
\end{equation}
Before training, we resolve each dataset's stored axes and signs, interpolate
the vector components and re-normalize before RGB encoding. Because the
encoding is affine, bilinearly interpolating RGB before decoding and
normalization gives the same vectors as interpolating and re-normalizing
the vector components, apart from RGB quantization. The evaluation procedure is given in
Sec.~\ref{sec:metrics}.
Horizontal augmentation flips the image and negates the physical $x$
component before re-encoding, preserving the image-right convention.

\subsection{Instance-segmentation inference and readout}
\label{sec:instance-readout}

\paragraph{Training targets.}
When the queried category is present in a training example, we keep the 32 largest instance masks of that category (the cap applies only to training targets), rank them left-to-right by their first foreground pixel with ties broken top-to-bottom, and paint the $i$-th mask in the $i$-th prompt color on a black background. Prompt and target therefore always agree on which instance gets which color. Negative examples without the queried category use the same 32-color prompt with an all-black target.

\paragraph{RGB-to-instance decoder.}
If the presence gate judges the queried category absent, no prediction is emitted. Otherwise, we first snap the generated image to the palette: at the generation resolution, every pixel takes the nearest of the 33 allowed colors (black and the 32 palette colors) in Euclidean RGB distance. Legend ensembling only permutes the order of the 32 colors across passes, so every pass snaps to the same 33 colors.
The snapped image is then resized to the ground-truth resolution with nearest-neighbor interpolation, which introduces no blended colors, and decoded.
Let \(\hat{y}(u)\in[0,255]^3\) be the snapped RGB value at pixel \(u\), and let \(c_{\mathrm{bg}}=(0,0,0)\). We first define the foreground as
\begin{equation}
    \mathcal{F}
    =
    \left\{
        u:
        \left\|\hat{y}(u)-c_{\mathrm{bg}}\right\|_2^2
        \geq \tau^2
    \right\}.
\end{equation}
We use \(\tau=58\); the same \(\tau\) also sets the color-bin width and the merge threshold below.
Pixels outside \(\mathcal{F}\) are treated as background. Each foreground RGB channel is quantized as
\(q_j(u)=\lfloor \hat{y}_j(u)/\tau\rfloor\), and 8-connected components are extracted separately within each resulting three-dimensional color bin. After snapping, \(\tau\) thus mainly decides which nearby palette colors are grouped, so fragments of one instance painted in neighboring palette colors can still be joined.

Each candidate component \(S\) then passes two filtering stages. First, we discard it when
\(\lvert S\rvert < \theta_{\mathrm{size}}HW\), where
\(\theta_{\mathrm{size}}=8\times10^{-4}\) and \(H,W\) are the image height and width. Second, using a \(3\times3\) binary erosion \(E(S)\), we discard thin boundary artifacts when
\(\lvert E(S)\rvert/\lvert S\rvert < \theta_{\mathrm{erosion}}=0.1\).
We subsequently connect two surviving components when their mean RGB values are sufficiently similar and their joint bounding box is sufficiently compact:
\begin{equation}
\begin{aligned}
    \left\|\bar c_a-\bar c_b\right\|_2^2
    &\leq \tau^2, \\
    A\!\left(B(S_a\cup S_b)\right)
    &\leq \gamma\!\left[A(B(S_a))+A(B(S_b))\right].
\end{aligned}
\end{equation}

Here, \(B(\cdot)\) denotes an axis-aligned bounding box, \(A(\cdot)\) its area, and \(\gamma=1\). The final instance masks are the connected components of this merge graph, and component colors and bounding boxes are not updated while merging. This procedure repairs nearby fragments of a painted instance, while the bounding-box constraint discourages direct merging of spatially distant same-colored fragments.

\paragraph{Gap filling and boundary refinement.}
Let \(\tilde y\) be the raw generated image, nearest-neighbor resized to the ground-truth resolution. A pixel \(u\) outside every decoded mask with \(\lVert\tilde y(u)-c_{\mathrm{bg}}\rVert_2\geq\tau\) joins the nearest decoded instance if it lies within 3 pixels of it.
Each mask \(M\) is then refined with GrabCut~\citep{rother2004grabcut}, run on the input image bilinearly resized to the ground-truth resolution and cropped to the bounding box of \(M\) enlarged by 15 pixels on each side. Let \(M^{-}\) and \(M^{+}\) be \(M\) eroded and dilated by 3 pixels with a \(3\times3\) kernel. GrabCut is initialized with \(M^{-}\) as sure foreground, \(M\setminus M^{-}\) as probable foreground, \(M^{+}\setminus M\) as probable background, and the rest of the crop as sure background. After two iterations, the result replaces \(M\) only if it keeps at least 90\% of \(\lvert M\rvert\); \(M\) is also kept when its eroded core is empty. OpenCV runs single-threaded with its random seed fixed per query and pass.
Every pass is refined independently before the vote of \cref{sec:test-time-scaling}. The instance-prompt ablation of \cref{sec:supp-instance-prompting} uses the same decoder.

The resulting instance masks are scored with the SA-Co/Gold protocol described in \cref{sec:apdx-evaluation}.

\section{Implementation Details}
\label{sec:experimental-protocol}

This section details the training configurations, training data,
and evaluation of all models. Unless stated otherwise, each ablation changes only the variable it studies.

\subsection{Training configurations}
\label{sec:supp-default-optimization}

Table~\ref{tab:training-fullscale} gives the training configuration of the
full-scale models, and Table~\ref{tab:training-defaults} those of the
segmentation and geometry ablations.

\providecommand{\tdsep}{\arrayrulecolor{white}\specialrule{1pt}{0pt}{0pt}\arrayrulecolor{black}}
\providecommand{\fsspan}[1]{\multicolumn{2}{>{\raggedright\arraybackslash}p{\dimexpr0.70\columnwidth+2\tabcolsep\relax}@{}}{#1}}
\colorlet{tdshared}{gray!12}
\colorlet{tddiff}{blue!8}
\begin{table}[!htbp]
  \centering
  \caption{Training configuration of the full-scale \model{} models. Gray rows
  are shared by both model sizes; blue rows differ. Task sampling weights are
  given in \cref{tab:task-sampling-weights}.}
  \label{tab:training-fullscale}
  \footnotesize
  \setlength{\tabcolsep}{3pt}
  \setlength{\aboverulesep}{0pt}
  \setlength{\belowrulesep}{0pt}
  \renewcommand{\arraystretch}{1.25}
  \begin{tabular}{@{}>{\raggedright\arraybackslash}p{0.22\columnwidth}>{\raggedright\arraybackslash}p{0.35\columnwidth}>{\raggedright\arraybackslash}p{0.35\columnwidth}@{}}
    \toprule
    \textbf{Setting} & \textbf{\model{} 4B} & \textbf{\model{} 9B} \\
    \midrule
    \rowcolor{tdshared} Optimizer & \fsspan{AdamW, $\beta_1=0.9$, $\beta_2=0.999$, $\epsilon=10^{-8}$, weight decay $10^{-4}$, gradient-norm clipping at 1.0} \\ \tdsep
    \rowcolor{tdshared} Schedule & \fsspan{constant learning rate after linear warm-up} \\ \tdsep
    \rowcolor{tdshared} Training length & \fsspan{100K optimizer steps (${\approx}$19 epochs)} \\ \tdsep
    \rowcolor{tdshared} Sample plan & \fsspan{1M rows per epoch, drawn with replacement, seed 42} \\ \tdsep
    \rowcolor{tdshared} Batch & \fsspan{global batch 192} \\ \tdsep
    \rowcolor{tdshared} FLUX guidance embedding & \fsspan{3.5 during training and inference} \\ \tdsep
    \rowcolor{tdshared} Parallelism & \fsspan{FSDP, BF16 mixed precision} \\ \tdsep
    \rowcolor{tdshared} Resolution & \fsspan{512 base; $512^2$, $576{\times}448$, $448{\times}576$, $608{\times}416$, $416{\times}608$} \\ \tdsep
    \rowcolor{tdshared} Loss & \fsspan{flow matching; invalid geometry targets masked for the non-zero-shot models (\cref{sec:recipe_training})} \\
    \midrule
    \rowcolor{tddiff} Base model & FLUX.2-Klein-Base 4B & FLUX.2-Klein-Base 9B \\ \tdsep
    \rowcolor{tddiff} Learning rate & $5\!\times\!10^{-6}$ & $2\!\times\!10^{-6}$ \\ \tdsep
    \rowcolor{tddiff} Warm-up & 500 steps & 1,000 steps \\     \bottomrule
  \end{tabular}
\end{table}

\providecommand{\tdsep}{\arrayrulecolor{white}\specialrule{1pt}{0pt}{0pt}\arrayrulecolor{black}}
\providecommand{\spanboth}[1]{\multicolumn{2}{>{\raggedright\arraybackslash}p{\dimexpr0.70\columnwidth+2\tabcolsep\relax}@{}}{#1}}
\colorlet{tdshared}{gray!12}
\colorlet{tddiff}{blue!8}
\begin{table}[!htbp]
  \centering
  \caption{Default controlled \fluxkninebshort{} training configurations of the
  segmentation and geometry ablations. Gray rows are shared by both ablation
  types; blue rows differ. These values are held fixed unless a study
  explicitly varies one of them; the class--color mapping study
  (\cref{rq:color-schema}) uses its own shorter schedule.}
  \label{tab:training-defaults}
  \footnotesize
  \setlength{\tabcolsep}{3pt}
  \setlength{\aboverulesep}{0pt}
  \setlength{\belowrulesep}{0pt}
  \renewcommand{\arraystretch}{1.25}
  \begin{tabular}{@{}>{\raggedright\arraybackslash}p{0.22\columnwidth}>{\raggedright\arraybackslash}p{0.37\columnwidth}>{\raggedright\arraybackslash}p{0.33\columnwidth}@{}}
    \toprule
    \textbf{Setting} & \textbf{Segmentation} & \textbf{Geometry} \\
    \midrule
    \rowcolor{tdshared} Optimizer & \spanboth{AdamW, $\beta_1=0.9$, $\beta_2=0.999$, $\epsilon=10^{-8}$, weight decay $10^{-4}$, gradient-norm clipping at 1.0} \\ \tdsep
    \rowcolor{tdshared} Learning rate & \spanboth{$2\!\times\!10^{-6}$} \\ \tdsep
    \rowcolor{tdshared} Schedule & \spanboth{constant with 1,000-step warm-up} \\ \tdsep
    \rowcolor{tdshared} Training length & \spanboth{37.5K optimizer steps} \\ \tdsep
    \rowcolor{tdshared} Evaluated checkpoint & \spanboth{final 37.5K-step checkpoint} \\ \tdsep
    \rowcolor{tdshared} FLUX guidance embedding & \spanboth{3.5 during training and inference} \\ \tdsep
    \rowcolor{tdshared} Parallelism & \spanboth{FSDP, BF16 mixed precision} \\ \tdsep
    \rowcolor{tdshared} Resolution & \spanboth{512 base; $512^2$, $576{\times}448$, $448{\times}576$, $608{\times}416$, $416{\times}608$} \\ \tdsep
    \rowcolor{tdshared} Auxiliary losses & \spanboth{none} \\
    \midrule
    \rowcolor{tddiff} Adaptation scope & Full transformer & Full except input/context embeddings \\ \tdsep
    \rowcolor{tddiff} Sample plan & 250K rows per epoch, seed 42 & 160K rows per epoch, seed 42 \\ \tdsep
    \rowcolor{tddiff} Batch & 2/GPU, accumulation 1--2; global batch 64 & 1/GPU, accumulation 2; global batch 8 \\ \tdsep
    \rowcolor{tddiff} Gradient checkpointing & disabled & enabled \\
    \bottomrule
  \end{tabular}
\end{table}

\paragraph{Model averaging.}
Each final model averages the transformer weights of two to five checkpoints from the last 20K steps of training, selected by their overall performance on the evaluation benchmarks. The exact steps and weights are listed with the released models. Table~\ref{tab:checkpoint-averaging} compares the averaged zero-shot 9B model with its final 100K checkpoint under matched settings: gains are modest, within one point on every task.
\begin{table}[t]
  \centering
  \caption{\textbf{Checkpoint averaging vs.\ the final checkpoint} for zero-shot \model{} 9B model. Matched single-pass settings: SA-Co/Gold uses the 14K-query subset. Depth and normals average NYUv2, KITTI, ETH3D, ScanNet and DIODE, and NYUv2, ScanNet, iBims-1, Sintel and DIODE indoor, respectively. $\Delta$ is averaged minus 100K.}
  \label{tab:checkpoint-averaging}
  \setlength{\tabcolsep}{4pt}
  \renewcommand{\arraystretch}{1.05}
  \small
  \begin{tabular}{@{}llccr@{}}
    \toprule
    Benchmark & Metric & Averaged & 100K & $\Delta$ \\
    \midrule
    Cityscapes & mIoU$\uparrow$ & 65.4 & 64.4 & $+1.0$ \\
    RefCOCOg & cIoU$\uparrow$ & 57.4 & 56.8 & $+0.6$ \\
    ReasonSeg & gIoU$\uparrow$ & 66.9 & 66.3 & $+0.6$ \\
    SA-Co/Gold & cgF1$\uparrow$ & 36.2 & 35.6 & $+0.6$ \\
    ImgEdit Basic & score$\uparrow$ & 3.86 & 3.88 & $-0.02$ \\
    ImgEdit UGE & score$\uparrow$ & 3.95 & 3.89 & $+0.06$ \\
    Depth & AbsRel (\%)$\downarrow$ & 6.52 & 6.51 & $+0.01$ \\
    Depth & $\delta_1$ (\%)$\uparrow$ & 95.9 & 95.9 & $0.0$ \\
    Normals & Mean ($^\circ$)$\downarrow$ & 21.20 & 21.17 & $+0.03$ \\
    Normals & $<\!11.25^\circ$ (\%)$\uparrow$ & 40.8 & 40.4 & $+0.4$ \\
    \bottomrule
  \end{tabular}
\end{table}

\subsection{Training data}
\label{sec:data-protocol}

Table~\ref{tab:datasets} lists the full-scale training pool and the evaluation
datasets, and Table~\ref{tab:task-sampling-weights} gives the per-task sampling
weights of the full-scale (OVB) and ablation mixtures.

\begin{table}[t]
\centering
\footnotesize

\caption{
\textbf{Per-task sampling weights} of the segmentation- and geometry-ablations and the full-scale OVB training mixtures.
}

\label{tab:task-sampling-weights}

\setlength{\tabcolsep}{3pt}
\renewcommand{\arraystretch}{1.15}

\begin{tabular}{
    p{0.43\columnwidth}
    p{0.15\columnwidth}
    p{0.15\columnwidth}
    >{\columncolor{ovbzeroshot}} p{0.15\columnwidth}}
\hline
\textbf{Task type} &
\textbf{Segm.} &
\textbf{Geom.} &
\textbf{OVB} \\
\hline

Semantic segmentation &
24.8\% &
14\% &
14.3\% \\

Referring segmentation &
18.8\% &
8\% &
13.2\% \\

Instance segmentation &
24.8\% &
8\% &
14.3\% \\
\hline

Depth estimation &
0\% &
20\% &
10.0\% \\

Surface normals &
0\% &
10\% &
20.0\% \\
\hline

Image editing &
11.9\% &
15\% &
11.3\% \\

Text-to-image &
11.9\% &
15\% &
11.3\% \\

Identity reconstruction &
7.9\% &
10\% &
5.5\% \\
\hline
\multicolumn{4}{p{\dimexpr0.88\columnwidth+6\tabcolsep\relax}}{%
  \vspace{-4pt}%
  \scriptsize Columns may not sum to 100\% due to rounding.
} \\
\end{tabular}
\end{table}

\paragraph{Full-scale training data.}
The zero-shot \model{} models use the non-underlined sources in
Table~\ref{tab:datasets}, excluding target-benchmark training splits
and evaluation images. The five geometry sources
(Synscapes, Replica, Matterport3D, Hypersim and Taskonomy) cover rendered
street scenes and synthetic, reconstructed and scanned interiors. We convert
all sources to the conventions of
Sec.~\ref{sec:supp-recipe-additional-taskrgb-details}.

Non-zero-shot models also use the underlined sources and their training
splits, which grows the pool from 2.1M to 3.8M training pairs. For VKITTI2, training uses non-clone renderings of the five scenes
used for normal evaluation. Training and test images therefore show the
same scenes under different rendering conditions.

For geometry sources added to the non-zero-shot pool, we identify contiguous
border rows and columns with at least half their pixels invalid. We crop
these from the image, target and mask, unless the crop would remove more
than 25\% at any edge or retain less than 60\% of the image area.
After this step, we discard examples with more than 25\% invalid area.
Remaining invalid regions inside the image are excluded from the loss by
the validity mask.

The non-zero-shot Replica pool contains 3,341 upright views rendered with mesh vertex colors; the zero-shot models use the earlier pool. For the added ScanNet source, we fit local planes to depth to obtain normals and orient them toward the camera. Released manifests record source versions, counts, image IDs and crop coordinates.

\paragraph{Ablation training data.}
The ablations draw from smaller subsets of the zero-shot pool and follow the
same exclusions.
\begin{itemize}
  \item \emph{Segmentation ablations} use no geometry data. They use the non-underlined 2D sources but leave out Hypersim (semantic and instance segmentation) as well as RailSem19 and Trans10K-v2 (semantic segmentation).
    Referring segmentation uses 150K samples from Visual Genome (without COCO images) and Mapillary Vistas.
  \item \emph{Geometry ablations} restrict the 2D tasks to ACDC, BDD100K,
    ADE20K, and PASCAL VOC for semantic segmentation and to ADE20K, Mapillary
    Vistas, and PASCAL VOC for instance segmentation. The depth and normal weights are each
split equally among the five non-underlined geometry sources. The depth-
    and normal-prompt variants of the prompt study assign the full 30\%
    geometry share to a single geometry task. Editing replay uses the first 50K valid Pico-Banana pairs after shuffling, identity replay uses 10K of their source images as both input and target, and text-to-image replay uses 500K PD12M examples. All three selections use seed 42. Each geometry-ablation epoch draws 160K examples across all tasks according to the task weights.
\end{itemize}

\definecolor{srcseg}{HTML}{6C4AB6}
\definecolor{srcgeo}{HTML}{1F8A70}
\definecolor{srcgen}{HTML}{D9730D}
\providecommand{\sh}[2]{\nobreak\,{\setlength{\fboxsep}{0.7pt}\scriptsize\colorbox{ovbzeroshot}{#1}\kern0.8pt\colorbox{ovbnonzeroshot}{#2}}}
\providecommand{\srcsep}{,\hskip 0.65em plus 0.25em minus 0.1em\ignorespaces}
\providecommand{\shnz}[1]{\nobreak\,{\setlength{\fboxsep}{0.7pt}\scriptsize\colorbox{ovbnonzeroshot}{#1}}}
\begin{table*}[t]
\centering
\footnotesize

\caption{ \textbf{Per-task training and evaluation sources.} \underline{Underlined} sources are used only by the non-zero-shot models. For those that are also evaluation benchmarks (Cityscapes, RefCOCOg, ReasonSeg, NYUv2, DIODE, ScanNet v2), only the training split is used. For Virtual KITTI 2, we train on non-clone renderings of the evaluated scenes. Training and evaluation images are therefore always disjoint. Each training source is followed by its share of the task's samples (\%, rounded) for the \colorbox{ovbzeroshot}{zero-shot} and \colorbox{ovbnonzeroshot}{non-zero-shot} models (Sec.~\ref{sec:data-protocol}). Evaluation split and subset sizes are given in Table~\ref{tab:geometry-dataset-protocols} and Sec.~\ref{sec:subset-construction}. }

\label{tab:datasets}

\setlength{\tabcolsep}{3pt}
\renewcommand{\arraystretch}{1.15}
\resizebox{\textwidth}{!}{
\begin{tabular}{
    >{\raggedright\arraybackslash\hyphenpenalty=10000}p{0.15\linewidth}
    >{\raggedright\arraybackslash\hyphenpenalty=10000}p{0.45\linewidth}
    >{\raggedright\arraybackslash}p{0.36\linewidth}}
\hline
\textbf{Task type} &
\textbf{Training sources} &
\textbf{Evaluation sources} \\
\hline

\cellcolor{srcseg!14}\textcolor{srcseg}{\textbf{Semantic segmentation}}\newline{\scriptsize \textcolor{srcseg}{14.3\% of samples}} &
ACDC~\cite{sakaridis2021acdc}\sh{1.0}{0.6}\srcsep
ADE20K~\cite{zhou2017ade20k}\sh{11.0}{6.9}\srcsep
BDD100K~\cite{yu2020bdd100k}\sh{4.0}{2.5}\srcsep
FoodSeg 103~\cite{wu2021largescalebenchmarkfoodimage}\sh{3.5}{2.2}\srcsep
GOOSE~\cite{goose-dataset}\sh{4.4}{2.7}\srcsep
Hypersim~\cite{roberts2021hypersim}\sh{7.3}{4.6}\srcsep
Mapillary Vistas~\cite{neuhold2017mapillary}\sh{9.9}{6.2}\srcsep
MUSES~\cite{brodermann2024muses}\sh{0.9}{0.5}\srcsep
PASCAL VOC~\cite{everingham2010pascalvoc}\sh{1.4}{0.9}\srcsep
PASCAL-Context~\cite{mottaghi_cvpr14}\sh{5.0}{3.1}\srcsep
RailSem19~\cite{zendel2019railsem19}\sh{4.2}{2.6}\srcsep
Trans10K-v2~\cite{ijcai2021p165}\sh{3.0}{1.9}\srcsep
VIPSeg~\cite{miao2022vipseg}\sh{37.3}{23.3}\srcsep
Waymo PVPS~\cite{mei2022waymo}\sh{7.0}{4.4}\srcsep
\underline{Cityscapes}~\cite{cordts2016cityscapes}\shnz{0.9}\srcsep
\underline{COCONut-XL} (COCO images)~\cite{deng2024coconut,lin2014coco}\shnz{15.0}\srcsep
\underline{COCONut-XL} (Objects365 images)~\cite{deng2024coconut,shao2019objects365}\shnz{5.0}\srcsep
\underline{IDD}~\cite{varma2019idd}\shnz{2.5}\srcsep
\underline{nuImages}~\cite{caesar2020nuscenes}\shnz{5.1}\srcsep
\underline{Synscapes}~\cite{wrenninge2018synscapes}\shnz{7.8}\srcsep
\underline{WildDash2}~\cite{zendel2022wilddash2}\shnz{1.3}
&
Cityscapes~\cite{cordts2016cityscapes}, \newline
COCO thing/stuff~\cite{lin2014coco,caesar2018cocostuff}
(w/ 12/15 super-categories and 500 image subsets for ablations), \newline
COCO-Object (80 Categories)~\cite{lin2014coco} \\
\hline

\cellcolor{srcseg!14}\textcolor{srcseg}{\textbf{Referring segmentation}}\newline{\scriptsize \textcolor{srcseg}{13.2\% of samples}} &
Visual Genome~\cite{krishna2017visualgenome}
(w/o COCO-derived samples)\sh{87.6}{20.5} and
Mapillary Vistas~\cite{neuhold2017mapillary}\sh{12.4}{9.5} (annotations from~\citep{ruthardt2026steervit})\srcsep
\underline{RefCOCO}~\cite{kazemzadeh2014referitgame}\shnz{8.6}\srcsep
\underline{RefCOCO+}~\cite{yu2016modeling}\shnz{8.6}\srcsep
\underline{RefCOCOg}~\cite{mao2016refcocog}\shnz{5.6}\srcsep
\underline{gRefCOCO}~\cite{liu2023gres}\shnz{11.7}\srcsep
\underline{PhraseCut}~\cite{wu2020phrasecut}\shnz{11.5}\srcsep
\underline{LVIS}~\cite{gupta2019lvis} (annotations~\citep{ruthardt2026steervit})\shnz{23.1}\srcsep
\underline{ReasonSeg}~\cite{lai2024lisa}\shnz{1.0}
&
RefCOCOg UMD~\cite{mao2016refcocog,nagaraja2016context},
ReasonSeg~\cite{lai2024lisa} \\
\hline

\cellcolor{srcseg!14}\textcolor{srcseg}{\textbf{Instance segmentation}}\newline{\scriptsize \textcolor{srcseg}{14.3\% of samples}} &
ACDC~\cite{sakaridis2021acdc}\sh{1.3}{0.5}\srcsep
ADE20K~\cite{zhou2017ade20k}\sh{12.4}{5.1}\srcsep
GOOSE~\cite{goose-dataset}\sh{5.6}{2.3}\srcsep
Hypersim~\cite{roberts2021hypersim}\sh{8.7}{3.6}\srcsep
Mapillary Vistas~\cite{neuhold2017mapillary}\sh{13.1}{5.4}\srcsep
MUSES~\cite{brodermann2024muses}\sh{1.1}{0.4}\srcsep
OVIS~\cite{qi2022ovis}\sh{5.6}{2.3}\srcsep
PASCAL VOC~\cite{everingham2010pascalvoc}\sh{1.9}{0.8}\srcsep
VIPSeg~\cite{miao2022vipseg}\sh{42.8}{17.6}\srcsep
Waymo PVPS~\cite{mei2022waymo}\sh{7.5}{3.1}\srcsep
\underline{COCONut-XL} (COCO images)~\cite{deng2024coconut,lin2014coco}\shnz{24.0}\srcsep
\underline{COCONut-XL} (Objects365 images)~\cite{deng2024coconut,shao2019objects365}\shnz{6.0}\srcsep
\underline{IDD}~\cite{varma2019idd}\shnz{2.1}\srcsep
\underline{nuImages}~\cite{caesar2020nuscenes}\shnz{4.1}\srcsep
\underline{PACO-LVIS}~\cite{ramanathan2023paco}\shnz{14.7}\srcsep
\underline{Synscapes}~\cite{wrenninge2018synscapes}\shnz{6.7}\srcsep
\underline{WildDash2}~\cite{zendel2022wilddash2}\shnz{1.1}
&
SA-Co/Gold~\cite{carion2025sam3} (Qwen3-VL-8B presence gate) \newline
Ablations -- 14K-query seed-0 subset (7 subsets $\times$ 2K queries) \newline
Main result comparisons -- all 168K queries \\
\hline

\cellcolor{srcgeo!14}\textcolor{srcgeo}{\textbf{Depth estimation}}\newline{\scriptsize \textcolor{srcgeo}{10.0\% of samples}} &
Synscapes~\cite{wrenninge2018synscapes}\sh{20.0}{11.1}\srcsep
Hypersim~\cite{roberts2021hypersim}\sh{20.0}{11.1}\srcsep
Taskonomy~\cite{zamir2018taskonomy}\sh{20.0}{11.1}\srcsep
Matterport3D~\cite{chang2017matterport}\sh{20.0}{11.1}\srcsep
Replica~\cite{straub2019replica}\sh{20.0}{11.1}\srcsep
\underline{NYUv2}~\cite{silberman2012nyu}\shnz{11.1}\srcsep
\underline{DIODE}~\cite{vasiljevic2019diode}\shnz{11.1}\srcsep
\underline{ScanNet v2}~\cite{dai2017scannet}\shnz{11.1}\srcsep
\underline{Virtual KITTI 2}~\cite{cabon2020virtualkitti2}\shnz{11.1}
&
Final benchmarks: NYUv2~\cite{silberman2012nyu},
DIODE~\cite{vasiljevic2019diode},
KITTI Eigen~\cite{geiger2012kitti},
ETH3D~\cite{schops2017eth3d},
ScanNet v2~\cite{dai2017scannet}.\newline
Development diagnostics (Sec.~\ref{sec:subset-construction}):
iBims-1~\cite{koch2020ibims},
nuScenes~\cite{caesar2020nuscenes},
Virtual KITTI 2~\cite{cabon2020virtualkitti2} \\
\hline

\cellcolor{srcgeo!14}\textcolor{srcgeo}{\textbf{Surface normals}}\newline{\scriptsize \textcolor{srcgeo}{20.0\% of samples}} &
Synscapes~\cite{wrenninge2018synscapes}\sh{20.0}{11.1}\srcsep
Hypersim~\cite{roberts2021hypersim}\sh{20.0}{11.1}\srcsep
Taskonomy~\cite{zamir2018taskonomy}\sh{20.0}{11.1}\srcsep
Matterport3D~\cite{chang2017matterport}\sh{20.0}{11.1}\srcsep
Replica~\cite{straub2019replica}\sh{20.0}{11.1}\srcsep
\underline{NYUv2}~\cite{silberman2012nyu}\shnz{11.1}\srcsep
\underline{DIODE}~\cite{vasiljevic2019diode}\shnz{11.1}\srcsep
\underline{ScanNet v2}~\cite{dai2017scannet}\shnz{11.1}\srcsep
\underline{Virtual KITTI 2}~\cite{cabon2020virtualkitti2}\shnz{11.1}
&
Final benchmarks: NYUv2~\cite{silberman2012nyu},
DIODE~\cite{vasiljevic2019diode} (official normals),
iBims-1~\cite{koch2020ibims} and Sintel (reference normal targets),
Virtual KITTI 2~\cite{cabon2020virtualkitti2},
ScanNet v2~\cite{dai2017scannet}.\newline
Development diagnostics: depth-derived normals for
ETH3D~\citep{schops2017eth3d}, KITTI~\cite{geiger2012kitti},
nuScenes~\cite{caesar2020nuscenes}, iBims-1, Virtual KITTI 2 and ScanNet v2 \\
\hline

\cellcolor{srcgen!14}\textcolor{srcgen}{\textbf{Image editing}}\newline{\scriptsize \textcolor{srcgen}{11.3\% of samples}} &
Pico-Banana-400K~\cite{qian2025picobanana}\sh{100}{100}
&
ImgEdit-Bench~\cite{ye2025imgedit} \\
\hline

\cellcolor{srcgen!14}\textcolor{srcgen}{\textbf{Text-to-image}}\newline{\scriptsize \textcolor{srcgen}{11.3\% of samples}} &
PD12M~\cite{meyer2024pd12m}\sh{100}{100}
&
GenAI-Bench~\cite{li2024genaibench} \\
\hline

\cellcolor{srcgen!14}\textcolor{srcgen}{\textbf{Identity reconstruction}}\newline{\scriptsize \textcolor{srcgen}{5.5\% of samples}} &
Pico-Banana-400K~\cite{qian2025picobanana}\sh{100}{100}
&
-- \\
\hline

\end{tabular}
}
\end{table*}

\subsection{Evaluation details and metrics}
\label{sec:apdx-evaluation}

Our evaluation protocol follows Vision Banana~\citep{visionbanana2026} where applicable and extends it to a broader benchmark suite. The exact datasets used for evaluation are listed in Table~\ref{tab:datasets}.

\paragraph{Generation.}
We compare each final model with its base generator using matched inference and judging settings, with 28 steps and guidance 3.5. GenAI-Bench~\citep{li2024genaibench} uses VQAScore with the CLIP--FlanT5 [XXL] judge, for the final models and the geometry ablations alike (IG-C, Table~\ref{tab:geometry-generation-retention}); ImgEdit-Bench~\citep{ye2025imgedit} uses ImgEdit-Judge. Both results are averaged over three generation seeds. Editing uses a fixed assignment to 16 logical shards because the noise generator advances between examples within each shard.

\paragraph{Semantic segmentation.} \textit{Mean intersection-over-union} (mIoU), computed as the arithmetic mean of the per-class IoUs is used for evaluation. We consider three evaluation protocols. \textit{Fixed} uses the complete dataset taxonomy with its canonical class--color mapping. \textit{Random-draw} (\textit{rd}) retains the complete taxonomy but samples a new class--color mapping for each image, thereby measuring generalization across colors. \textit{Set} requests a random subset of the classes present in each image and assigns them random colors. Its size is drawn uniformly between 1 and one fewer than the number of present classes, or is 1 if only one class is present. Only the requested classes are scored and pixels of other classes are ignored, so this protocol measures how accurately the model paints the requested classes, without penalizing painting in unrequested regions. Unless otherwise stated, we use random-draw evaluation throughout the paper. 

\paragraph{Referring segmentation.} Following LISA~\cite{lai2024lisa}, we report cumulative IoU (cIoU), the summed foreground intersection divided by the summed foreground union over all queries, and gIoU, the mean per-query foreground IoU. cIoU weights queries by their union area, whereas gIoU gives every query equal weight.

\paragraph{Instance segmentation.} \looseness=1 We use the official SA-Co/Gold evaluator of SAM 3, as Vision Banana does. It retains only queries that are instance-exhaustive in all three oracle annotations. Each decoded instance (\cref{sec:instance-readout}) is exported as a binary COCO RLE mask with confidence 1.0, so the evaluator's 0.5 score threshold discards nothing. Queries that the VLM presence gate judges absent receive no prediction, Vision Banana filters queries the same way with Gemini 3.1 Flash-Lite. For each query, we score the same predictions against all three annotations and keep the result with the highest local F1, averaged over the IoU thresholds from 0.50 to 0.95 in steps of 0.05. An empty prediction for an empty target counts as perfect. Evaluation uses all ground-truth instances, without the training cap of 32. Positive micro-F1 (pmF1) measures instance matching and mask localization on positive queries. Image-level Matthews correlation coefficient (IL-MCC) measures category-presence accuracy. Within each subset, $\mathrm{cgF1}=\mathrm{pmF1}\times\mathrm{IL\text{-}MCC}$, so hallucinated predictions on negative queries are penalized. We average each metric over the seven subsets, so mean cgF1 need not equal mean pmF1 times mean IL-MCC.

\paragraph{Geometry.}
Table~\ref{tab:geometry-dataset-protocols} lists the final evaluation splits.
Unless a caption specifies a smaller set, depth ablations use all five depth benchmarks and normal ablations all normal benchmarks except Sintel. Depth benchmarks use the image lists and reference depths distributed with Marigold~\citep{ke2024repurposingdiffusionbasedimagegenerators}.

\paragraph{Crops and valid-pixel masks.}
\label{sec:final-geometry-protocol}
\looseness=1 For both inverse-gray and metric-RGB predictions, depth evaluation retains valid ground-truth pixels within the dataset-specific ranges in Table~\ref{tab:geometry-dataset-protocols}. NYUv2 uses the standard crop, rows 45--470 and columns 41--600. KITTI applies a bottom-centered $1216\!\times\!352$ crop to the input and ground-truth depth, then scores rows 117--320 and columns 43--1171 within that cropped grid. All row and column ranges are zero-indexed and inclusive. ETH3D is scored at $2048\!\times\!1360$, with nearest-neighbor resizing of depth, masks and predictions. DIODE combines its official mask with an edge filter: the summed absolute
horizontal and vertical Sobel responses on metric depth must be at most $0.3$. The same mask is used for alignment and scoring.
After disparity-affine alignment, inverse-gray predictions are clipped to the
dataset depth bounds; ETH3D uses the largest valid ground-truth depth in each
image as its upper bound.

\looseness=1 Normal evaluation uses the benchmark targets and masks. ScanNet and iBims-1 use float normal maps distributed with Marigold; the iBims-1 maps are DSINE plane fits to laser-scanner depth~\citep{DSINE}. Target vectors must be finite and nonzero. The axis convention is fixed before evaluation and never fitted on test images.

\begin{table}[t]
  \centering
  \caption{Final geometry evaluation splits. Counts are images; dashes indicate
  tasks not evaluated on that dataset. Depth ranges are in meters and apply to valid ground-truth pixels. DIODE depth includes indoor and outdoor scenes; its normal split is indoor only.
  Masks and crops are specified in Sec.~\ref{sec:final-geometry-protocol}.}
  \label{tab:geometry-dataset-protocols}
  \scriptsize
  \setlength{\tabcolsep}{4pt}
  \begin{tabular}{@{}lrrc@{}}
    \toprule
    Dataset & Depth & Normals & Depth range \\
    \midrule
    NYUv2~\cite{silberman2012nyu} & 654 & 654 & $(0.1,10)$ \\
    KITTI Eigen~\cite{geiger2012kitti} & 652 & -- & $(0.1,80)$ \\
    ETH3D~\cite{schops2017eth3d} & 454 & -- & $(0.01,\infty)$ \\
    ScanNet~\cite{dai2017scannet} & 800 & 300 & $(0.01,10)$ \\
    DIODE~\cite{vasiljevic2019diode} & 771 & 325 & $(0.6,350)$ \\
    iBims-1~\cite{koch2020ibims} & -- & 100 & -- \\
    Sintel & -- & 627 & -- \\
    Virtual KITTI 2~\cite{cabon2020virtualkitti2} & -- & 1,000 & -- \\
    \bottomrule
  \end{tabular}
\end{table}

\paragraph{Depth metrics and alignment.}
\label{sec:metrics}
For ground-truth camera-$z$ depth $z_i$ and predicted depth $\hat z_i$ on the
valid set $\Omega$, with $N=|\Omega|$, we report
\begin{align}
 \mathrm{AbsRel} &= \frac{1}{N}\sum_{i\in\Omega}
       \frac{|\hat z_i-z_i|}{z_i},\\
 \delta_1 &= \frac{1}{N}\sum_{i\in\Omega}
 \mathbf 1\!\left[\max\!\left(\frac{\hat z_i}{z_i},
 \frac{z_i}{\hat z_i}\right)<1.25\right].
\end{align}
\looseness=1 Tables report $\delta_1$ in percent. AbsRel is a fraction in the ablations and a percentage in Table~\ref{tab:final-depth-full}.
We obtain inverse-gray predictions $p_i$ by converting generated RGB to
8-bit grayscale (Pillow mode \texttt{L}) and dividing by 255.
We fit a scale and shift by least squares,
$ap_i+b\approx1/z_i$, on each image's valid pixels. We clamp the fitted
disparity below at $10^{-6}$, invert it and clip the resulting depth before
scoring on the same pixels. The metric-RGB ablation instead decodes RGB to
depth and fits scale and shift in log-depth with RANSAC, following
Marigold~V2~\citep{pavlovic2026marigoldv2}. It clips aligned depth to
$[10^{-3},\max_{i\in\Omega}z_i]$\,m.
Both protocols measure relative depth after ground-truth-assisted alignment,
not recovery of absolute distance.

\raggedbottom %
\paragraph{Surface normals.}
We bilinearly resize generated RGB to the ground-truth grid, decode it using
Eq.~\ref{eq:normal-encoding}, and normalize prediction and target vectors.
For each valid pixel, the angular error in degrees is
\begin{equation}
 e_i=\frac{180}{\pi}\arccos\!\left(
 \operatorname{clip}(\hat{\mathbf n}_i^\top\mathbf n_i,-1,1)\right).
\end{equation}
\flushbottom%
Final-model tables report mean angular error and the percentage below
$11.25^\circ$. Ablation tables report mean and median error.
Predictions with a non-finite decoded norm or a norm below $0.01$ receive
a $90^\circ$ error and remain included in all reported metrics.

\paragraph{Aggregation and ranking.}
\looseness=1 We average per-image metrics within each dataset. Thus, the reported normal median is the mean of the per-image medians, not a median over pooled pixels. Geometry averages weight datasets equally and use the same dataset intersection for all variants in a comparison. Bold and underline mark the best and second-best values, ranked before rounding unless a caption specifies otherwise. 

\subsection{Evaluation sets}
\label{sec:subset-construction}

\looseness=1 Final models are evaluated on the complete evaluation sets listed in Table~\ref{tab:datasets}. Because the design-choice analyses only compare variants with each other, they use cheaper subsets of COCO-Stuff, COCO-Things and SA-Co/Gold. COCO-Stuff and COCO-Things use the first 500 validation images in order of image ID, scored on their 15 and 12 super-categories. For SA-Co/Gold, we sample 2,000 instance-exhaustive queries from each of its seven subsets (seed 0), giving 14K queries, and reuse the same presence-gate decisions in all comparisons. Geometry studies use the splits of Appendix~\ref{sec:geometry-ablations}. Studies that deviate from these defaults state their evaluation sets in their setup or caption.

\paragraph{Development examples.}
The validity-mask examples in Figure~\ref{fig:qual-validity-mask} come from an earlier development suite with different depth bounds and splits. That suite used ETH3D two-view training scenes, already-cropped KITTI depth-selection frames, nuScenes \texttt{CAM\_FRONT} validation images with projected-LiDAR validity, and Virtual KITTI 2 \texttt{clone} views from camera 0. Its settings do not apply to the final evaluations in Table~\ref{tab:geometry-dataset-protocols}. Figure~\ref{fig:qual-klein-depth-finetuning} instead uses the final zero-shot 9B model and benchmark targets.

\section{Additional Results}
\label{sec:consolidated-benchmark-results}

This section shows extended evaluations with the final \model{} models.

\subsection{Legend pruning on COCO-Object}
\label{sec:coco-object-filter-analysis}

\looseness=1 Because accuracy drops as more classes are requested (\cref{sec:supp-set-segmentation}), legend pruning first asks an MLLM which of the 80 categories appear in the image and prompts \model{} only with those. 
Pruning raises the mean single-pass mIoU of the zero-shot 9B model from 38.7 to 62.2 with Qwen3-VL-8B-Instruct and to 63.5 with Gemma4 31B, and that of the non-zero-shot model from 45.0 to 65.2 and 66.1 (\cref{tab:coco-object,tab:coco-object-filter}). Both zero-shot and non-zero-shot \model{} 9B models score above the listed specialists.

\begin{table}[ht]
  \centering
  \caption{\textbf{COCO-Object with legend pruning.}
  $^\diamond$20-pass majority vote; mean single-pass mIoU in parentheses.}
  \label{tab:coco-object-filter}
  \scriptsize
  \setlength{\tabcolsep}{4pt}
  \begin{tabular}{@{}>{\raggedright\arraybackslash}p{0.72\linewidth}c@{}}
    \toprule
    \textbf{Method} & \textbf{mIoU $\uparrow$} \\
    \midrule
    \multicolumn{2}{@{}l}{\textit{Task-specialist models}} \\
    CorrCLIP~\cite{zhang2025corrclip}     & 49.4 \\
    OV-Stitcher~\cite{moon2026ovstitcher} & 46.5 \\
    FSeg-LLaVA~\cite{zhang2026fsegllava}  & 42.0 \\
    \midrule
    \multicolumn{2}{@{}l}{\textit{Ours with legend pruning (Qwen3-VL-8B-Instruct)}} \\
    \rowcolor{ovbzeroshot}
    \model{} 9B (zero-shot)     & 63.6$^\diamond$ (62.2) \\
    \rowcolor{ovbnonzeroshot}
    \model{} 9B (non-zero-shot) & \second{66.4}$^\diamond$ (65.2) \\
    \midrule
    \multicolumn{2}{@{}l}{\textit{Ours with legend pruning (Gemma4 31B)}} \\
    \rowcolor{ovbzeroshot}
    \model{} 9B (zero-shot)     & 64.5$^\diamond$ (63.5) \\
    \rowcolor{ovbnonzeroshot}
    \model{} 9B (non-zero-shot) & \best{67.1}$^\diamond$ (66.1) \\
    \bottomrule
  \end{tabular}
\end{table}

\subsection{Confidence estimates from vote agreement}
\label{sec:cityscapes-vote-agreement}

\looseness=2 We define the vote agreement of a pixel as the fraction of passes that vote
for the majority class, counted over the passes that assign the pixel a class.
It requires no additional computation and serves as a per-pixel confidence
estimate. We evaluate it for the zero-shot \model{} 9B on the Cityscapes
validation set with \(R{=}30\), counting unassigned pixels as errors. Pixels
with agreement \(\geq 0.9\) cover 93.7\% of the labeled pixels, and 97.8\% of
them are classified correctly. The remaining 6.3\% of pixels have agreement
below 0.9; 44.9\% of them are misclassified, and they hold 57.6\% of all
errors in the validation set. These low-agreement pixels concentrate at class
boundaries: 37.8\% of them lie within 2 pixels of a ground-truth class
boundary, whereas only 7.2\% of all labeled pixels do. Agreement measures the consistency of the passes: it ranks pixels by reliability and should not be read as a calibrated probability of being correct. \Cref{fig:cityscapes-vote-agreement}
shows agreement maps for the two images of
\cref{fig:cityscapes-scaling-progression}.

\begin{figure*}[t]
    \centering
    \includegraphics[width=\textwidth]{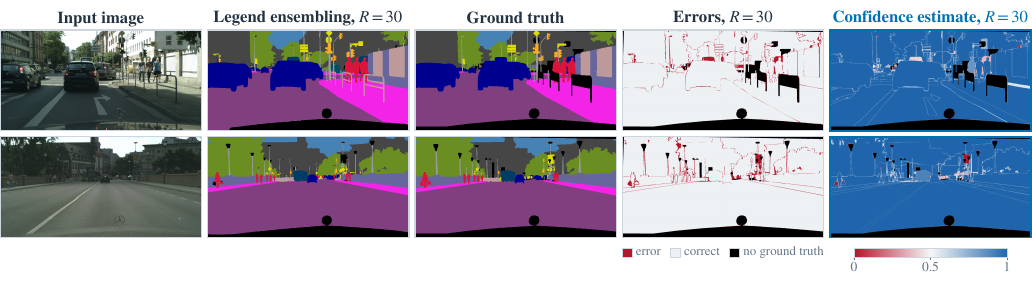}
    \caption{\textbf{Vote agreement as a confidence estimate.}
    \model{} 9B with \(R{=}30\) passes on the two Cityscapes images of \cref{fig:cityscapes-scaling-progression}. \emph{Errors}: pixels where the majority vote differs from the ground truth or is unassigned. \emph{Confidence estimate}: vote agreement (\cref{sec:cityscapes-vote-agreement}). Black: no ground truth. Errors and low agreement appear in the same places, mainly along object outlines and on small objects.}
    \label{fig:cityscapes-vote-agreement}
\end{figure*}

\subsection{Full final-model geometry results}
\label{sec:final-geometry-results}

Tables~\ref{tab:final-depth-full} and \ref{tab:final-normal-full} extend \cref{tab:final-geometry} with more benchmarks and published specialists. We evaluate the same four merged models as in \cref{tab:main_perception} (\cref{sec:supp-default-optimization}).

\looseness=1 We evaluate geometry with the training prompts (pixel-explicit inverse depth and concise normals; Sec.~\ref{sec:exact-geometry-prompts}) and generation seed 0. Depth generation uses the preprocessed input dimensions instead of the 512-pixel training buckets: $1008\!\times\!672$ for ETH3D and the input crop size for KITTI. Normal predictions are resized to the reference grid before decoding. Masks, alignment and scoring follow Sec.~\ref{sec:apdx-evaluation}.

\begin{table}[t]
\centering
\caption{Full depth comparison. Each cell gives AbsRel $\downarrow$ / $\delta_1\uparrow$, both in percent. Bold and underline mark the best and second-best reported OVB values; displayed ties share a rank. Published references are unranked. OVB inverse-gray predictions use disparity-affine alignment; published log-depth and metric-depth references retain their own protocols.}
\label{tab:final-depth-full}
\scriptsize
\setlength{\tabcolsep}{1.1pt}
\begin{adjustbox}{max width=\linewidth}
\begin{tabular*}{\columnwidth}{@{\extracolsep{\fill}}lccccc@{}}
\toprule
Method & NYU & KITTI & ETH3D & ScanNet & DIODE \\
\midrule
Marigold V2~\cite{pavlovic2026marigoldv2}$^{\dagger}$ & 3.6\,/\,98.0 & 5.4\,/\,97.4 & 2.8\,/\,99.2 & 3.7\,/\,97.9 & 5.2\,/\,97.1 \\
Marigold V1~\cite{ke2024repurposingdiffusionbasedimagegenerators}$^{\dagger}$ & 5.5\,/\,96.4 & 9.9\,/\,91.6 & 6.5\,/\,96.0 & 6.4\,/\,95.1 & 10.0\,/\,90.7 \\
Lotus-2~\cite{he2026lotus2advancinggeometricdense}$^{\dagger}$ & 3.7\,/\,97.6 & 6.7\,/\,94.1 & 4.1\,/\,98.6 & 4.0\,/\,97.2 & 6.5\,/\,95.5 \\
FE2E~\cite{wang2026editordensegeometryestimator}$^{\dagger}$ & 3.8\,/\,97.6 & 6.5\,/\,96.0 & 3.8\,/\,98.7 & 4.3\,/\,97.1 & 5.6\,/\,96.4 \\
PPD (1024)~\cite{xu2025pixelperfectdepth}$^{\dagger}$ & 4.1\,/\,97.7 & 7.0\,/\,95.5 & 4.3\,/\,98.0 & 4.6\,/\,97.2 & 6.8\,/\,95.9 \\
PPD (512)~\cite{xu2025pixelperfectdepth}$^{\dagger}$ & 4.3\,/\,97.4 & 8.0\,/\,93.1 & 4.5\,/\,97.7 & 4.5\,/\,97.3 & 7.0\,/\,95.5 \\
DAV2 (54K)~\cite{yang2024depthanythingv2}$^{\dagger}$ & 5.4\,/\,97.2 & 8.6\,/\,92.8 & 12.3\,/\,88.4 & -- & 8.8\,/\,93.7 \\
DepthFM~\cite{gui2025depthfm}$^{\dagger}$ & 5.5\,/\,96.3 & 8.9\,/\,91.3 & 5.8\,/\,96.2 & 6.3\,/\,95.4 & -- \\
GeoWizard~\cite{fu2024geowizardunleashingdiffusionpriors}$^{\dagger}$ & 5.2\,/\,96.6 & 9.7\,/\,92.1 & 6.4\,/\,96.1 & 6.1\,/\,95.3 & 12.0\,/\,89.8 \\
Vision Banana~\cite{visionbanana2026}$^{\ddagger}$ & 8.1\,/\,94.8 & 10.7\,/\,91.5 & 10.3\,/\,93.5 & -- & -- \\
\midrule
OVB 9B ZS & 5.82\,/\,97.4 & \best{6.56}\,/\,\best{95.3} & \best{4.38}\,/\,98.2 & 8.87\,/\,93.4 & \best{6.96}\,/\,\best{95.4} \\
OVB 9B NZS & \best{3.23}\,/\,\best{99.0} & 14.80\,/\,83.2 & 4.63\,/\,\second{98.3} & \best{4.49}\,/\,\best{97.8} & 8.89\,/\,\second{94.1} \\
OVB 4B ZS & 5.12\,/\,97.5 & \second{7.70}\,/\,\second{92.5} & 4.62\,/\,98.2 & 6.39\,/\,95.8 & 8.38\,/\,\second{94.1} \\
OVB 4B NZS & \second{3.56}\,/\,\second{98.7} & 13.72\,/\,85.9 & \second{4.39}\,/\,\best{98.4} & \second{4.72}\,/\,\second{97.5} & \second{7.75}\,/\,93.6 \\
\bottomrule
\end{tabular*}
\end{adjustbox}
\vspace{2pt}
\parbox{\columnwidth}{\scriptsize
$^{\dagger}$Published scores from~\cite{pavlovic2026marigoldv2}. $^{\ddagger}$Vision Banana~\cite{visionbanana2026} scores report metric depth whereas OVB predicts aligned depth per image. VB's KITTI split is not matched to Eigen, and its indoor-only DIODE score is omitted here. ZS/NZS: zero-shot/non-zero-shot, PPD: Pixel-Perfect Depth, and DAV2: Depth Anything V2.}
\end{table}

\begin{table}[t]
\centering
\caption{Full surface-normal comparison. Each cell gives mean angular error $\downarrow$ (degrees) / percentage below $11.25^\circ\uparrow$. Bold and underline mark the best and second-best reported OVB values; displayed ties share a rank. Published references are unranked. The non-zero-shot OVB models improve NYUv2 mean error but remain slightly worse on ScanNet.}
\label{tab:final-normal-full}
\scriptsize
\setlength{\tabcolsep}{1.1pt}
\begin{tabular*}{\columnwidth}{@{\extracolsep{\fill}}lcccc@{}}
\toprule
Method & NYU & ScanNet & iBims-1 & Sintel \\
\midrule
Marigold V2~\cite{pavlovic2026marigoldv2}$^{\dagger}$ & 16.6\,/\,61.2 & 14.1\,/\,67.4 & 15.9\,/\,70.9 & 28.7\,/\,27.6 \\
DSINE~\cite{DSINE}$^{\dagger}$ & 16.4\,/\,59.6 & 16.2\,/\,61.0 & 17.1\,/\,67.4 & 34.9\,/\,21.5 \\
Lotus-2~\cite{he2026lotus2advancinggeometricdense}$^{\dagger}$ & 16.9\,/\,59.0 & 14.2\,/\,66.8 & 15.4\,/\,70.4 & 30.3\,/\,27.6 \\
FE2E~\cite{wang2026editordensegeometryestimator}$^{\dagger}$ & 16.2\,/\,59.6 & 13.8\,/\,67.2 & 15.1\,/\,70.6 & 31.2\,/\,22.3 \\
GeoWizard~\cite{fu2024geowizardunleashingdiffusionpriors}$^{\dagger}$ & 18.9\,/\,50.7 & 17.4\,/\,53.8 & 19.3\,/\,63.0 & 40.3\,/\,12.3 \\
Vision Banana~\cite{visionbanana2026} & 17.78\,/\,-- & 15.05\,/\,-- & -- & -- \\
\midrule
OVB 9B ZS & 19.85\,/\,44.3 & \second{19.07}\,/\,\second{45.5} & \second{16.66}\,/\,\second{67.0} & 29.88\,/\,\second{28.5} \\
OVB 9B NZS & \second{13.51}\,/\,\second{67.0} & 19.86\,/\,40.7 & 17.18\,/\,65.4 & \best{28.63}\,/\,\best{30.6} \\
OVB 4B ZS & 19.87\,/\,45.1 & \best{18.81}\,/\,\best{48.7} & \best{16.37}\,/\,\best{67.4} & 34.18\,/\,22.0 \\
OVB 4B NZS & \best{13.21}\,/\,\best{67.8} & 20.26\,/\,39.5 & 17.45\,/\,62.6 & \second{29.42}\,/\,25.8 \\
\bottomrule
\end{tabular*}
\vspace{2pt}
\parbox{\columnwidth}{\scriptsize $^{\dagger}$Published values from~\cite{pavlovic2026marigoldv2}. OVB uses a fixed camera-facing convention without test-image calibration.}
\par\medskip
\textbf{Additional benchmarks}\par\smallskip
\label{tab:final-normal-extra}
\scriptsize
\setlength{\tabcolsep}{4pt}
\begin{tabular*}{\columnwidth}{@{\extracolsep{\fill}}lcc@{}}
\toprule
Model & DIODE indoor (325) & VKITTI2 (1000) \\
\midrule
Vision Banana~\cite{visionbanana2026} & 13.82\,/\,-- & 29.06\,/\,-- \\
\midrule
9B zero-shot & 20.53\,/\,18.5 & 34.54\,/\,48.7 \\
9B non-zero-shot & \best{11.43}\,/\,\best{76.4} & \best{30.48}\,/\,\best{53.4} \\
4B zero-shot & 20.79\,/\,25.6 & 35.93\,/\,46.3 \\
4B non-zero-shot & \second{12.37}\,/\,\second{72.1} & \second{32.15}\,/\,\second{51.7} \\
\bottomrule
\end{tabular*}
\vspace{2pt}

\end{table}

\paragraph{Paired geometry examples.}
Figure~\ref{fig:geometry-nzs-comparison} compares the selected 9B models using the same prompts and inference settings as above. The NYUv2 depth and ScanNet normal examples were selected by visual inspection; the latter includes an artifact case. The KITTI depth and NYUv2 normal examples have the median zero-shot error within fixed sets of 20 candidate images, not within the full benchmarks.
\begin{figure*}[t]
\centering
\includegraphics[width=\textwidth]{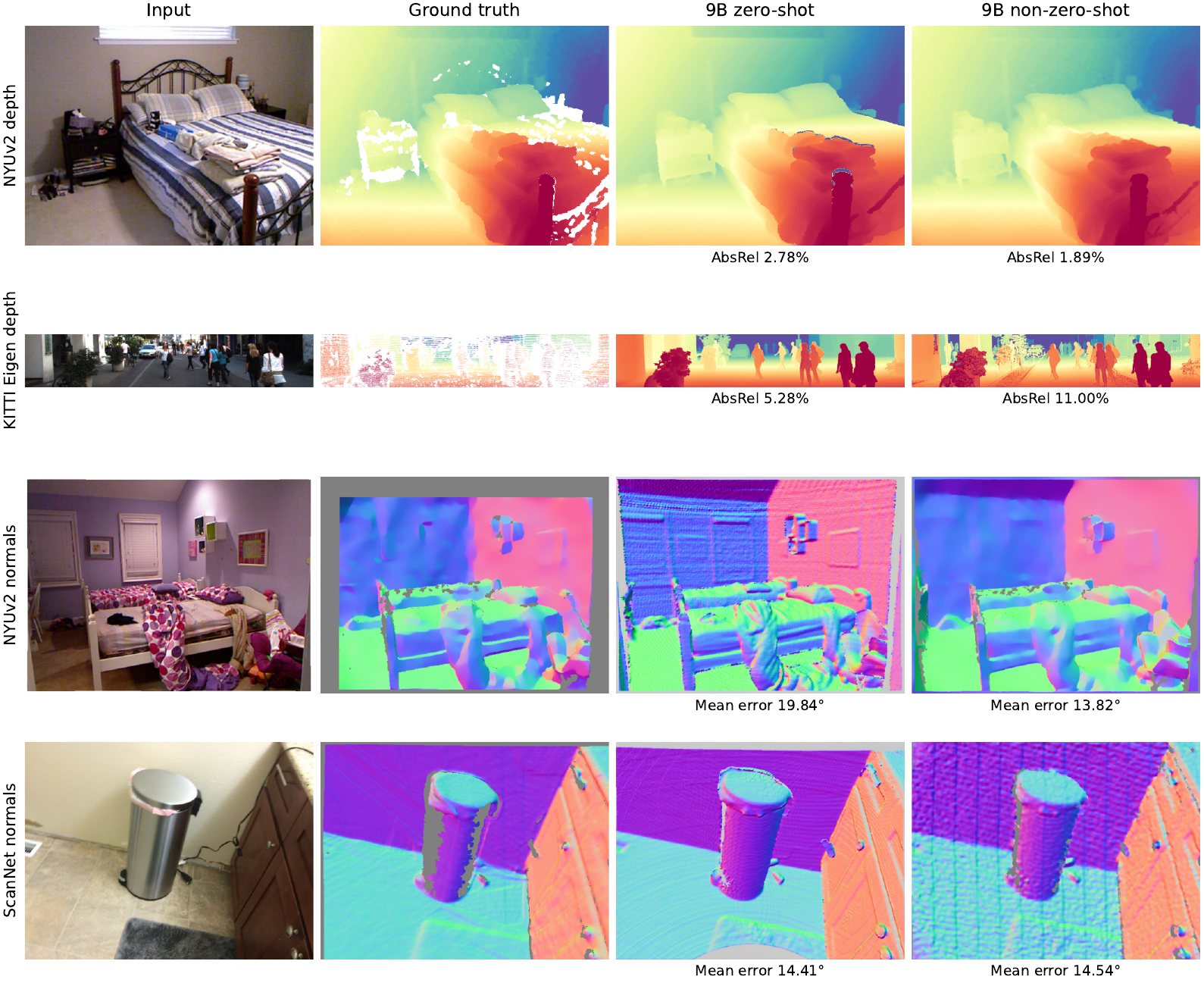}
\caption{\textbf{Paired zero-shot and non-zero-shot predictions.} The 9B non-zero-shot model improves both NYUv2 examples, with less texture in the normal map. KITTI depth does not improve, and the ScanNet wall remains textured. Depth uses disparity-affine alignment and a shared color scale within each row (red near, blue far). White depth and gray normal regions mark invalid ground truth; predictions are shown without a validity overlay. Scores below predictions are per-image errors.}
\label{fig:geometry-nzs-comparison}
\end{figure*}

\raggedbottom %
\subsection{Extended image editing results}
\label{sec:extended-imgeditbench-results}

\cref{tab:imgedit_results_full} gives the per-edit-type breakdown of the ImgEdit-Bench summary reported in \cref{tab:imgedit_summary}.

\providecommand{\xmark}{\textcolor{red!80!black}{\ding{55}}}
\providecommand{\second}[1]{\underline{#1}}

\begin{table*}[t]
\centering
\caption{\textbf{ImgEdit-Bench results, per edit type.} Mean over three
generation seeds, extending \cref{tab:imgedit_summary} with the Basic-suite
breakdown. \emph{Zero-shot mix} and \emph{non-zero-shot mix} refer to the
perception training mixture (\cref{sec:data-protocol}), neither includes ImgEdit-Bench data, and all rows are zero-shot on this benchmark.}
\label{tab:imgedit_results_full}

\setlength{\tabcolsep}{1.5pt}
\renewcommand{\arraystretch}{1.05}
\small
\begin{tabular}{@{}lccccccccc|cc@{}}
\toprule
\textbf{Model} &
\textbf{adjust} & \textbf{style} & \textbf{backgr.} & \textbf{extract} &
\textbf{remove} & \textbf{replace} & \textbf{add} & \textbf{compose} &
\textbf{action} &
\textbf{Basic avg.} $\uparrow$ & \textbf{UGE avg.} $\uparrow$ \\
\midrule
\multicolumn{12}{@{}l}{\textit{FLUX.2 Klein 4B backbone}} \\
\fluxkfourb{} & 3.43 & \best{2.97} & \best{4.13} & 4.00 & 2.97 & \second{3.29} & \second{4.64} & 3.38 & \best{3.98} & 3.66 & 3.69 \\
\rowcolor{ovbzeroshot}  \quad \model{} 4B (zero-shot mix) & \best{3.82} & \second{2.93} & 4.05 & \best{4.05} & \second{3.81} & 3.19 & 4.60 & \best{3.51} & \second{3.91} & \second{3.78} & \second{3.75} \\
\rowcolor{ovbnonzeroshot} \quad \model{} 4B (non-zero-shot mix) & \second{3.81} & 2.70 & \second{4.06} & \second{4.01} & \best{4.16} & \best{3.37} & \best{4.67} & \second{3.49} & 3.90 & \best{3.82} & \best{3.79} \\
\midrule
\multicolumn{12}{@{}l}{\textit{FLUX.2 Klein 9B backbone}} \\
\fluxknineb{} & \best{4.06} & \second{3.02} & \best{4.22} & \best{4.03} & 3.71 & \best{3.45} & \best{4.75} & 3.49 & \best{4.00} & \best{3.89} & \best{4.03} \\
\rowcolor{ovbzeroshot}  \quad \model{} 9B (zero-shot mix) & \second{3.84} & \best{3.14} & 3.99 & \second{3.89} & \best{4.19} & 3.31 & \best{4.75} & \second{3.52} & \second{3.94} & \second{3.86} & \second{3.95} \\
\rowcolor{ovbnonzeroshot} \quad \model{} 9B (non-zero-shot mix) & 3.80 & \best{3.14} & \second{4.12} & \best{4.03} & \second{4.11} & \second{3.35} & \second{4.74} & \best{3.58} & \best{4.00} & \best{3.89} & 3.68 \\
\bottomrule
\end{tabular}
\end{table*}

\section{Extended Analysis of Design Choices}
\label{sec:design-analysis}
This appendix extends the design-choice analysis in Sec.~\ref{sec:ablations}.
Its studies use ablation checkpoints rather than the selected models of
Appendix~\ref{sec:consolidated-benchmark-results}, so scores are compared
only within each study.

\flushbottom
\subsection{Segmentation}
\label{sec:supp-2d-analysis}
These studies examine how the requested categories and instance-color
conventions affect segmentation.

\subsubsection{Number of requested classes}
\label{sec:supp-set-segmentation}
\setrqprefix{S}{violet}\setcounter{researchquestion}{2}
\researchquestion{S7}{How does accuracy change with the number of requested foreground classes?}

For each requested class count $K$, we evaluate the Top-64 palette model of \cref{rq:color-segmentation} (\cref{tab:palette_sampling}) and request $\min(K,p_i)$ classes per image, drawn uniformly at random from the $p_i$ foreground classes present in image~$i$. The class draw is seeded per image, so the legends are fixed for a given $K$. The generation seed is fixed as
well.

Unlike the \textit{set} evaluation protocol, which requests random strict-subset, large $K$ can request all present classes. As in the protocol, only pixels of requested classes are scored, so predictions in unrequested regions are not penalized.
Cityscapes uses 500 images, COCO-Stuff and COCO-Things use 493 and 496 after excluding images with no foreground classes. The largest present foreground class counts are 16, 9 and 6, respectively. Larger requests repeat the full-set condition (requesting all classes present in each image), so the curves plateau beyond these counts. \cref{fig:requested-classes} plots mIoU against $K$ for the three benchmarks.

\begin{figure}[t]
\centering
\includegraphics[width=\columnwidth]{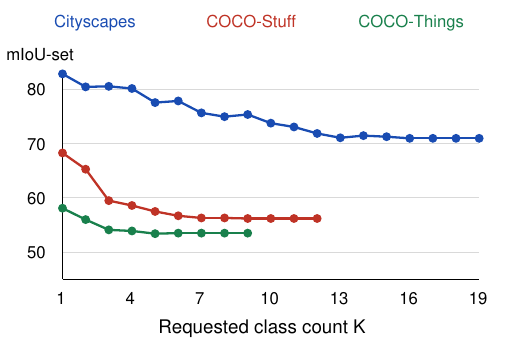}
\caption{mIoU versus the number of requested classes $K$.}
\label{fig:requested-classes}
\end{figure}

Cityscapes falls from 82.9 mIoU at \(K{=}1\) to 71.0 at \(K{=}16\).
Both COCO datasets also score lower at their largest requests than at \(K{=}1\). Requesting fewer classes therefore helps, which motivates the legend pruning of \cref{sec:coco-object-filter-analysis}.

\subsubsection{Instance colors and prompts}
\label{sec:supp-instance-prompting}
\researchquestion{color-instance-seg}{How should instance colors, ordering and prompts be chosen?}
All variants use the same 32 separated named colors and matched training and evaluation conventions. We vary fixed versus random assignments, spatial ordering, and whether the prompt lists assignments, states only an ordering rule, or leaves both unstated.

\begin{table}[H]
    \centering
    \rqcaption[color-instance-seg]{
    \textbf{Instance-segmentation on SA-Co/Gold.} 
    \emph{Random}/\emph{Fixed} denotes per-sample permutation/fixed
    instance$\rightarrow$color mapping.
    Scans use the indicated primary direction, with the other direction for
    tie-breaking.
    \emph{Stated} enumerates all assignments in the prompt;
    \emph{Unstated} specifies neither assignments nor scan order; and
    \emph{Implicit} states the scan and the
    $i\text{-th instance} \rightarrow i\text{-th color}$
    rule without listing the colors.
    }
    \label{tab:instance_prompting}
    \setlength{\tabcolsep}{4pt}
    \renewcommand{\arraystretch}{1.08}
    \small
    \begin{tabular}{@{}cccccc@{}}
        \toprule
        \textbf{Map} &
        \textbf{Scan} &
        \textbf{List} &
        \textbf{cgF1} &
        \textbf{MCC} &
        \textbf{pmF1} \\
        \midrule
        Random & Top$\rightarrow$Bottom & Stated
            & 30.82 & 0.80 & 38.18 \\
        Random & Unordered & Unstated
            & 33.94 & \underline{0.81} & 41.59 \\
        Fixed & Top$\rightarrow$Bottom & Stated
            & 32.43 & \textbf{0.82} & 39.27 \\
        \rowcolor{segmentationreferencecolor} Fixed & Left$\rightarrow$Right & Stated
            & \underline{34.32} & \underline{0.81}
            & \underline{41.74} \\
        Fixed & Top$\rightarrow$Bottom & Implicit
            & \textbf{34.72} & \underline{0.81} & \textbf{42.54} \\
        \bottomrule
    \end{tabular}
\end{table}

As can be seen in \cref{tab:instance_prompting}, with stated top-to-bottom assignments, fixed colors beat random ones by 1.61 cgF1 (32.43 vs.\ 30.82), while random, unordered, unstated assignments reach 33.94. Changing the fixed scan to left-to-right raises cgF1 from 32.43 to 34.32. Fixed top-to-bottom assignments with only the ordering rule stated score highest, 34.72. MCC stays between 0.80 and 0.82. We nevertheless adopt fixed left-to-right assignments, stated in full, as the default: the gap to the implicit variant is 0.40 cgF1, that variant also changes the scan direction, and a stated legend fixes the instance-to-color mapping explicitly rather than relying on the model to apply the rule. Scan direction and prompt wording thus matter at least as much as color randomization.

\setlength{\floatsep}{10pt plus 2pt minus 2pt}
\setlength{\dblfloatsep}{10pt plus 2pt minus 2pt}
\setlength{\textfloatsep}{12pt plus 2pt minus 2pt}
\setlength{\dbltextfloatsep}{12pt plus 2pt minus 2pt}
\setlength{\intextsep}{6pt plus 1pt minus 1pt}
\makeatletter
\setlength{\@fptop}{0pt}
\setlength{\@fpsep}{6pt}
\setlength{\@fpbot}{0pt plus 1fil}
\setlength{\@dblfptop}{0pt}
\setlength{\@dblfpsep}{6pt}
\setlength{\@dblfpbot}{0pt plus 1fil}
\makeatother
\subsection{Geometry}
\label{sec:geometry-ablations}
\label{sec:geometry-full-splits}
These studies use the reduced geometry training setup of
Sec.~\ref{sec:ablations}. Evaluation follows~\cref{sec:metrics}.
Other-task scores measure retention in the same checkpoints.

\subsubsection{Prompt and representation}
\label{sec:prompt-ablation}
\setrqprefix{S}{violet}\setcounter{researchquestion}{4}
\researchquestion{setting-geometry}{How do prompts and output representations affect geometry accuracy?}
For each task, we train a separate 37.5K-step model for each prompt: concise, pixel-explicit and scene-preserving. Each run assigns the full 30\% geometry share to that task. All three depth prompts use inverse-gray targets; a fourth depth run pairs the scene-preserving prompt with metric-RGB targets. All normal runs use canonical RGB targets.
Each checkpoint is evaluated with its training prompt. Exact instructions appear in
Sec.~\ref{sec:exact-geometry-prompts}.

\begin{table}[t]
\centering
\caption{Prompt--representation configurations after matched 37.5K fine-tunes. Depth and normals each use the five complete splits defined in Sec.~\ref{sec:geometry-ablations}; entries average dataset scores equally. Depth and normal scores come from separate task-isolated checkpoints, each evaluated with its training prompt.}
\label{tab:appendix-geometry-prompts-refreshed}
\scriptsize
\setlength{\tabcolsep}{2pt}
\begin{tabularx}{\columnwidth}{@{}Xcccc@{}}
\toprule
 & \multicolumn{2}{c}{Depth} & \multicolumn{2}{c}{Normal} \\
\cmidrule(lr){2-3} \cmidrule(lr){4-5}
Configuration & AbsRel$\downarrow$ & $\delta_1$ (\%)$\uparrow$ & Mean$\downarrow$ & Med.$\downarrow$ \\
\midrule
Concise inverse / normal          & \second{0.060} & \second{96.2} & \second{23.95} & \second{14.75} \\
Pixel-explicit inverse / normal   & \best{0.059}   & \second{96.4}   & \best{23.64}   & \best{14.34} \\
Scene-preserving inverse / normal & 0.061          & \best{96.4}   & 23.98          & 14.91 \\
Scene-preserving metric RGB depth & 0.062          & 96.0          & --             & -- \\
\bottomrule
\end{tabularx}
\end{table}

Pixel-explicit wording gives the lowest depth AbsRel and normal errors, although normal means differ by less than $0.4^\circ$. Pixel-explicit and scene-preserving depth $\delta_1$ both round to 96.4\%. Metric RGB improves neither depth metric under its corresponding alignment. The full-scale models use the concise normal prompt selected in an earlier 20K-step study; its mean error here is $0.3^\circ$ higher than pixel-explicit.

\subsubsection{Supervision and task mixture}
\label{sec:training-supervision-composition}
\label{sec:mask-ablation}
\label{sec:source-ablation}
\label{sec:ratio-ablation}
Table~\ref{tab:appendix-joint-full}(a) varies one training choice at a time from a 37.5K-step reference run with validity masks. Training masks follow
Sec.~\ref{sec:recipe_training}; evaluation masks remain fixed.
Erosion removes valid pixels within eight pixels of an invalid region
before projection to the latent grid.
\begin{table*}[t]
\centering
\caption{Geometry training studies. Geometry scores average the five complete splits per task in Sec.~\ref{sec:geometry-full-splits}; CS, CO, RS, IS and IE are the higher-is-better retention metrics in Sec.~\ref{sec:ablation-evaluation}. (a)~Supervision and data mixture at 37.5K, with a masked all-source reference run. (b)~Training duration under a fixed recipe. (c)~Combined changes at 50K, using a separately trained reference run. The full combination uses the 10/20 allocation, eight-pixel erosion, and removal of Replica depth and Synscapes normals. Only the full combination is repeated across three seeds.}
\label{tab:appendix-joint-full}
\label{tab:main-training-duration}
\label{tab:appendix-geometry-duration-refreshed}
\label{tab:appendix-phase2-full}
\footnotesize
\setlength{\tabcolsep}{3pt}
\begin{tabular*}{\textwidth}{@{\extracolsep{\fill}}lccccccccc@{}}
\toprule
& \multicolumn{2}{c}{Depth} & \multicolumn{2}{c}{Normals} & \multicolumn{5}{c}{Retention} \\
\cmidrule(lr){2-3}\cmidrule(lr){4-5}\cmidrule(l){6-10}
Variant & AbsRel$\downarrow$ & $\delta_1$ (\%)$\uparrow$ & Mean ($^\circ$)$\downarrow$ & Med. ($^\circ$)$\downarrow$ & CS & CO & RS & IS & IE \\
\midrule
\multicolumn{10}{@{}l}{\textbf{(a) Supervision and data mixture, 37.5K}} \\
Masked reference (20/10) & 0.0645 & 96.35 & 23.10 & 14.29 & 15.85 & 18.51 & 36.71 & 18.28 & 3.56 \\
\multicolumn{10}{@{}l}{\textit{Training masks}} \\
No mask & 0.0719 & 95.72 & 27.27 & 18.37 & 19.58 & 11.93 & 43.40 & 18.20 & 3.54 \\
Erosion, 8 px & 0.0656 & 96.28 & 23.65 & 15.11 & 15.51 & 16.56 & 41.05 & 17.76 & 3.64 \\
\multicolumn{10}{@{}l}{\textit{Geometry sources}} \\
No Hypersim & 0.0625 & 96.25 & 29.90 & 20.53 & 30.82 & 20.86 & 42.64 & 16.53 & 3.64 \\
No Matterport3D & 0.0579 & 96.66 & 26.10 & 17.43 & 22.89 & 15.91 & 43.16 & 17.18 & 3.56 \\
No Replica & 0.0582 & 96.83 & 25.97 & 17.44 & 27.64 & 18.08 & 35.55 & 18.15 & 3.48 \\
No Synscapes & 0.0615 & 96.13 & 21.75 & 13.16 & 32.43 & 20.51 & 39.11 & 18.21 & 3.24 \\
No Taskonomy & 0.0589 & 96.74 & 25.06 & 16.31 & 29.67 & 19.91 & 34.41 & 16.64 & 3.84 \\
\multicolumn{10}{@{}l}{\textit{Depth/normal allocation, 30\% total}} \\
10/20 & 0.0668 & 96.02 & 22.13 & 13.17 & 24.02 & 17.73 & 37.48 & 13.48 & 3.30 \\
15/15 & 0.0651 & 95.95 & 24.14 & 15.31 & 17.32 & 13.45 & 45.87 & 16.56 & 3.44 \\
\midrule
\multicolumn{10}{@{}l}{\textbf{(b) Training duration}} \\
10K & 0.07264 & 95.401 & 26.00 & 18.18 & 21.45 & 5.89 & 11.65 & 2.84 & 3.96 \\
37.5K & 0.06454 & 96.354 & 23.10 & 14.29 & 15.85 & 18.51 & 36.71 & 18.28 & 3.56 \\
50K & 0.06080 & 96.497 & 24.90 & 16.54 & 27.40 & 21.19 & 45.76 & 17.47 & 4.00 \\
\midrule
\multicolumn{10}{@{}l}{\textbf{(c) Combined changes, 50K}} \\
Reference, seed 42 & 0.0621 & 96.53 & 22.65 & 13.24 & 30.19 & 13.71 & 39.56 & 17.33 & 3.52 \\
10/20 ratio & 0.0670 & 95.97 & 24.83 & 15.80 & 21.80 & 17.48 & 43.00 & 16.23 & 3.68 \\
Ratio + 8 px erosion & 0.0699 & 95.70 & 25.61 & 16.56 & 25.61 & 20.63 & 47.00 & 16.76 & 4.08 \\
Complete, seed 42 & 0.0630 & 96.29 & 22.00 & 13.13 & 26.62 & 16.65 & 40.15 & 17.10 & 3.90 \\
Complete, seed 123 & 0.0672 & 95.74 & 20.99 & 11.99 & 24.98 & 18.10 & 44.49 & 18.13 & 3.44 \\
Complete, seed 456 & 0.0666 & 95.65 & 21.11 & 12.14 & 19.61 & 14.79 & 34.67 & 16.69 & 3.56 \\
\bottomrule
\end{tabular*}
\end{table*}

\begin{table*}[t]
\centering
\begin{tabular}{@{}p{\columnwidth}@{\hspace{\columnsep}}p{\columnwidth}@{}}

\begin{minipage}[t]{\columnwidth}
\vspace{0pt}
  \centering
  \rqcaption[D1]{\textbf{Effect of replay and perception-data diversity.} 
  The 40\% replay reference uses the geometry-ablation mixture, not the full-scale OVB mixture. Depth scores average the full NYUv2 and KITTI Eigen splits with disparity-affine alignment; normal scores average NYUv2 and DIODE indoor. Each dataset has equal weight within its task. Other-task evaluations follow Sec.~\ref{sec:ablation-evaluation}.}
  \label{tab:main-data-scaling}
  \setlength{\tabcolsep}{0.8pt}
  \renewcommand{\arraystretch}{1.03}
  \resizebox{\columnwidth}{!}{%
  \begin{tabular}{@{}lccccccccc@{}}
    \toprule
    & \multicolumn{2}{c}{Depth} & \multicolumn{2}{c}{Normals} & \multicolumn{5}{c}{Retention} \\
    \cmidrule(lr){2-3}\cmidrule(lr){4-5}\cmidrule(l){6-10}
    Training data & AbsRel$\downarrow$ & $\delta_1\uparrow$ (\%) & Mean$\downarrow$ & Med.$\downarrow$ & CS$\uparrow$ & CO$\uparrow$ & RS$\uparrow$ & IS$\uparrow$ & IE$\uparrow$ \\
    \midrule
    \multicolumn{10}{@{}l}{\textit{(a) Replay fraction}} \\
    1\%  & 0.0670 & \best{96.01} & 21.91 & 16.62 & \best{28.77} & 11.71 & 41.58 & 16.32 & 2.74 \\
    5\%  & \best{0.0643} & \second{95.79} & 19.52 & 14.47 & 23.56 & \second{17.54} & 40.50 & \second{17.26} & \best{3.84} \\
    10\% & \second{0.0650} & 95.71 & \best{18.96} & \second{13.58} & \second{28.69} & \best{18.79} & \best{43.36} & 16.54 & 3.36 \\
    40\% reference & 0.0676 & 95.31 & \second{19.30} & \best{13.41} & 21.64 & 12.80 & \second{41.14} & \best{18.33} & \second{3.66} \\
    \addlinespace[2pt]
    \multicolumn{10}{@{}l}{\textit{(b) Unique perception examples}} \\
    10\% & \second{0.0704} & \best{95.34} & \second{21.06} & \second{15.94} & \best{22.71} & \best{17.51} & \second{41.09} & \best{21.42} & \best{3.76} \\
    30\% & 0.0709 & 95.24 & 21.12 & 15.97 & \second{22.06} & 12.40 & 30.43 & 16.47 & 3.32 \\
    100\% default & \best{0.0676} & \second{95.31} & \best{19.30} & \best{13.41} & 21.64 & \second{12.80} & \best{41.14} & \second{18.33} & \second{3.66} \\
    \bottomrule
  \end{tabular}}
\end{minipage}%
 &

\begin{minipage}[t]{\columnwidth}
\vspace{0pt}
\centering
\caption{Generation retention for the geometry ablations. IG-C is the GenAI-Bench CLIP--FlanT5 score, higher is better. Panels (a)--(c) use 37.5K checkpoints; (d) uses the 50K generation/editing-data runs from Table~\ref{tab:main-creative-data}. Compare rows within each panel, using its study's reference run.}
\label{tab:geometry-generation-retention}
\scriptsize
\setlength{\tabcolsep}{3pt}
\resizebox{\linewidth}{!}{%
\begin{tabular}[t]{@{}lr@{}}
\toprule
Setting & IG-C \\
\midrule
\multicolumn{2}{@{}l}{\textit{(a) Adaptation scope}} \\
Last 2 & 0.8118 \\
Last 4 & 0.8095 \\
Last 8 & 0.8078 \\
Last 12 & 0.8036 \\
All 24 & 0.7997 \\
Full except embeddings & 0.7962 \\
Full transformer & 0.7893 \\
\multicolumn{2}{@{}l}{\textit{(b) Replay fraction}} \\
1\% & 0.7954 \\
5\% & 0.7787 \\
10\% & 0.7919 \\
40\% & 0.7962 \\
\bottomrule
\end{tabular}\hspace{6pt}%
\begin{tabular}[t]{@{}lr@{}}
\toprule
Setting & IG-C \\
\midrule
\multicolumn{2}{@{}l}{\textit{(c) Unique perception examples}} \\
10\% & 0.7919 \\
30\% & 0.7747 \\
100\% & 0.7962 \\
\multicolumn{2}{@{}l}{\textit{(d) Generation/editing allocation}} \\
Generation 0\% & 0.8074 \\
Generation 7.5\% & 0.7639 \\
Generation 15\% & 0.7871 \\
Generation 30\% & 0.7840 \\
Editing/identity 0\% & 0.7635 \\
Editing/identity 12.5\% & 0.7673 \\
Editing/identity 25\% & 0.7871 \\
Editing/identity 50\% & 0.7731 \\
\bottomrule
\end{tabular}
}
\end{minipage}%

\end{tabular}
\end{table*}
Validity masking improves both geometry tasks; eight-pixel erosion gives
no additional gain at this budget.

Source-removal runs redistribute the omitted source's probability equally
among the remaining sources within each geometry task, keeping total
geometry sampling probability at 30\%.
Source removal often trades depth against normal accuracy; no exclusion improves
both depth metrics and normal mean error.

The allocation study changes only the division of the 30\% geometry budget. The 10/20 depth/normal allocation improves normals relative to 20/10 but worsens depth and several retention scores. Full-scale training uses 10/20 (Table~\ref{tab:task-sampling-weights}), fixed before this reduced-budget comparison favored 20/10. These results do not establish the best allocation at full scale.

\subsubsection{Training duration}
\label{sec:length-ablation}
\setrqprefix{T}{blue}
\researchquestion{geometry-training-duration}{How does performance change with training duration?}
We compare 10K, 37.5K and 50K checkpoints under the same data, prompts,
masks and optimization settings (Table~\ref{tab:main-training-duration}(b)).
Depth improves through 50K, whereas normal error is lowest at 37.5K.
The 50K checkpoint leads all retention metrics except instance segmentation,
which favors 37.5K.

\subsubsection{Combined choices and seed variation}
\label{sec:appendix-combined-choices}
\label{sec:phase2-validation}
Table~\ref{tab:appendix-phase2-full}(c) starts from a separately trained 50K-step reference using the duration-study recipe: 20/10 allocation, all five sources, uneroded validity masks, pixel-explicit depth and concise normals. We add the listed changes cumulatively. Neither switching to 10/20 nor then adding erosion improves geometry. The full combination also removes Replica depth and Synscapes normals; all three seeds lower normal error but raise depth error. Only this combination was repeated, so the normal gain cannot be separated from variation between reference runs.

\subsubsection{Replay and perception-data diversity}
\label{sec:phase2-scope-data}
\setrqprefix{D}{green}\setcounter{researchquestion}{1}
\researchquestion{D1}{How much replay and perception-data diversity are required?}
Table~\ref{tab:main-data-scaling} compares two 37.5K-step studies. The first varies the replay fraction from a 40\% reference, keeping generation, editing and identity at a 3:3:2 ratio and allocating the remaining probability to perception tasks. The second varies the perception pool size through deterministic nested subsets at fixed task probabilities. Lower replay improves depth relative to 40\%, but editing declines at 1\% and the preferred fraction varies by metric. The full perception pool gives the lowest depth AbsRel and normal mean error, while some retention metrics favor smaller pools.

\subsection{Parameter-Efficient Fine-Tuning}
\label{sec:supp-lora}

\looseness=1 Full fine-tuning of the 4B model needs 16 GH200 GPUs for about three days
(Sec.~\ref{sec:recipe_training}). To test whether the recipe transfers to a
parameter-efficient setting, we train LoRA adapters~\citep{hu2022lora} on
\fluxkfourbshort{} with the zero-shot mixture and evaluate them with the
protocol of Sec.~\ref{sec:apdx-evaluation}.

\paragraph{Setup.}
\looseness=1 All adapters are trained on the full zero-shot pool (\cref{tab:datasets}). Each run attaches a LoRA adapter to every linear layer of the transformer, with scale $\alpha/r=1$. The base weights stay frozen in bf16 and the adapters are trained in fp32. Optimizer, schedule, resolution buckets and the 1M-sample epochs follow~\cref{tab:training-fullscale}, with 500 warm-up steps. Because the base weights are frozen, we use plain data parallelism instead of FSDP, with 4 samples per GPU on 16 GPUs (global batch 64) and a peak memory of 66\,GiB per GPU. A 37.5K-step run takes about 9\,h on 16 GH200 GPUs, and the same recipe should also fit, more slowly, on a single 4-GPU node with 4-step gradient accumulation. Each result is a single evaluation run with the fixed seeds of the evaluation protocol. A rank-256 learning-rate sweep found $3$--$5\!\times\!10^{-5}$ usable; $10^{-4}$ and above degrade or diverge, while $10^{-5}$ learns too slowly. All runs below use $5\!\times\!10^{-5}$.

\begin{table*}[t]
  \centering
  \caption{\textbf{LoRA rank and training length on \fluxkfourbshort{}.} Shaded rows are the unmerged 37.5K and 100K full fine-tuning checkpoints. 100K LoRA steps correspond to 6.4M samples, and 37.5K full fine-tuning steps to 7.2M. SA-Co/Gold uses the 14K-query subset (Sec.~\ref{sec:subset-construction}), and ReasonSeg uses the Gemma4~31B query rewrites of the main paper. Bold and underline rank the LoRA rows only.}
  \label{tab:lora}
  \setlength{\tabcolsep}{3pt}
  \renewcommand{\arraystretch}{1.05}
  \resizebox{\textwidth}{!}{%
  \begin{tabular}{@{}llrccccccccc@{}}
    \toprule
    & & & \multicolumn{2}{c}{Semantic seg.\ (mIoU$\uparrow$)} & Ref.\ seg. & Reas.\ seg. & Inst.\ seg. & \multicolumn{2}{c}{ImgEdit (score$\uparrow$)} & Depth & Normals \\
    \cmidrule(lr){4-5}\cmidrule(lr){9-10}
    Adaptation & Steps & Trainable & Cityscapes & COCO-Object & RefCOCOg cIoU$\uparrow$ & ReasonSeg gIoU$\uparrow$ & SA-Co/Gold cgF1$\uparrow$ & Basic & UGE & NYUv2 $\delta_1\uparrow$ (\%) & NYUv2 Mean ($^\circ$)$\downarrow$ \\
    \midrule
    \rowcolor{defaultcolor} Full fine-tuning & 37.5K & 3.9B & 63.0 & 24.6 & 53.0 & 66.4 & 30.2 & 3.75 & 3.59 & 95.1 & 19.76 \\
    \rowcolor{defaultcolor} Full fine-tuning & 100K & 3.9B & 65.8 & 30.6 & 55.9 & 69.2 & 26.8 & 3.80 & 3.51 & 97.2 & 19.34 \\
    \midrule
    \multirow{5}{*}{LoRA $r{=}64$} & 25K & \multirow{5}{*}{102M} & 42.7 & 11.2 & 46.3 & 63.3 & 28.5 & \best{3.79} & \best{3.88} & 93.6 & 19.96 \\
     & 37.5K &  & 50.3 & 11.4 & 46.3 & 63.6 & 27.5 & \second{3.78} & 3.63 & 95.9 & \second{18.95} \\
     & 50K &  & 53.5 & 12.2 & 42.2 & 62.2 & \second{28.8} & 3.75 & \second{3.74} & 96.1 & 19.56 \\
     & 75K &  & \second{56.8} & \best{20.5} & \best{50.6} & \best{65.7} & 27.9 & 3.65 & 3.51 & 97.0 & 19.00 \\
     & 100K &  & \best{57.7} & 19.0 & 47.0 & 60.6 & \best{29.4} & 3.70 & 3.56 & \second{97.1} & 19.83 \\
    \midrule
    \multirow{5}{*}{LoRA $r{=}128$} & 25K & \multirow{5}{*}{203M} & 49.7 & 13.9 & 44.2 & 61.4 & 28.4 & 3.71 & 3.19 & 96.8 & 19.16 \\
     & 37.5K &  & 50.8 & 14.6 & \second{48.8} & \second{64.4} & 27.6 & 3.71 & 3.61 & 96.1 & 19.66 \\
     & 50K &  & 51.2 & 11.0 & 41.1 & 60.6 & 19.1 & 3.74 & 3.70 & \second{97.1} & 19.22 \\
     & 75K &  & 52.3 & 16.3 & 46.1 & 59.7 & 27.4 & 3.60 & 3.60 & 96.4 & 19.11 \\
     & 100K &  & 55.7 & \second{19.3} & 45.9 & 61.6 & 25.9 & 3.67 & 3.21 & 96.2 & 19.48 \\
    \midrule
    \multirow{5}{*}{LoRA $r{=}256$} & 25K & \multirow{5}{*}{406M} & 46.5 & 10.7 & 41.3 & 59.5 & 26.3 & 3.57 & 3.13 & 96.4 & 19.59 \\
     & 37.5K &  & 52.1 & 14.0 & 44.6 & 63.8 & 24.3 & 3.69 & 3.21 & 94.8 & 20.69 \\
     & 50K &  & 52.4 & 11.5 & 44.6 & 61.8 & 27.5 & 3.56 & 3.23 & 96.4 & \best{18.80} \\
     & 75K &  & 52.3 & 13.4 & 44.3 & 60.4 & 27.4 & 3.42 & 3.48 & \best{97.6} & 19.51 \\
     & 100K &  & 55.1 & 12.8 & 42.4 & 53.1 & 23.6 & 3.50 & 3.21 & 96.6 & 19.01 \\
    \midrule
    LoRA $r{=}512$ & 37.5K & 812M & 47.6 & 10.3 & 37.1 & 58.1 & 26.6 & 3.41 & 3.21 & 95.4 & 20.18 \\
    LoRA $r{=}1024$ & 37.5K & 1.63B & 48.6 & 7.0 & 34.0 & 47.3 & 23.9 & 3.11 & 2.77 & 95.1 & 19.32 \\
    \bottomrule
  \end{tabular}}
\end{table*}

\paragraph{Rank and training length.}
\looseness=1 Table~\ref{tab:lora} compares ranks 64 to 1024 at 37.5K steps and continues ranks 64, 128 and 256 to 100K. At 37.5K, ranks 64 and 128 are the best adapters on all tasks except Cityscapes, and ranks 512 and 1024 are worse on nearly every task. With 102M--203M trainable parameters, ranks 64 and 128 match or exceed full fine-tuning at the same step count on image editing, depth and surface normals, trail it by 2--7 points on referring, reasoning and instance segmentation, and trail most on semantic segmentation. Longer training improves semantic segmentation for ranks 64 and 128, and beyond 50K it hurts rank 256 on most tasks. Because full fine-tuning uses a three times larger batch, we also compare the 100K-step adapters with the 37.5K-step full fine-tuning checkpoint, which has seen a similar number of samples. The best adapter still trails it by 5 mIoU on Cityscapes and 4 on COCO-Object, but is close on image editing and better on depth and normals. Overall, LoRA recovers most of the editing and geometry ability of \model{} with a fraction of the trainable parameters, while dense semantic segmentation benefits most from updating all weights.

\section{Prompt Definitions}
\label{sec:prompt-definitions}

This section records the model-facing prompt families used across tasks.

\subsection{Prompts for 2D tasks}
\label{sec:prompt-templates}

\paragraph{Semantic segmentation.}
Each prompt begins with \texttt{[TASK: SEMANTIC\_SEGMENTATION]}, followed by
one of ten templates sampled with equal probability. The complete prompt therefore reads
\begin{promptblock}\texttt{[TASK: SEMANTIC\_SEGMENTATION]} $\langle$sampled template$\rangle$\end{promptblock}
Six dictionary-style templates embed the class--color mapping as a legend
\texttt{\{legend\}} of the form \texttt{\{"road": "sky blue", "car": "red", ...\}}:
\begin{promptblock}
\begin{enumerate}[leftmargin=*]
    \item Generate a semantic segmentation visualization of the input image, using this color mapping: \texttt{\{legend\}} Color each pixel of a listed class with its color and leave everything else black. Output only the segmentation visualization image.

    \item Segment only the following classes, using exactly these colors: \texttt{\{legend\}} All other pixels must be black. Output only the segmentation image.

    \item Produce a segmentation mask that highlights only these classes with the given colors: \texttt{\{legend\}} Everything not listed must be black. Output just the mask image.

    \item Using the color legend below, paint each named class in its color and leave the rest black: \texttt{\{legend\}} Output only the segmentation visualization image.

    \item Generate a semantic segmentation visualization image, using this color mapping: \texttt{\{legend\}}.

    \item Generate a visualization image of semantic segmentation, using this color mapping: \texttt{\{legend\}}.
\end{enumerate}
\end{promptblock}
Four sentence-style templates consist of a lead-in followed by one clause per
requested class:
\begin{promptblock}
\begin{enumerate}[leftmargin=*]
    \item This image is a per-pixel class labeling of the input. The \texttt{\{class\}} are represented by \texttt{\{color\}}. \ldots

    \item Generate a semantic segmentation visualization of the input. The \texttt{\{class\}} is \texttt{\{color\}}. \ldots

    \item Produce a per-pixel semantic segmentation of the image. The \texttt{\{class\}} should be colored \texttt{\{color\}}. \ldots

    \item Color the input image by semantic class. Paint the \texttt{\{class\}} \texttt{\{color\}}. \ldots
\end{enumerate}
\end{promptblock}
\noindent Each prompt writes all of its colors in one notation: an everyday
color name (e.g., \texttt{sky blue}), an RGB list (e.g., \texttt{[220,30,30]}),
or a hexadecimal code (e.g., \texttt{\#DC1E1E}). In dictionary legends, names
are quoted, whereas RGB and hexadecimal values are not. The notation is sampled
uniformly per prompt; names are unavailable when the legend exceeds the
37-color named palette. Only the first four dictionary templates state that
unlisted pixels are black, but the target renders them black for every
template. The released per-example manifest stores the fully rendered prompt.

\raggedbottom %
\paragraph{Referring segmentation.}
Each prompt begins with \texttt{[TASK: REFERRING\_SEGMENTATION]}, followed by
one of three templates sampled uniformly. So, the complete prompt structure is
\begin{promptblock}\texttt{[TASK: REFERRING\_SEGMENTATION]} $\langle$sampled template$\rangle$\end{promptblock}
\flushbottom
Here, \texttt{\{expression\}} is the referring expression, \texttt{\{color\}} a
color name sampled from the 37-color named palette, and \texttt{\{rgb\}} its RGB
value (e.g., \texttt{[220,30,30]}):
\begin{promptblock}
\begin{enumerate}[leftmargin=*]
    \item Given the referring expression ``\texttt{\{expression\}}'', produce only the mask image: paint the target region \texttt{\{color\}} (\texttt{\{rgb\}}), and leave everything else black ([0,0,0]).

    \item Segment the region referred to by: ``\texttt{\{expression\}}'' Output only a binary segmentation mask with the referred region in \texttt{\{color\}} RGB \texttt{\{rgb\}} and all background pixels black RGB [0,0,0].

    \item Create a binary mask for ``\texttt{\{expression\}}'' Use \texttt{\{color\}} for the referred object or region and black for the background. Output only the mask image.
\end{enumerate}
\end{promptblock}
\noindent All three templates specify a black background; the third names the
foreground color without its RGB value.

\paragraph{Instance segmentation.}
Each prompt begins with \texttt{[TASK: INSTANCE\_SEGMENTATION]}, followed by
one of four base templates (sampled uniformly) and the instance mapping
\textit{$\langle$instance mapping$\rangle$}. The complete prompt is \begin{promptblock}\texttt{[TASK: INSTANCE\_SEGMENTATION]} $\langle$sampled base template$\rangle$ $\langle$instance mapping$\rangle$ \end{promptblock}. The base templates are:
\begin{promptblock}
\begin{enumerate}[leftmargin=*]
    \item This image shows segmentation masks for \texttt{\{category\}} from the input image. The background is set to black color [0, 0, 0]. Each \texttt{\{category\}} instance is represented by a solid mask, and a different color is used for each mask.

    \item This image shows segmentation masks for \texttt{\{category\}} from the input image. The background is set to black color [0, 0, 0]. Each \texttt{\{category\}} is represented by a different color.

    \item Generate an instance segmentation visualization of this image. Each ``\texttt{\{category\}}'' is colored differently. Leave everything else black [0, 0, 0].

    \item This image is a segmentation task derived from the input. The ``\texttt{\{category\}}'' instances are each represented by a unique, solid color on a black [0, 0, 0] background.
\end{enumerate}
\end{promptblock}
\textit{$\langle$instance mapping$\rangle$}:
\begin{promptblock}
Instances are numbered by where their mask first
appears in a left-to-right scan of the image (ties broken top to bottom). The
1st \texttt{\{category\}} is blue, the 2nd is lime, the 3rd is red, the 4th
is sky blue, the 5th is dark green, the 6th is pink, the 7th is apricot, the
8th is mint, the 9th is wine, the 10th is magenta, the 11th is navy, the 12th
is green, the 13th is yellow, the 14th is brown, the 15th is white, the 16th
is olive, the 17th is lavender, the 18th is teal, the 19th is orange, the 20th
is hot pink, the 21st is plum, the 22nd is gray, the 23rd is purple, the 24th
is beige, the 25th is steel blue, the 26th is cyan, the 27th is royal blue,
the 28th is sea green, the 29th is rust, the 30th is mustard, the 31st is
mauve, the 32nd is maroon.
\end{promptblock}
\noindent The order of the 32 colors is the same for every training example and single-pass evaluation, and legend ensembling permutes it across passes.

\paragraph{Image editing, text-to-image generation, and identity.}
Image-editing prompts preserve each Pico-Banana instruction and prepend the
task tag when it is absent:
\begin{promptblock}\texttt{[TASK: IMAGE\_EDITING]} $\langle$Pico-Banana instruction$\rangle$\end{promptblock}
Text-to-image examples use the normalized PD12M caption without a task tag or
fixed wrapper. Identity reconstruction uses a fixed instruction:
\begin{promptblock}\texttt{[TASK: IDENTITY]} Reproduce this image exactly without making any changes.\end{promptblock}

\paragraph{Evaluation prompts.}
COCO-Object evaluation samples the semantic-segmentation templates above in
RGB notation, SA-Co/Gold evaluation uses the instance-segmentation templates
above, and depth and normal evaluation use the training prompts of
\cref{sec:exact-geometry-prompts}. Cityscapes and the two referring benchmarks use dedicated prompts.
Cityscapes lists all 19 classes in a fixed order, with a new random color draw
(\texttt{\{rgb\}}, e.g., \texttt{[220,30,30]}) per image:
\begin{promptblock}
\texttt{[TASK: SEMANTIC\_SEGMENTATION]} Generate a semantic segmentation visualization of the input image, using exactly these 19 Cityscapes classes and this color mapping: \{ 'road': \texttt{\{rgb\}}, 'sidewalk': \texttt{\{rgb\}}, \ldots, 'bicycle': \texttt{\{rgb\}} \}. Each pixel must be colored as exactly one of these semantic classes. Output only the segmentation visualization image.
\end{promptblock}
RefCOCOg and ReasonSeg use the following prompt, where \texttt{\{color\}} is a
sampled color name followed by its RGB value (e.g., \texttt{red [220, 30, 30]})
and, for ReasonSeg, \texttt{\{expression\}} is the original or rewritten query:
\begin{promptblock}
\texttt{[TASK: REFERRING\_SEGMENTATION]} Segment the region that best matches this description: ``\texttt{\{expression\}}''. Paint the matching region \texttt{\{color\}} and leave everything else black [0, 0, 0]. Output only the segmentation mask image.
\end{promptblock}

\subsection{Exact geometry prompts}
\label{sec:exact-geometry-prompts}

Full-scale models use pixel-explicit inverse depth and concise normals for
both training and evaluation.

\paragraph{Depth: concise inverse.}
\begin{promptblock}
Create a dense relative inverse-depth map of the input as neutral
grayscale (R=G=B): closer points are brighter and farther points are darker.
Use one global mapping across the full image, preserve thin structures and
sharp occlusion boundaries, keep smooth surfaces coherent, and return only the
registered depth map.
\end{promptblock}

\paragraph{Depth: pixel-explicit inverse.}
\begin{promptblock}
Estimate a dense relative inverse-depth map for the input image. Produce
exactly one output value for every input pixel at the same pixel location, with
closer points brighter and farther points darker. Encode the result as neutral
grayscale (R=G=B) using one consistent mapping over the complete image.
Preserve object silhouettes, thin structures, occlusion boundaries, smooth
surfaces, and small depth changes. Do not alter the scene or add texture,
shading, color, text, labels, legends, borders, or objects; return only the
registered depth map.
\end{promptblock}

\paragraph{Depth: scene-preserving inverse.}
\begin{promptblock}
Generate the dense depth representation corresponding to the input
image. Return a pixel-aligned neutral grayscale inverse-depth map (R=G=B), with
nearer geometry brighter and farther geometry darker under one image-wide
mapping. Preserve the input composition and fine boundaries, and output only
the depth map without text, labels, legends, or additional objects.
\end{promptblock}

\paragraph{Depth: scene-preserving metric RGB.}
\begin{promptblock}
Generate a dense metric projective camera-depth visualization
corresponding to the input image. Encode camera-z depth in meters with the
declared bounded RGB depth transform, maintaining exact pixel-to-pixel
registration and crisp depth boundaries. Return only the RGB depth
visualization without text, labels, legends, or additional objects.
\end{promptblock}

\paragraph{Normals: concise.}
\begin{promptblock}
Create a dense camera-space unit surface-normal map registered to the
input. Use +x image-right, +y image-up, +z toward the camera and encode (x,y,z)
as R=(1-x)/2, G=(1+y)/2, B=(1+z)/2. Preserve planes, curved surfaces, thin
structures, and sharp orientation boundaries; return only the RGB normal map.
\end{promptblock}

\paragraph{Normals: pixel-explicit.}
\begin{promptblock}
Estimate a dense camera-space unit surface normal for every input pixel
at the same pixel location. Use +x image-right, +y image-up, and +z toward the
camera. Encode physical (x,y,z) as R=(1-x)/2, G=(1+y)/2, B=(1+z)/2. Preserve
planar and curved geometry, thin structures, and sharp orientation
discontinuities. Do not copy image texture, change the scene, or add text,
labels, legends, borders, or objects; return only the registered RGB normal
map.
\end{promptblock}

\paragraph{Normals: scene-preserving.}
\begin{promptblock}
Generate the dense camera-space surface-normal representation
corresponding to the input image. Use +x image-right, +y image-up, +z toward
the camera and encode the unit vector as R=(1-x)/2, G=(1+y)/2, B=(1+z)/2.
Maintain pixel alignment and surface boundaries, and return only the RGB
normal map without text, labels, legends, or additional objects.
\end{promptblock}

\section{Additional Qualitative Results}
\label{sec:qualitative-results}
This appendix collects further qualitative examples for the selected models
and the geometry studies of Appendix~\ref{sec:geometry-ablations}.

\newlength{\ovbgallerycolumnwidth}
\setlength{\ovbgallerycolumnwidth}{\columnwidth}
\onecolumn
\begingroup
\setlength{\parindent}{0pt}
\setlength{\parskip}{0pt}

\newsavebox{\ovbgallerybox}
\newcommand{\ovbgallerystart}{%
  \par\noindent
  \begin{lrbox}{\ovbgallerybox}%
  \begin{minipage}[t]{\linewidth}\vspace{0pt}%
  \centering\captionsetup{type=figure}%
}
\newcommand{\ovbgalleryend}{%
  \end{minipage}\end{lrbox}%
  \typeout{OVB GALLERY figure \thefigure:
    height=\the\dimexpr\ht\ovbgallerybox+\dp\ovbgallerybox\relax,
    width=\the\wd\ovbgallerybox}%
  \usebox{\ovbgallerybox}\par
}
\renewenvironment{figure}[1][]{\ovbgallerystart}{\ovbgalleryend}
\renewenvironment{figure*}[1][]{\ovbgallerystart}{\ovbgalleryend}
\newcommand{\ovbplacegallery}[2]{%
  \par\noindent\makebox[\linewidth][c]{%
    \begin{minipage}[t]{#1}\vspace{0pt}#2\end{minipage}%
  }\par
}

\newsavebox{\ovbgallerypagebox}
\newenvironment{ovbgallerypage}{%
  \par\noindent\begin{lrbox}{\ovbgallerypagebox}%
  \begin{minipage}[t]{\textwidth}\vspace{0pt}%
}{%
  \end{minipage}\end{lrbox}%
  \typeout{OVB GALLERY PAGE: height=\the\dimexpr
    \ht\ovbgallerypagebox+\dp\ovbgallerypagebox\relax,
    limit=\the\textheight}%
  \raisebox{\dp\ovbgallerypagebox}{\usebox{\ovbgallerypagebox}}\par
}

\begin{ovbgallerypage}
\ovbplacegallery{.94\textwidth}{
\begin{figure*}[p]
  \centering
  \scriptsize
  \setlength{\tabcolsep}{1.2pt}
  \begin{tabular}{@{}ccccc@{}}
    Input & Target & No validity mask & Valid pixels only & Eroded by 8 px \\
    \includegraphics[width=.192\textwidth]{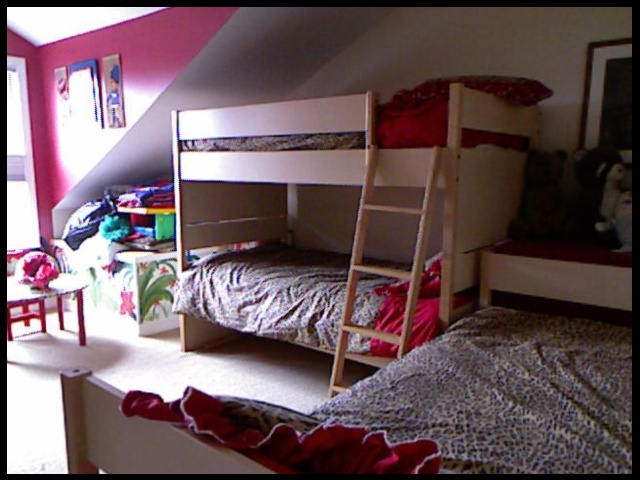} &
    \includegraphics[width=.192\textwidth]{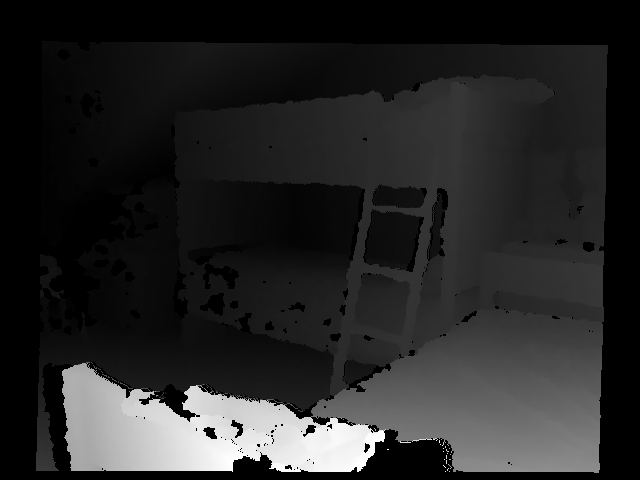} &
    \includegraphics[width=.192\textwidth]{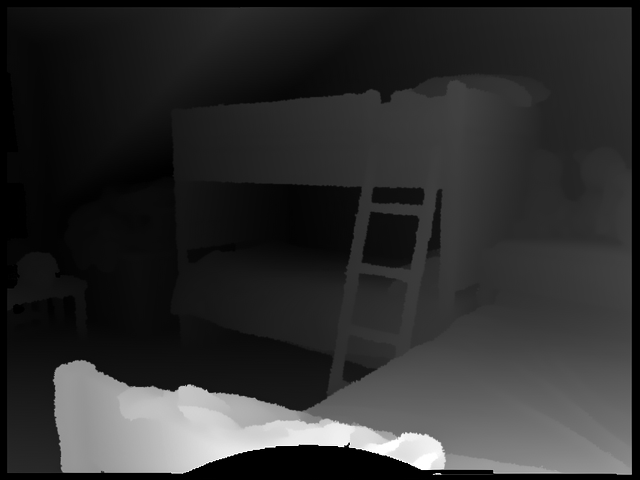} &
    \includegraphics[width=.192\textwidth]{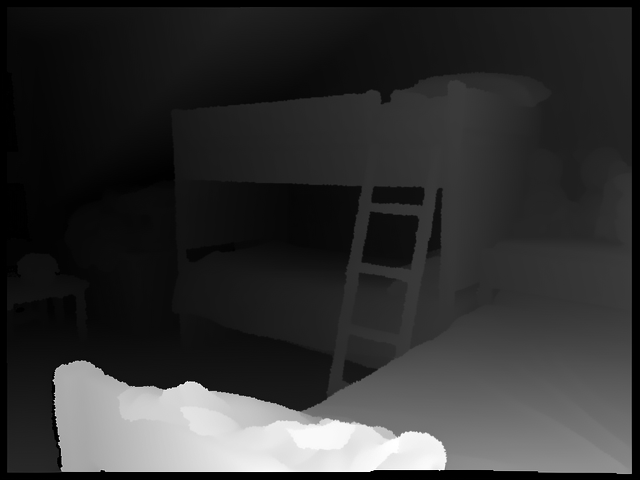} &
    \includegraphics[width=.192\textwidth]{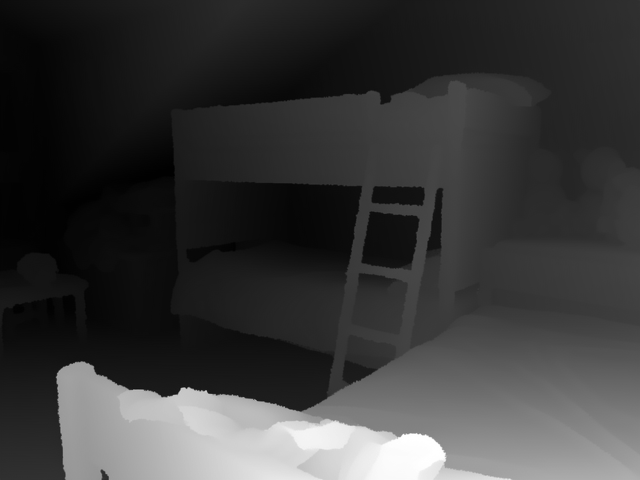} \\[1pt]
    \includegraphics[width=.192\textwidth]{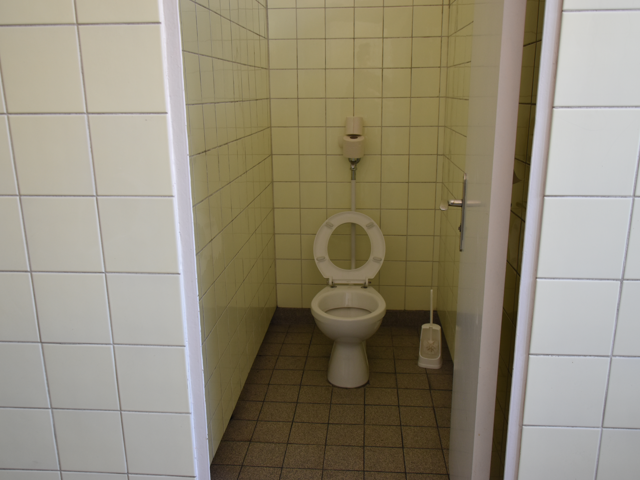} &
    \includegraphics[width=.192\textwidth]{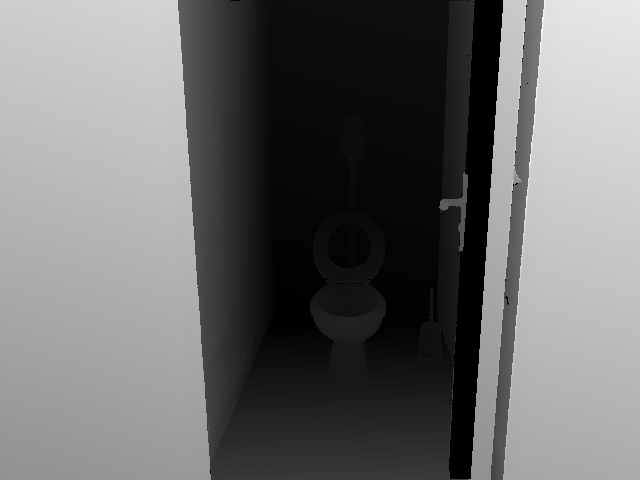} &
    \includegraphics[width=.192\textwidth]{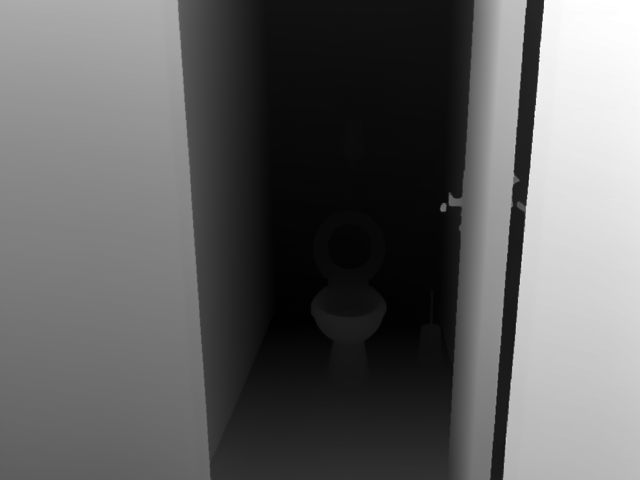} &
    \includegraphics[width=.192\textwidth]{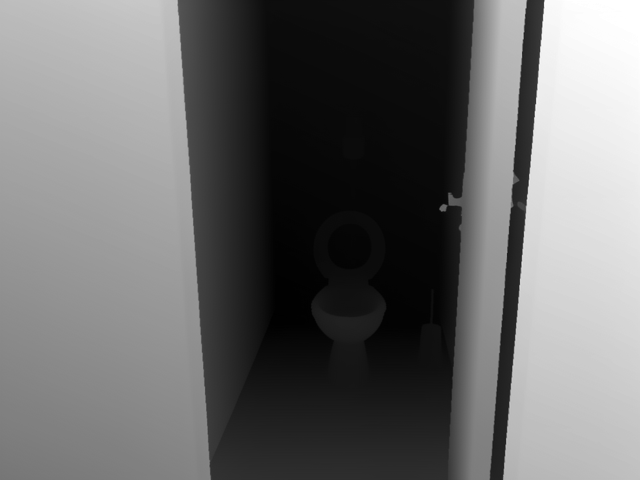} &
    \includegraphics[width=.192\textwidth]{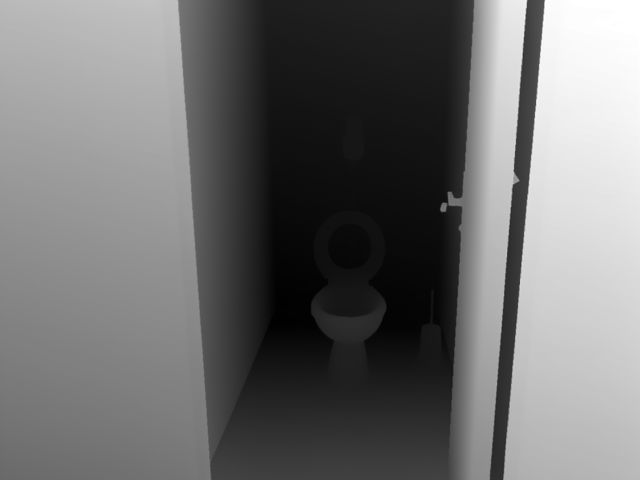} \\[1pt]
    \includegraphics[width=.192\textwidth]{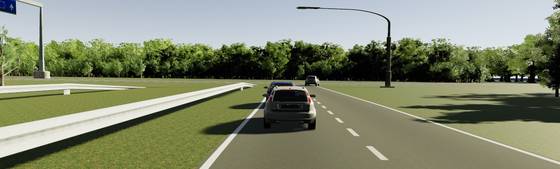} &
    \includegraphics[width=.192\textwidth]{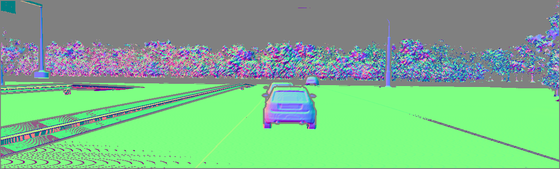} &
    \includegraphics[width=.192\textwidth]{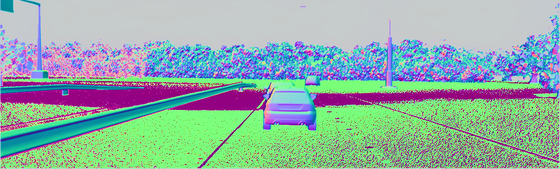} &
    \includegraphics[width=.192\textwidth]{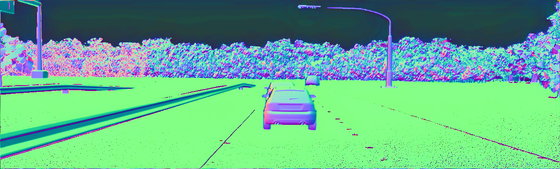} &
    \includegraphics[width=.192\textwidth]{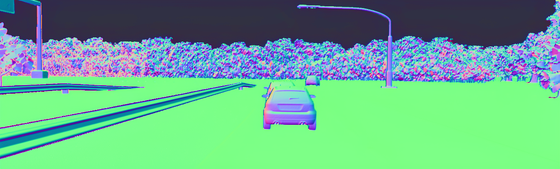} \\[1pt]
    \includegraphics[width=.192\textwidth]{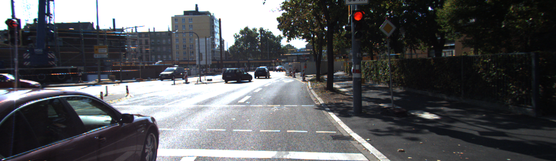} &
    \includegraphics[width=.192\textwidth]{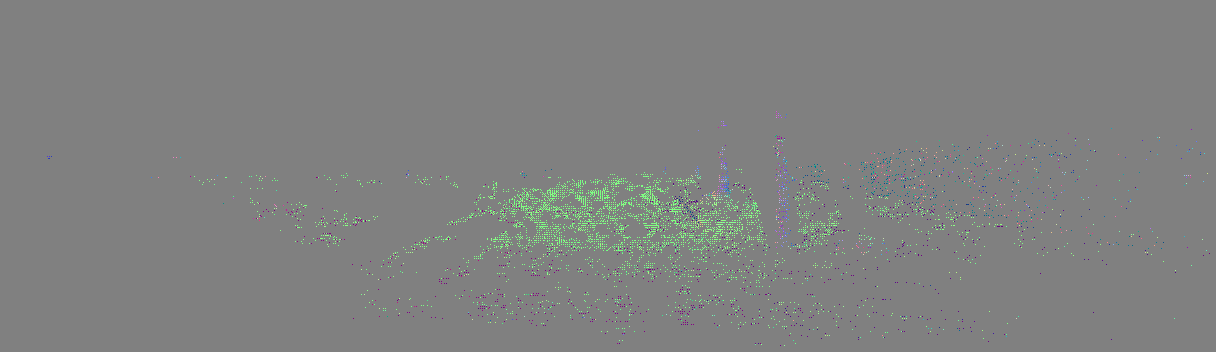} &
    \includegraphics[width=.192\textwidth]{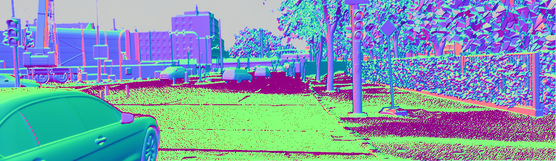} &
    \includegraphics[width=.192\textwidth]{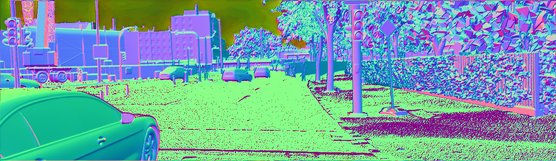} &
    \includegraphics[width=.192\textwidth]{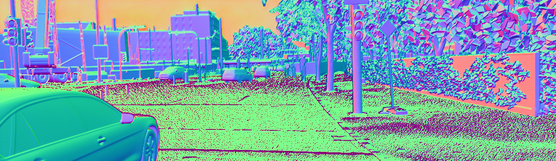}
  \end{tabular}
  \caption{Validity-mask supervision on NYUv2 and iBims-1 depth (top two rows)
  and Virtual KITTI 2 and KITTI normals (bottom two rows). Prediction columns
  differ only in whether training supervises all pixels, valid pixels, or valid
  pixels after removing an eight-pixel boundary band. These are qualitative
  development examples. KITTI normals are derived from depth, not part of
  the final normal benchmark suite.}
  \label{fig:qual-validity-mask}
\end{figure*}
}
\vspace{6pt}
\ovbplacegallery{.84\textwidth}{
\def\ovbeditingtabcolsep{1.2pt}

\begin{figure*}[t]
    \centering
    \providecommand{\ovbeditingtabcolsep}{2pt}
    \setlength{\tabcolsep}{\ovbeditingtabcolsep}
    \begin{tabular}{@{}cccccc@{}}
        \multicolumn{2}{@{}>{\centering\arraybackslash}p{0.32\textwidth}@{}}{\scriptsize\textbf{Add:} Add a group of deer grazing in the snow-covered field.} &
        \multicolumn{2}{@{}>{\centering\arraybackslash}p{0.32\textwidth}@{}}{\scriptsize\textbf{Background:} Change the beach to a mountain landscape.} &
        \multicolumn{2}{@{}>{\centering\arraybackslash}p{0.32\textwidth}@{}}{\scriptsize\textbf{Style:} Transfer the image into a clean graphite pencil sketch.} \\
        \includegraphics[width=0.16\textwidth]{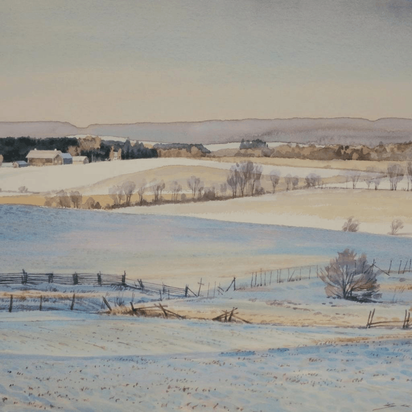} &
        \includegraphics[width=0.16\textwidth]{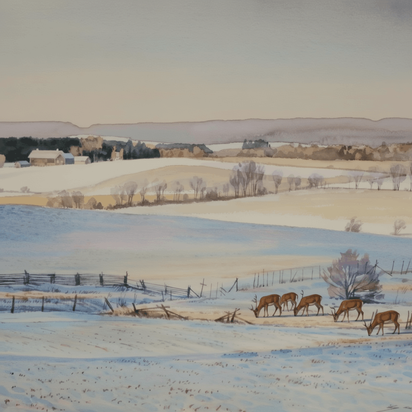} &
        \includegraphics[width=0.16\textwidth]{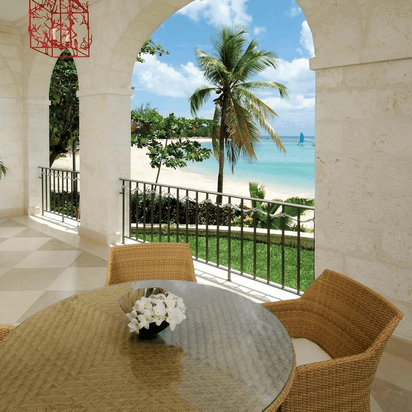} &
        \includegraphics[width=0.16\textwidth]{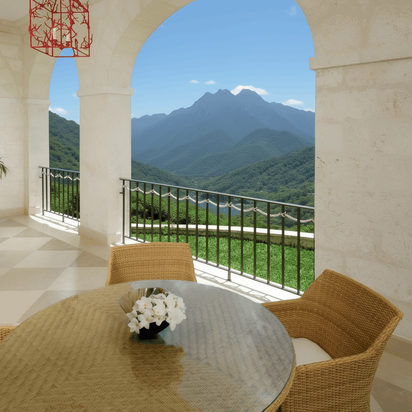} &
        \includegraphics[width=0.16\textwidth]{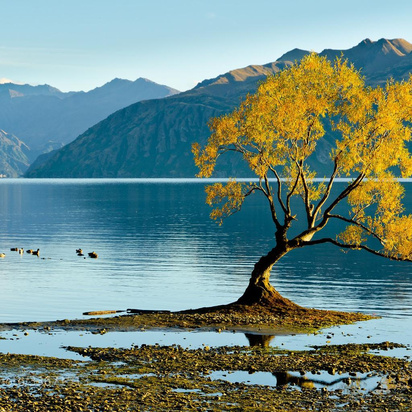} &
        \includegraphics[width=0.16\textwidth]{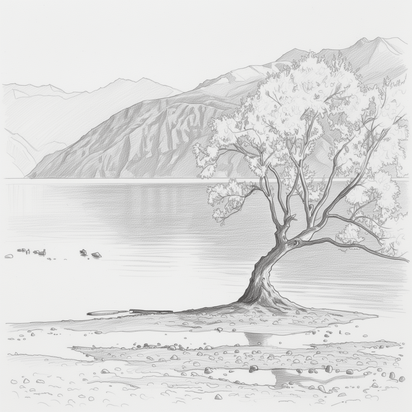} \\[2pt]
        \multicolumn{2}{@{}>{\centering\arraybackslash}p{0.32\textwidth}@{}}{\scriptsize\textbf{Remove:} Remove the green armchair.} &
        \multicolumn{2}{@{}>{\centering\arraybackslash}p{0.32\textwidth}@{}}{\scriptsize\textbf{Replace:} Replace the brown suitcase with a large potted plant.} &
        \multicolumn{2}{@{}>{\centering\arraybackslash}p{0.32\textwidth}@{}}{\scriptsize\textbf{Action:} Make the person smile slightly.} \\
        \includegraphics[width=0.16\textwidth]{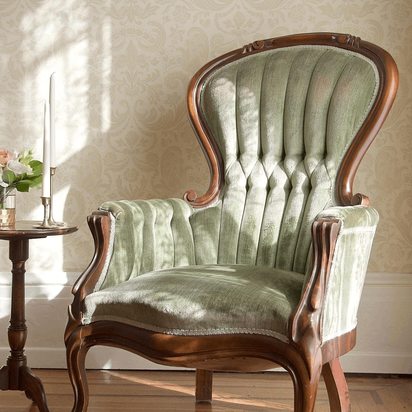} &
        \includegraphics[width=0.16\textwidth]{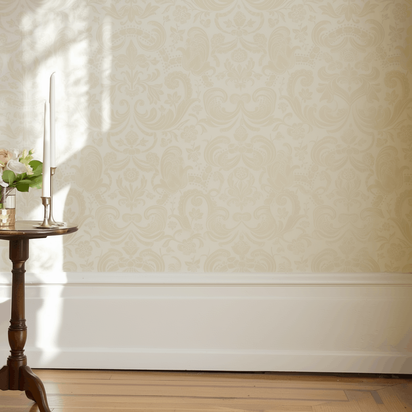} &
        \includegraphics[width=0.16\textwidth]{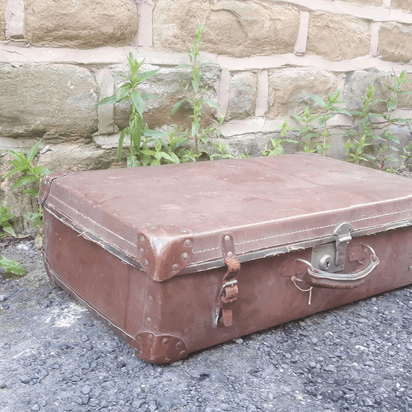} &
        \includegraphics[width=0.16\textwidth]{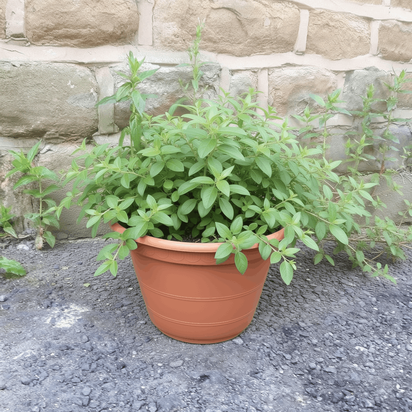} &
        \includegraphics[width=0.16\textwidth]{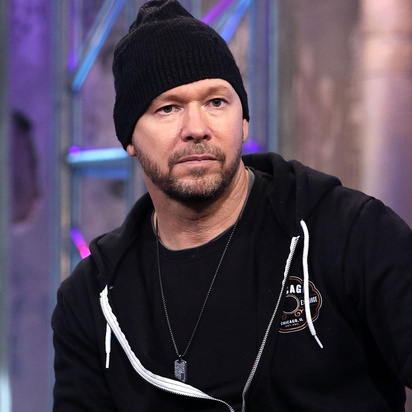} &
        \includegraphics[width=0.16\textwidth]{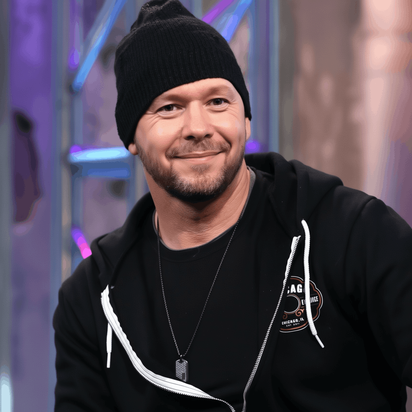} \\
        \noalign{\vspace{5pt}}\hline\noalign{\vspace{5pt}}
        \multicolumn{1}{@{}>{\centering\arraybackslash}b{0.16\textwidth}@{}}{\scriptsize\textbf{Text:} A graffiti mural on a city street depicting a phoenix, with `RISE AGAIN' boldly written.} &
        \multicolumn{1}{@{}>{\centering\arraybackslash}b{0.16\textwidth}@{}}{\scriptsize\textbf{Text:} A coffee mug with `Morning Magic Brew' written in a whimsical font.} &
        \multicolumn{1}{@{}>{\centering\arraybackslash}b{0.16\textwidth}@{}}{\scriptsize\textbf{Counting:} Two dogs playing with one red ball.} &
        \multicolumn{1}{@{}>{\centering\arraybackslash}b{0.16\textwidth}@{}}{\scriptsize\textbf{Counting:} Three flowers on the ground: one red, another yellow, and the third blue.} &
        \multicolumn{1}{@{}>{\centering\arraybackslash}b{0.16\textwidth}@{}}{\scriptsize\textbf{Differentiation:} A large pizza with pepperoni on the left half and mushrooms on the right half.} &
        \multicolumn{1}{@{}>{\centering\arraybackslash}b{0.16\textwidth}@{}}{\scriptsize\textbf{Differentiation:} A scene with two blue balls amidst many yellow ones.} \\
        \includegraphics[width=0.16\textwidth]{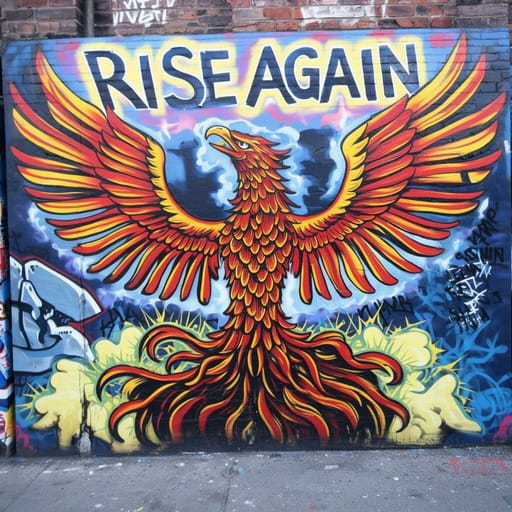} &
        \includegraphics[width=0.16\textwidth]{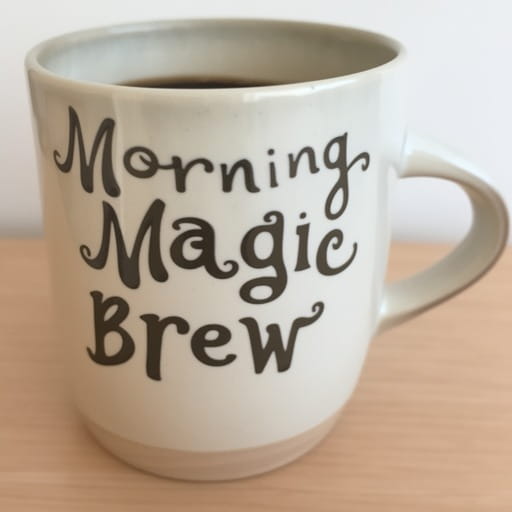} &
        \includegraphics[width=0.16\textwidth]{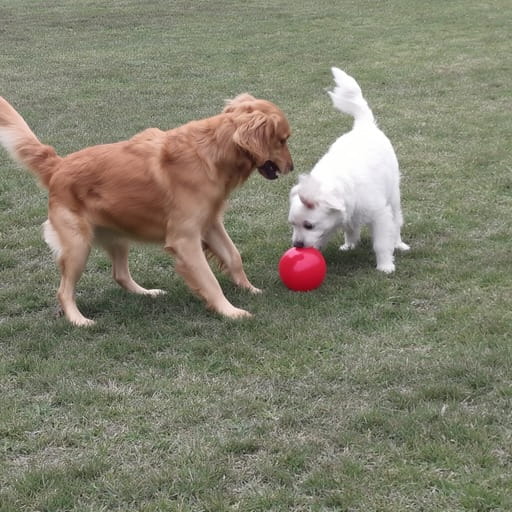} &
        \includegraphics[width=0.16\textwidth]{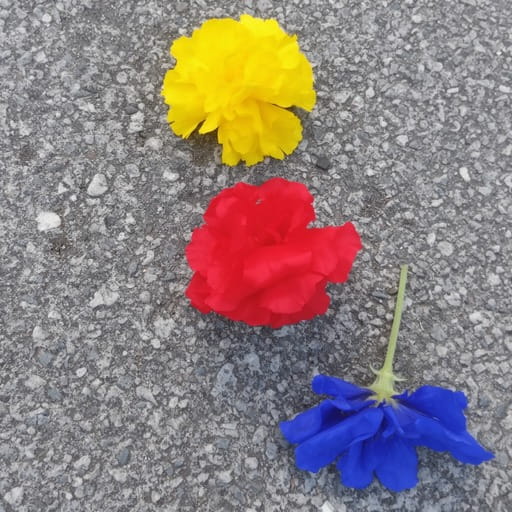} &
        \includegraphics[width=0.16\textwidth]{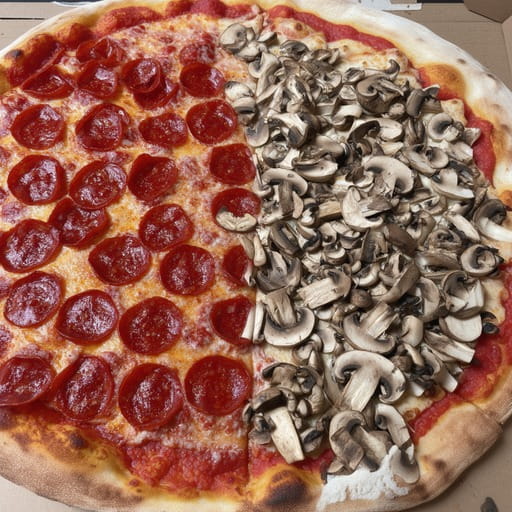} &
        \includegraphics[width=0.16\textwidth]{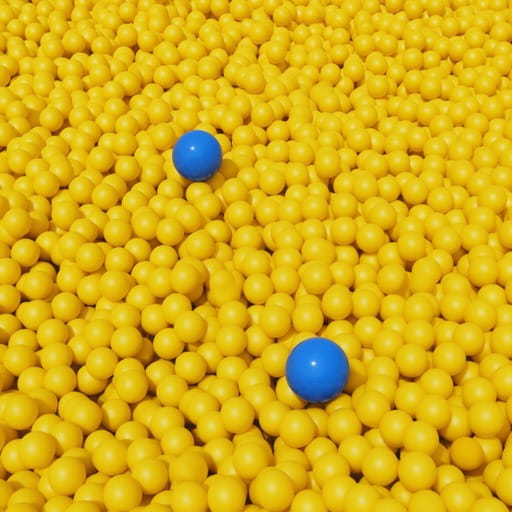}
    \end{tabular}
    \caption{\textbf{Image editing and text-to-image generation after perception tuning.}
    Above the line, image editing on ImgEdit-Bench: each pair shows the input image (left)
    transformed into the \model{} edit (right).
    ImgEdit-Bench has no reference edit; it evaluates instruction compliance and
    visual quality with a VLM judge.
    Below the line, GenAI-Bench samples from \model{} 9B (zero-shot mix) requiring
    rendered text, counting and differentiation.}
    \label{fig:qual-imgedit}
    \label{fig:genai-retention}
\end{figure*}

}
\end{ovbgallerypage}

\clearpage
\begin{ovbgallerypage}
{\def\featurequalitywidth{1.95\ovbgallerycolumnwidth}
 \captionsetup{width=\ovbgallerycolumnwidth}
 \ovbplacegallery{\textwidth}{

\begin{figure}[!htbp]
    \centering
    \captionsetup{skip=6pt}
    \providecommand{\featurequalitywidth}{.95\columnwidth}
    \includegraphics[width=\featurequalitywidth]{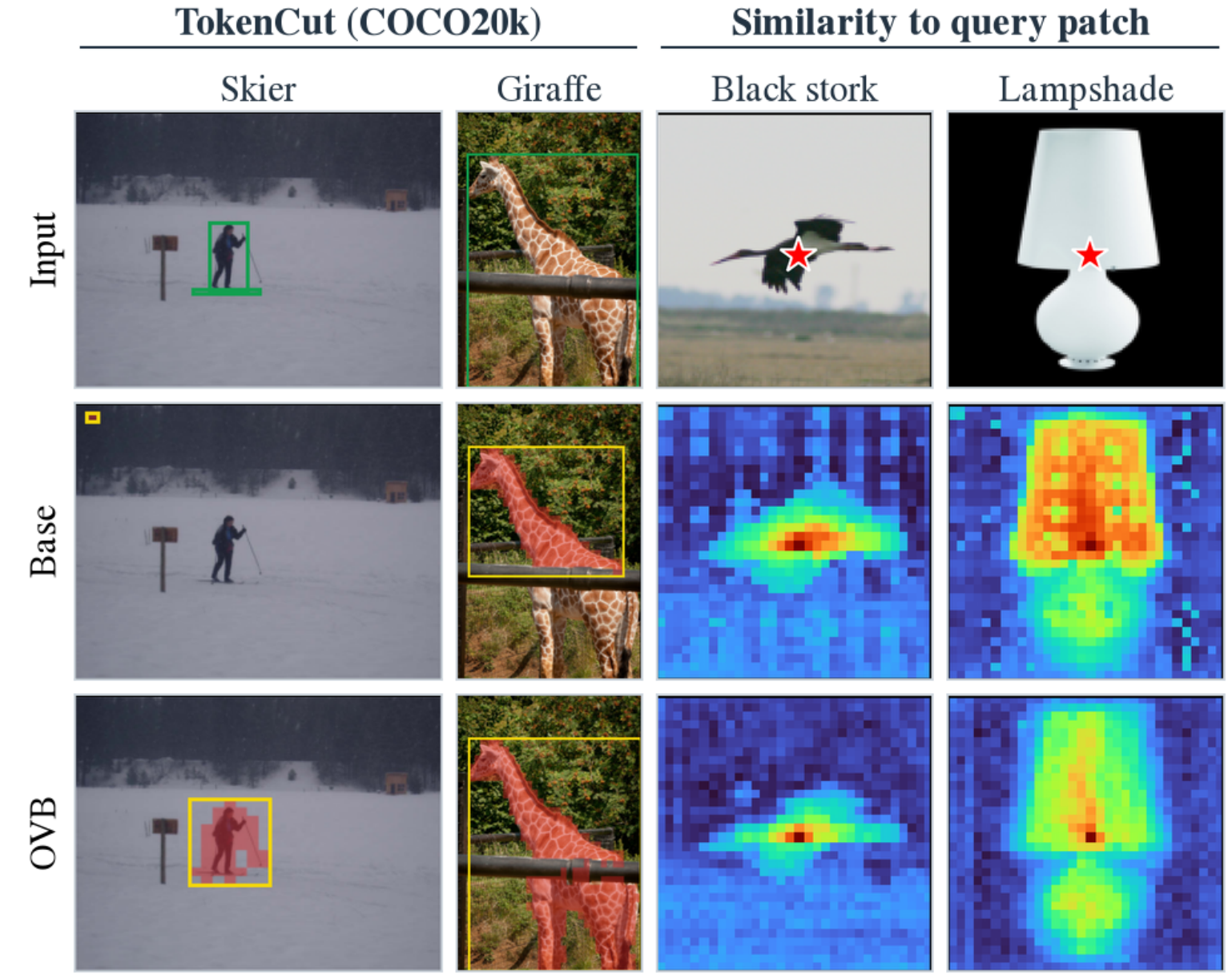}
    \caption{\textbf{Frozen features before and after perception tuning} (\fluxkfourbshort{}). Left: TokenCut object discovery on COCO20k (green: ground truth; yellow box and red mask: discovered object). Right: cosine sim. of the query patch (star) to all patches.}

    \label{fig:feature-quality-qual}
\end{figure}
}}
\vspace{8pt}
\ovbplacegallery{\textwidth}{
\begin{figure*}[p]
  \centering
  \scriptsize
  \setlength{\tabcolsep}{1.2pt}
  \begin{tabular}{@{}ccccccc@{}}
    & & \multicolumn{3}{c}{Inverse-gray prompts} & \multicolumn{2}{c}{Metric RGB} \\
    \cmidrule(lr){3-5}\cmidrule(l){6-7}
    Input & Inv. target & Concise & Pixel-explicit & Scene-preserving & Target & Output \\
    \includegraphics[width=.138\textwidth]{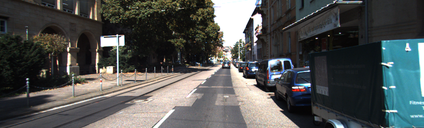} &
    \includegraphics[width=.138\textwidth]{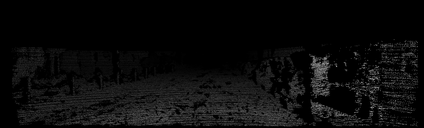} &
    \includegraphics[width=.138\textwidth]{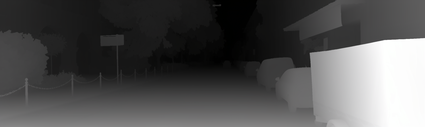} &
    \includegraphics[width=.138\textwidth]{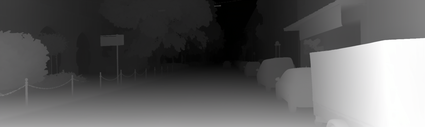} &
    \includegraphics[width=.138\textwidth]{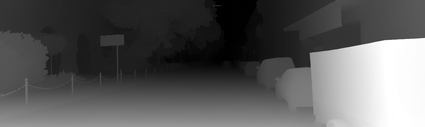} &
    \includegraphics[width=.138\textwidth]{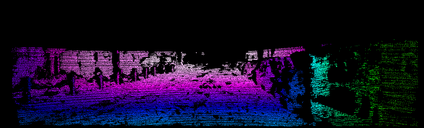} &
    \includegraphics[width=.138\textwidth]{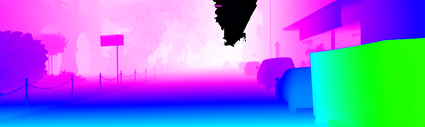} \\[1pt]
    \includegraphics[width=.138\textwidth]{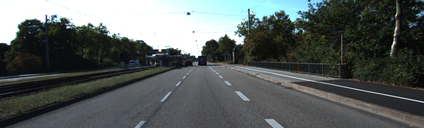} &
    \includegraphics[width=.138\textwidth]{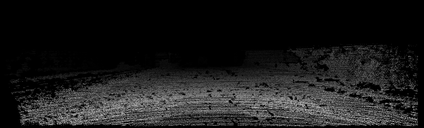} &
    \includegraphics[width=.138\textwidth]{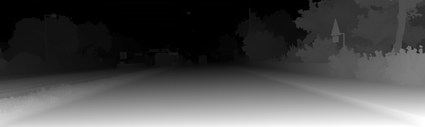} &
    \includegraphics[width=.138\textwidth]{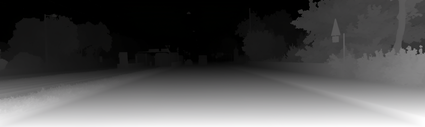} &
    \includegraphics[width=.138\textwidth]{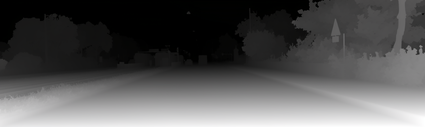} &
    \includegraphics[width=.138\textwidth]{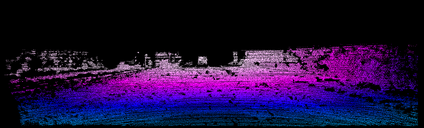} &
    \includegraphics[width=.138\textwidth]{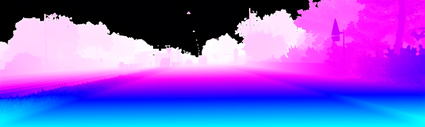} \\[1pt]
    \includegraphics[width=.138\textwidth]{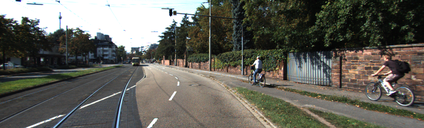} &
    \includegraphics[width=.138\textwidth]{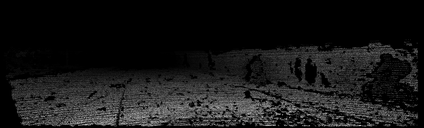} &
    \includegraphics[width=.138\textwidth]{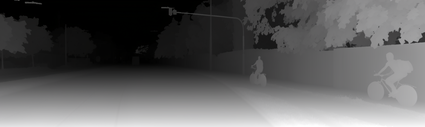} &
    \includegraphics[width=.138\textwidth]{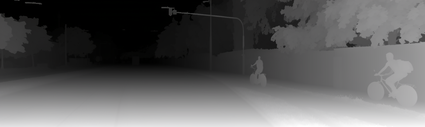} &
    \includegraphics[width=.138\textwidth]{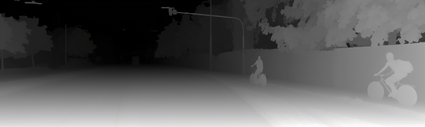} &
    \includegraphics[width=.138\textwidth]{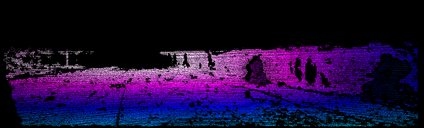} &
    \includegraphics[width=.138\textwidth]{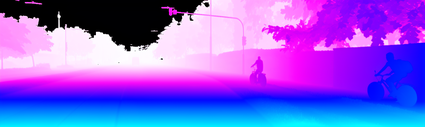} \\[1pt]
    \includegraphics[width=.138\textwidth]{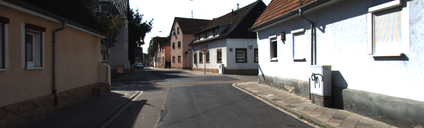} &
    \includegraphics[width=.138\textwidth]{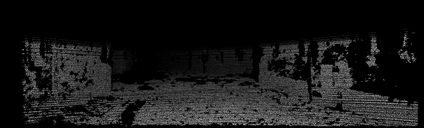} &
    \includegraphics[width=.138\textwidth]{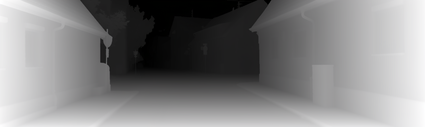} &
    \includegraphics[width=.138\textwidth]{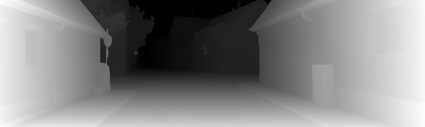} &
    \includegraphics[width=.138\textwidth]{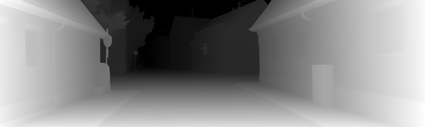} &
    \includegraphics[width=.138\textwidth]{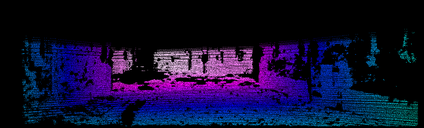} &
    \includegraphics[width=.138\textwidth]{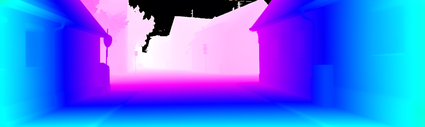}
  \end{tabular}
  \caption{Depth prompt and representation variants on four KITTI Eigen
  evaluation examples after task-isolated Klein fine-tuning. The three inverse-gray columns
  share a target encoding and differ in instruction style. Metric-RGB colors encode
  depth along a fixed RGB path and cannot be compared directly with grayscale
  intensity. Quantitative results use the encoding-specific alignment in
  Sec.~\ref{sec:metrics}.}
  \label{fig:qual-depth-representations}
  \label{fig:qual-depth-metric-rgb}
\end{figure*}
}
\end{ovbgallerypage}

\clearpage
\begin{ovbgallerypage}
\noindent
\begin{minipage}[t]{\ovbgallerycolumnwidth}\vspace{0pt}

\begin{figure}[t]
\centering
\captionsetup{skip=6pt}
\renewcommand{\arraystretch}{1}
\setlength{\tabcolsep}{2pt}

\setlength{\colw}{\dimexpr(\columnwidth-2\tabcolsep)/2\relax}

\newcommand{\qimg}[1]{\includegraphics[width=\colw]{#1}}

\begin{tabular}{@{}
  >{\centering\arraybackslash}p{\colw}
  >{\centering\arraybackslash}p{\colw}
@{}}

\textbf{Input} & \textbf{Prediction} \\[1pt]
\qimg{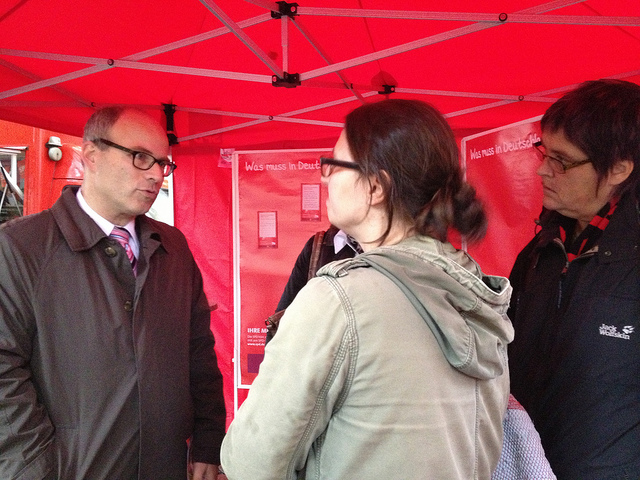} &
\qimg{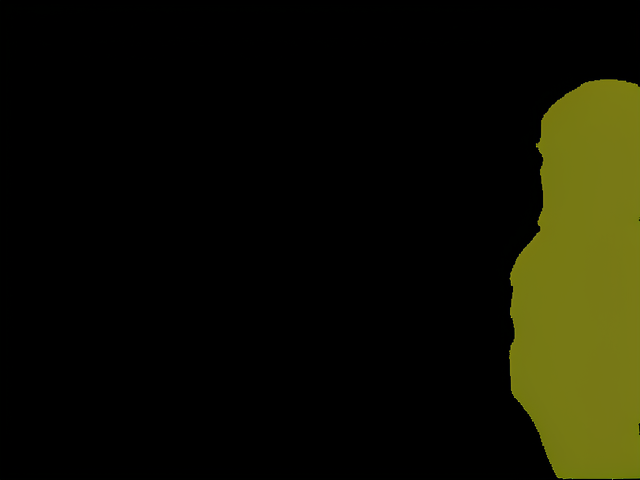} \\[2pt]
\multicolumn{2}{@{}>{\centering\arraybackslash}p{\columnwidth}@{}}{\prompttext{Woman in dark blue jacket with red and black scarf.}} \\[2pt]

\textbf{Input} & \textbf{Prediction} \\[1pt]
\qimg{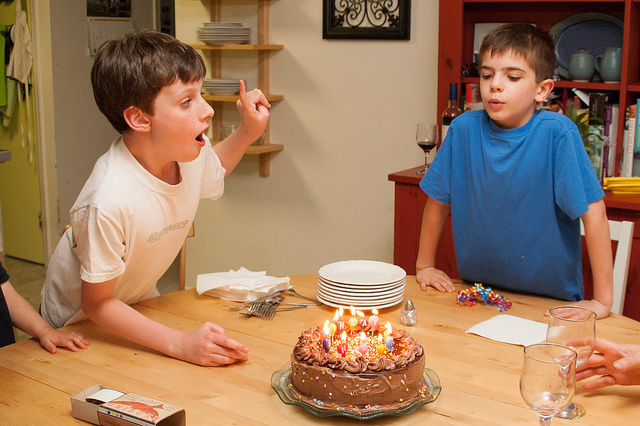} &
\qimg{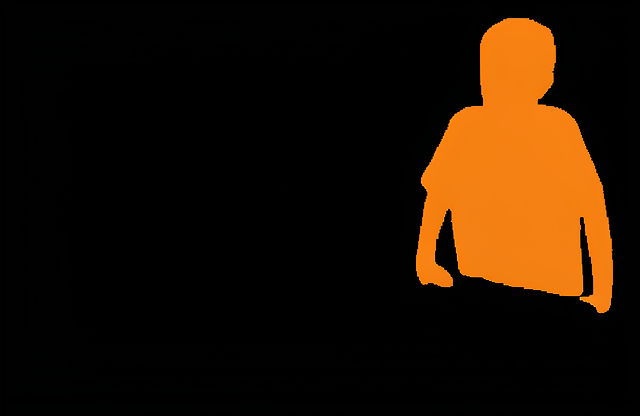} \\[2pt]
\multicolumn{2}{@{}>{\centering\arraybackslash}p{\columnwidth}@{}}{\prompttext{A boy in a blue shirt about to blow out his candles.}}

\end{tabular}
\caption{\textbf{Additional general referring-segmentation results on RefCOCOg with zero-shot \model{} 9B.}
Each example shows only the input image and the raw \model{} prediction;
the referring expression appears immediately below the pair.}
\label{fig:qual-refcocog-general}
\end{figure}

  \vspace{6pt}

\begin{figure}[t]
\centering
\captionsetup{skip=6pt}
\begin{adjustbox}{width=.66\columnwidth,center}
\begin{minipage}{\columnwidth}
\centering
\renewcommand{\arraystretch}{1}
\setlength{\tabcolsep}{2pt}
\newlength{\nzboundaryw}
\setlength{\nzboundaryw}{\dimexpr(\columnwidth-4\tabcolsep)/3\relax}
\newcommand{\nzboundaryimg}[1]{\includegraphics[width=\nzboundaryw]{#1}}
\begin{tabular}{@{}
  >{\centering\arraybackslash}p{\nzboundaryw}
  >{\centering\arraybackslash}p{\nzboundaryw}
  >{\centering\arraybackslash}p{\nzboundaryw}
@{}}

\textbf{Input} & \textbf{Original label} & \textbf{Prediction} \\[0pt]
\nzboundaryimg{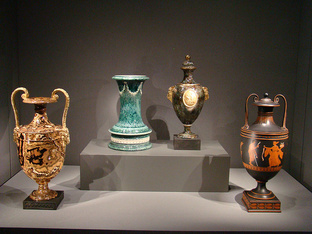} &
\nzboundaryimg{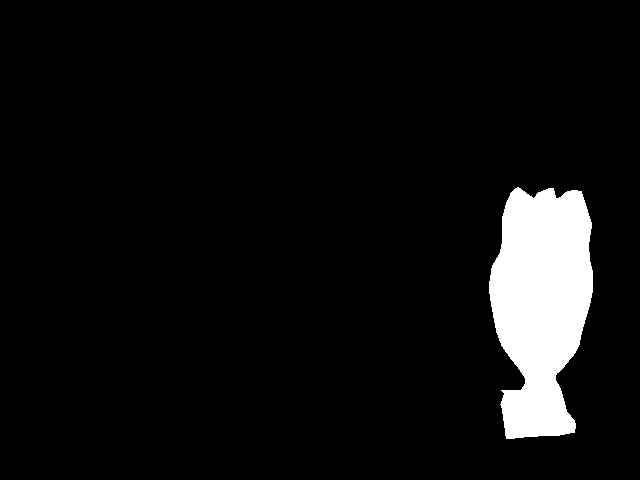} &
\nzboundaryimg{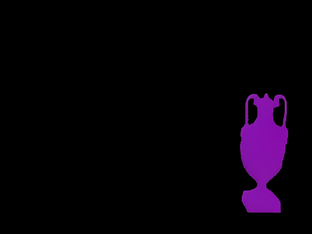} \\[0pt]
\multicolumn{3}{@{}>{\centering\arraybackslash}p{\columnwidth}@{}}{\prompttext{A black ancient Greek vase with figures.}} \\[1pt]
\nzboundaryimg{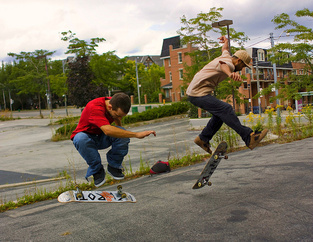} &
\nzboundaryimg{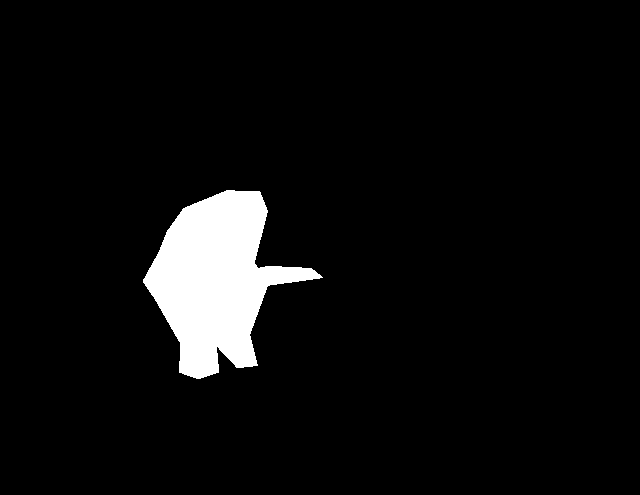} &
\nzboundaryimg{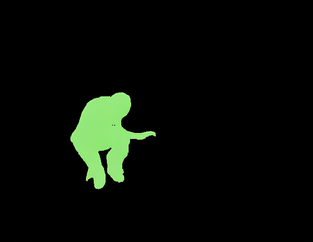} \\[0pt]
\multicolumn{3}{@{}>{\centering\arraybackslash}p{\columnwidth}@{}}{\prompttext{A man in a red shirt performing a kick flip.}} \\
\arrayrulecolor{black!25}\specialrule{0.4pt}{2pt}{2pt}\arrayrulecolor{black}
\nzboundaryimg{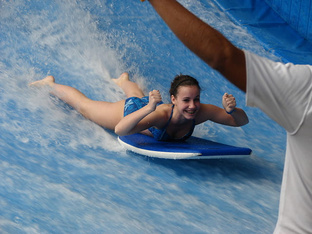} &
\nzboundaryimg{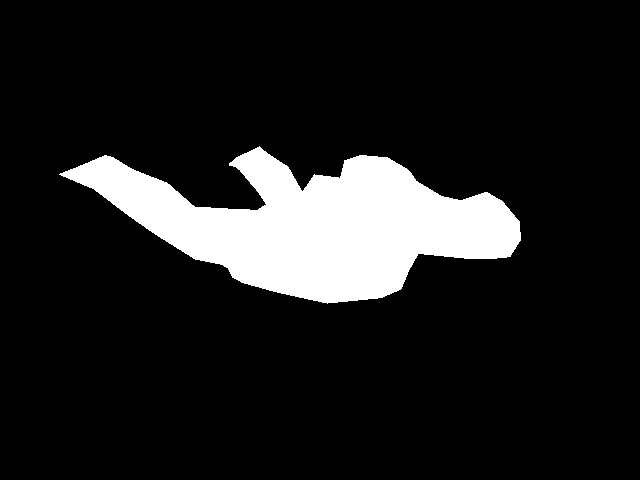} &
\nzboundaryimg{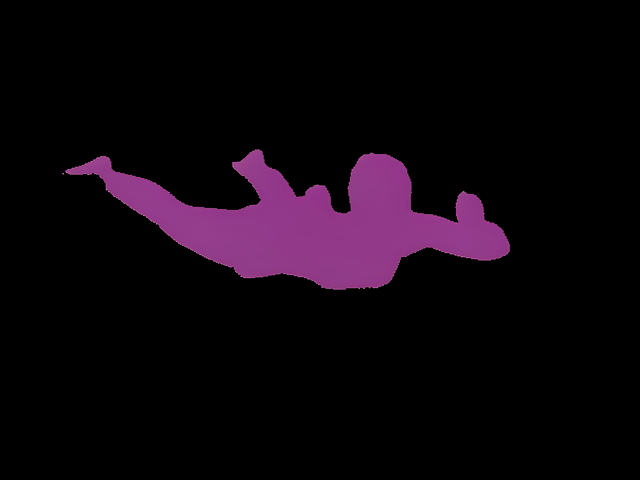} \\[0pt]
\multicolumn{3}{@{}>{\centering\arraybackslash}p{\columnwidth}@{}}{\prompttext{woman body surfing.}} \\[1pt]

\nzboundaryimg{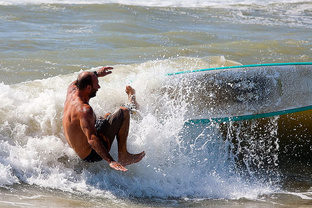} &
\nzboundaryimg{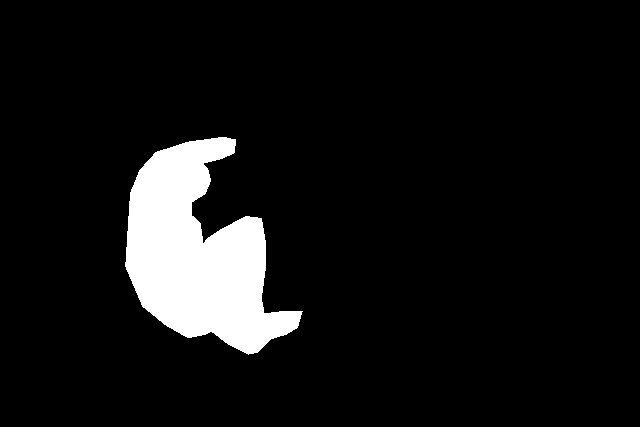} &
\nzboundaryimg{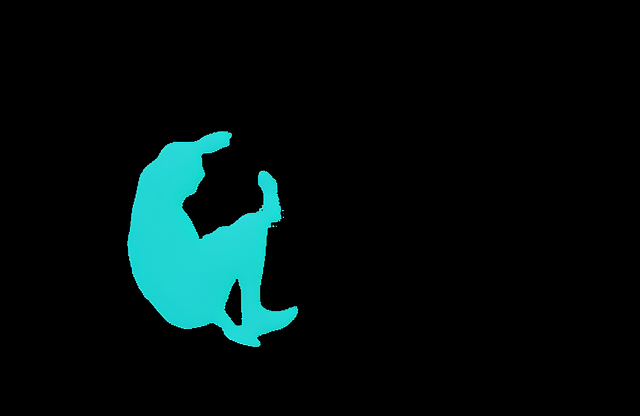} \\[0pt]
\multicolumn{3}{@{}>{\centering\arraybackslash}p{\columnwidth}@{}}{\prompttext{a man falling off his surfboard}} \\[1pt]

\nzboundaryimg{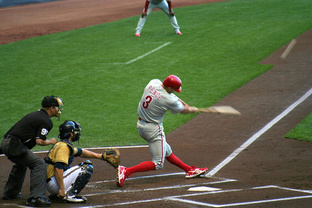} &
\nzboundaryimg{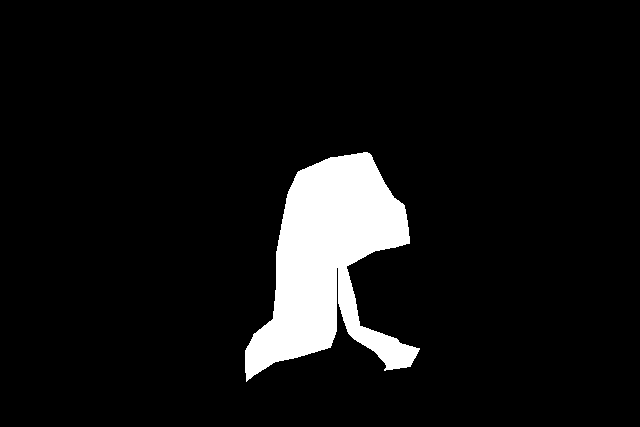} &
\nzboundaryimg{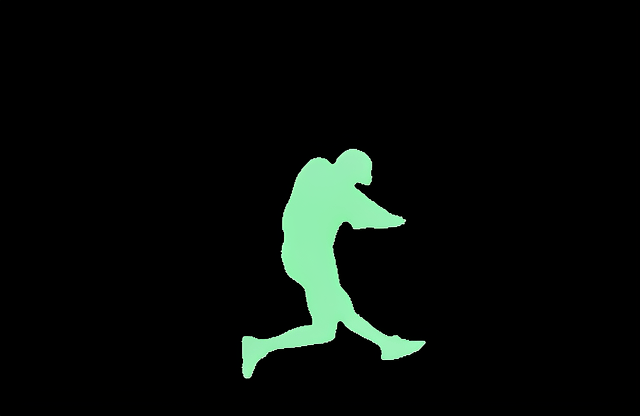} \\[0pt]
\multicolumn{3}{@{}>{\centering\arraybackslash}p{\columnwidth}@{}}{\prompttext{a baseball player with number 3 on his uniform.}} \\[1pt]

\nzboundaryimg{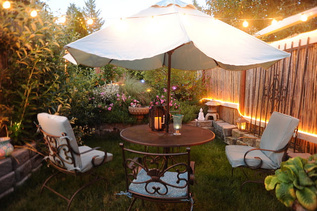} &
\nzboundaryimg{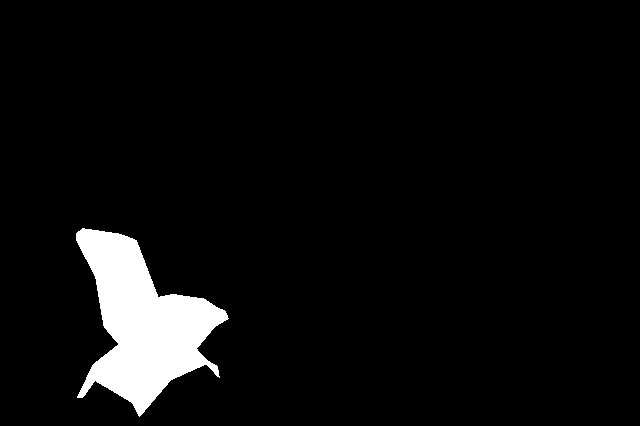} &
\nzboundaryimg{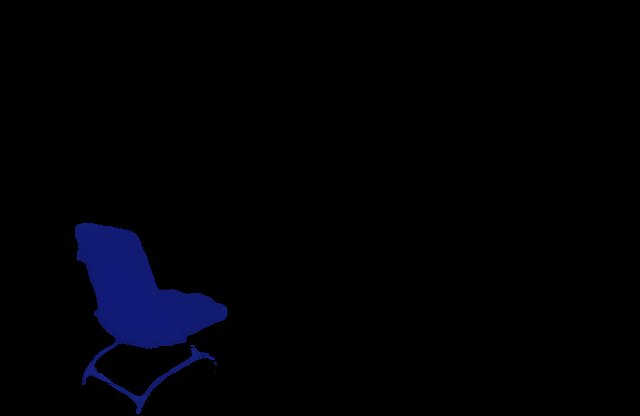} \\[0pt]
\multicolumn{3}{@{}>{\centering\arraybackslash}p{\columnwidth}@{}}{\prompttext{Chair with the white cushion on the left.}}

\end{tabular}
\end{minipage}
\end{adjustbox}
\caption{\textbf{Segmentation boundary refinement beyond the original RefCOCOg label with \model{} 9B.}
Above the line, the zero-shot model and below it, the non-zero-shot model. The predicted mask follows the visible referred region beyond the coarse binary annotation.}
\label{fig:supp-qual-refcocog-boundaries-nonzero}%
\label{fig:qual-refcocog-boundaries}
\end{figure}

\end{minipage}\hfill
\begin{minipage}[t]{\ovbgallerycolumnwidth}\vspace{0pt}

\begin{figure}[!tp]
\centering
\renewcommand{\arraystretch}{1}
\setlength{\tabcolsep}{2pt}

\setlength{\rscolwtwo}{0pt} %
\setlength{\rscolwtwo}{\dimexpr(\linewidth-2\tabcolsep)/2\relax}

\newcommand{\rsov}[1]{\raisebox{-0.5\height}{\includegraphics[width=\rscolwtwo]{#1}}}
\newcommand{\rsr}[1]{%
  \promptbox{\parbox[c]{0.85\rscolwtwo}{\raggedright\scriptsize
    \textcolor{promptaccent}{\textbf{Query:}}\ \itshape #1}}}
\newcommand{\rsq}[2]{%
  \promptbox{\parbox[c]{0.85\rscolwtwo}{\raggedright\scriptsize
    \textcolor{promptaccent}{\textbf{Original:}}\ {\itshape #1}\\[2pt]
    \textcolor{promptaccent}{\textbf{Transformed:}}\ {\itshape #2}}}}

\begin{tabular}{@{}
  >{\centering\arraybackslash}p{\rscolwtwo}
  >{\centering\arraybackslash}p{\rscolwtwo}
@{}}

\rsov{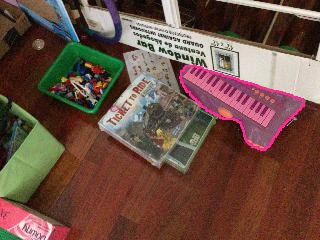} &
\rsr{What musical instrument in the picture is usually played with both hands on a keyboard?} \\
\noalign{\vspace{3pt}}
\rsov{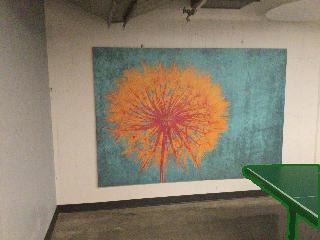} &
\rsr{If you want to play table tennis indoors, what furniture in the picture should be used as the playing surface?} \\
\noalign{\vspace{3pt}}
\rsov{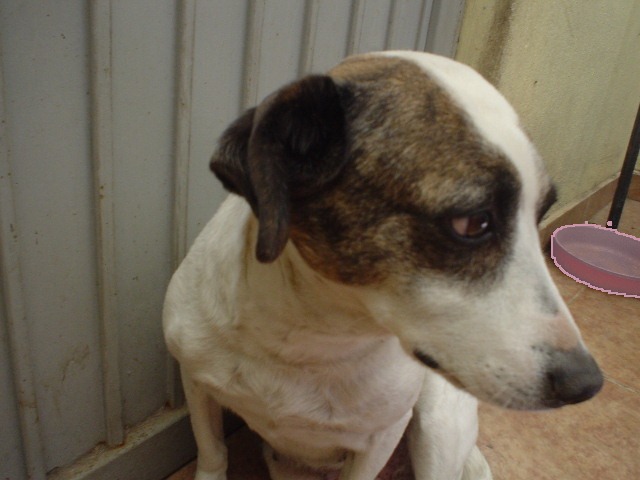} &
\rsr{something that the dog's food should be put into} \\
\noalign{\vspace{5pt}}\hline\noalign{\vspace{5pt}}

\rsov{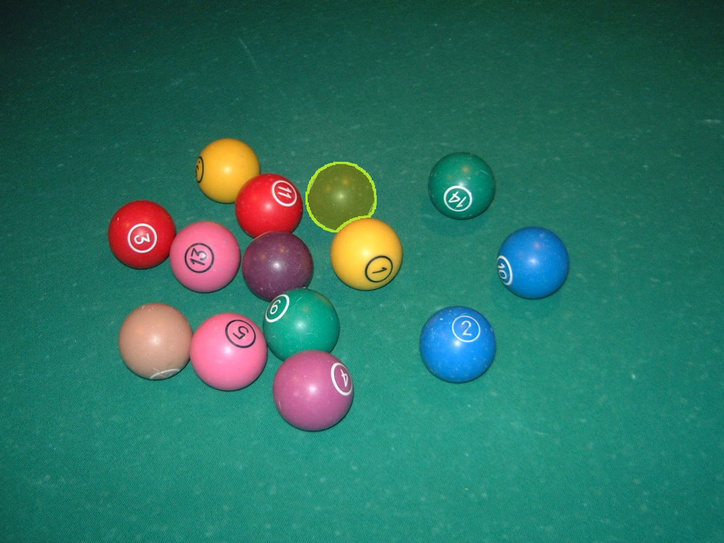} &
\rsq{the ball that can only be hit into the hole at last}{the black billiard ball} \\
\noalign{\vspace{3pt}}
\rsov{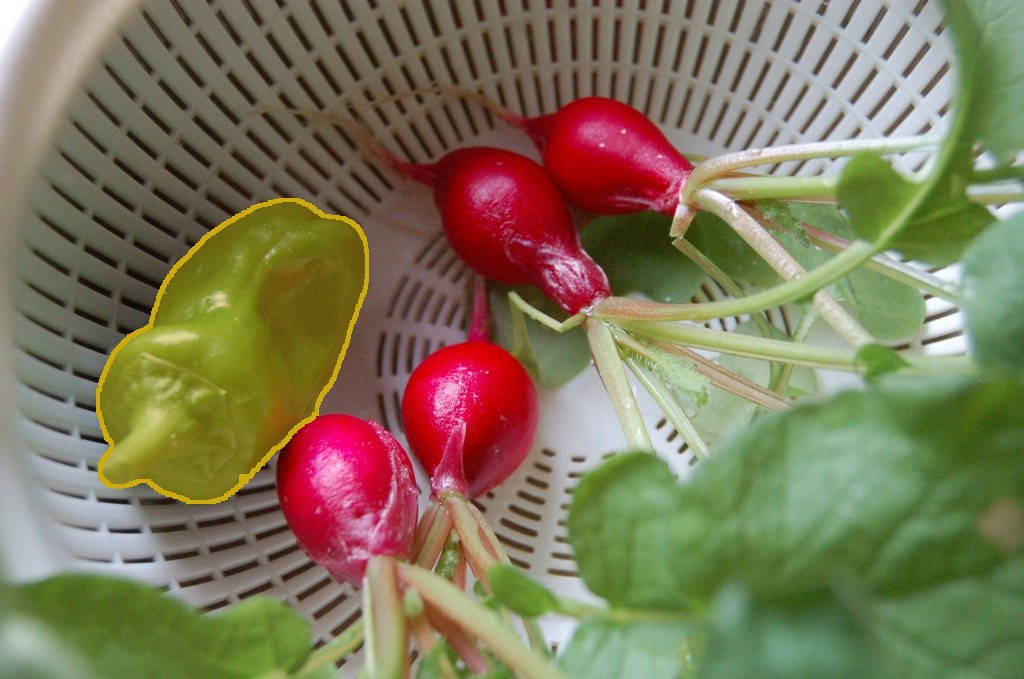} &
\rsq{the food that makes people feel spicy or hot}{the green pepper} \\
\noalign{\vspace{3pt}}
\rsov{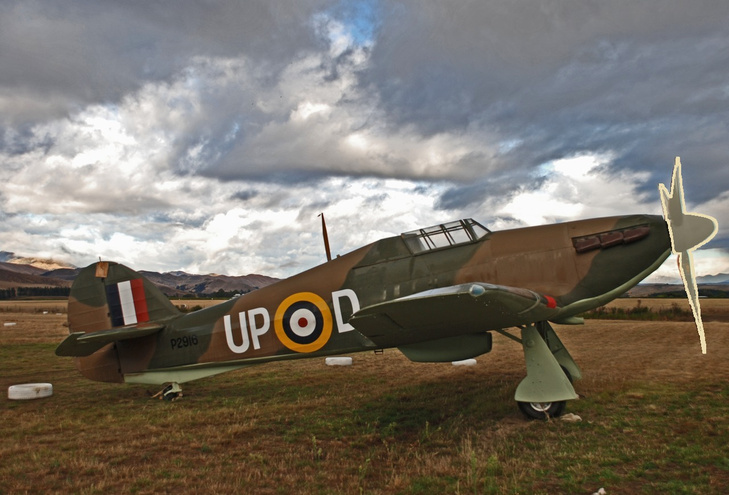} &
\rsq{If we were to take a flight in the sky, what part of the plane in the picture would be spinning to generate lift and propel the aircraft forward?}{the propeller at the front of the plane} \\

\end{tabular}
\caption{\textbf{Reasoning segmentation on ReasonSeg with the zero-shot \model{} 9B model.}
Above the line, the raw query is used; below it, Gemma4 31B rewrites each query
(\textcolor{promptaccent}{\textbf{Original}}) into an explicit expression
(\textcolor{promptaccent}{\textbf{Transformed}}). Predicted
masks are overlaid on the input. Expressions fill the prompt template of
Sec.~\ref{sec:prompt-templates}.}
\label{fig:qual-reasonseg-supp}
\label{fig:qual-reasonseg-part1}
\end{figure}

\end{minipage}\par
\end{ovbgallerypage}

\clearpage
\begin{ovbgallerypage}
\ovbplacegallery{\textwidth}{
\begin{figure*}[t]
\centering
\begin{adjustbox}{max totalsize={\textwidth}{0.89\textheight},center}
\begin{minipage}{\textwidth}
\setlength{\tabcolsep}{0pt}
\begin{minipage}[t]{\dimexpr0.4850\textwidth+1pt\relax}
  \centering
  \includegraphics[width=\dimexpr0.5\linewidth-0.5pt\relax]{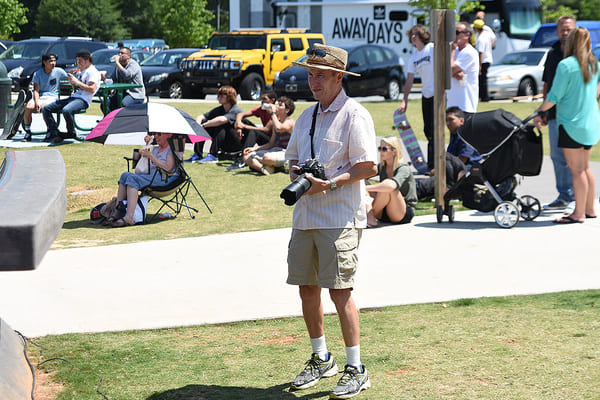}\kern1pt%
  \includegraphics[width=\dimexpr0.5\linewidth-0.5pt\relax]{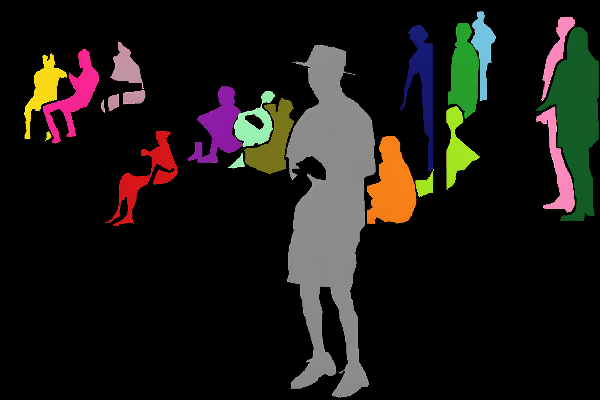}

  \vspace{1pt}\raggedright\scriptsize
  \texttt{[TASK: INSTANCE\_SEGMENTATION]} This image is a segmentation task derived from the input. The ``people'' instances are each represented by a unique, solid color on a black [0, 0, 0] background. \textit{$\langle$instance mapping$\rangle$}
\end{minipage}\hfill
\begin{minipage}[t]{\dimexpr0.4850\textwidth+1pt\relax}
  \centering
  \includegraphics[width=\dimexpr0.5\linewidth-0.5pt\relax]{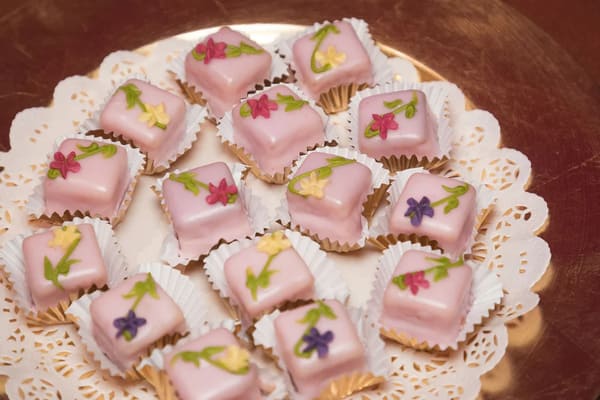}\kern1pt%
  \includegraphics[width=\dimexpr0.5\linewidth-0.5pt\relax]{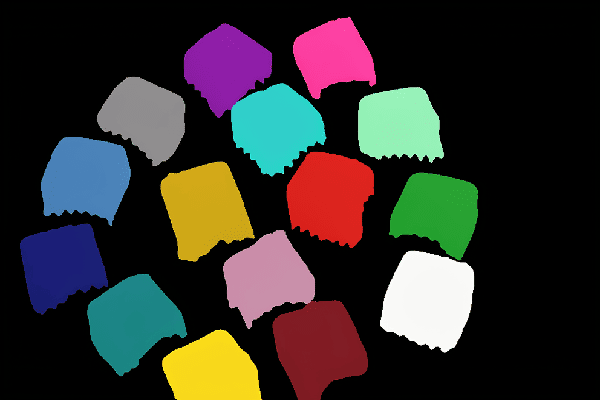}

  \vspace{1pt}\raggedright\scriptsize
  \texttt{[TASK: INSTANCE\_SEGMENTATION]} This image shows segmentation masks for petit four from the input image. The background is set to black color [0, 0, 0]. Each petit four is represented by a different color. \textit{$\langle$instance mapping$\rangle$}
\end{minipage}

\vspace{4pt}

\begin{minipage}[t]{\dimexpr0.4970\textwidth+1pt\relax}
  \centering
  \includegraphics[width=\dimexpr0.5\linewidth-0.5pt\relax]{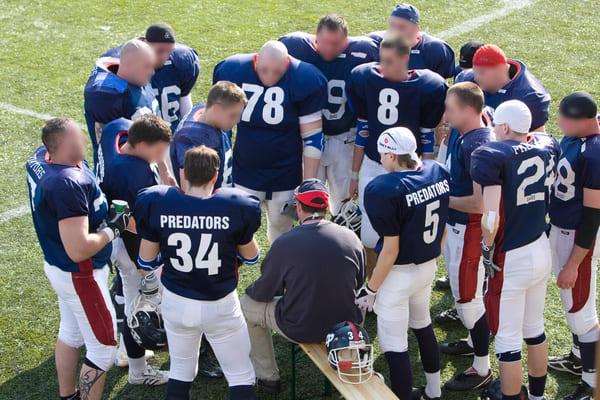}\kern1pt%
  \includegraphics[width=\dimexpr0.5\linewidth-0.5pt\relax]{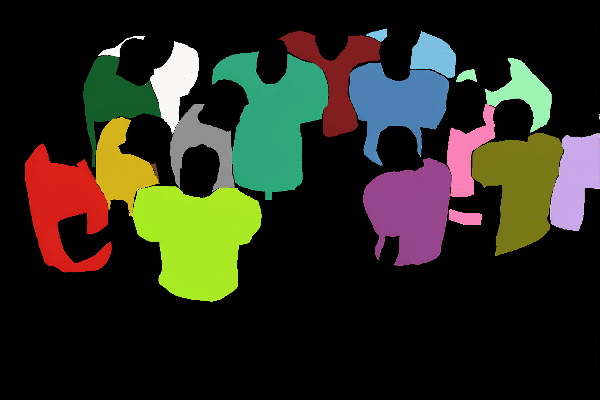}

  \vspace{1pt}\raggedright\scriptsize
  \texttt{[TASK: INSTANCE\_SEGMENTATION]} Generate an instance segmentation visualization of this image. Each ``dark blue shirt'' is colored differently. Leave everything else black [0, 0, 0]. \textit{$\langle$instance mapping$\rangle$}
\end{minipage}\hfill
\begin{minipage}[t]{\dimexpr0.4730\textwidth+1pt\relax}
  \centering
  \includegraphics[width=\dimexpr0.5\linewidth-0.5pt\relax]{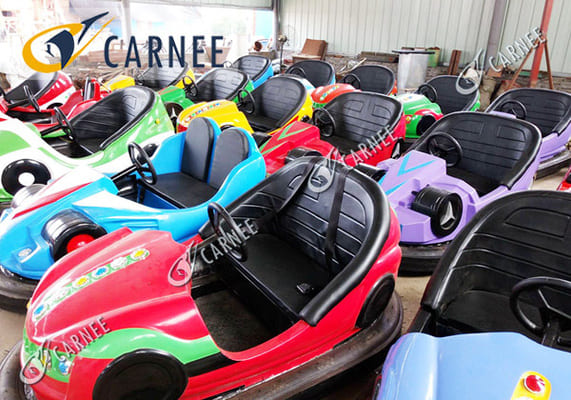}\kern1pt%
  \includegraphics[width=\dimexpr0.5\linewidth-0.5pt\relax]{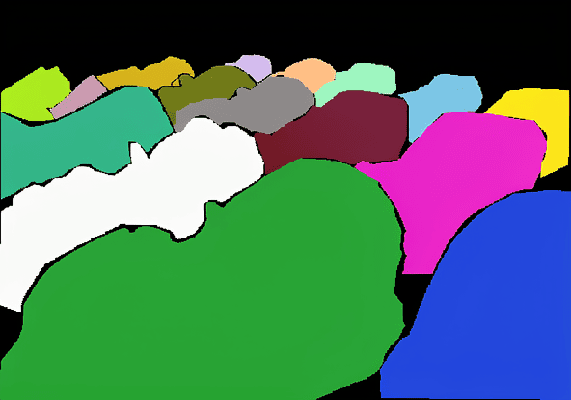}

  \vspace{1pt}\raggedright\scriptsize
  \texttt{[TASK: INSTANCE\_SEGMENTATION]} This image shows segmentation masks for bumper car from the input image. The background is set to black color [0, 0, 0]. Each bumper car instance is represented by a solid mask, and a different color is used for each mask. \textit{$\langle$instance mapping$\rangle$}
\end{minipage}

\vspace{4pt}

\begin{minipage}[t]{\dimexpr0.4126\textwidth+1pt\relax}
  \centering
  \includegraphics[width=\dimexpr0.5\linewidth-0.5pt\relax]{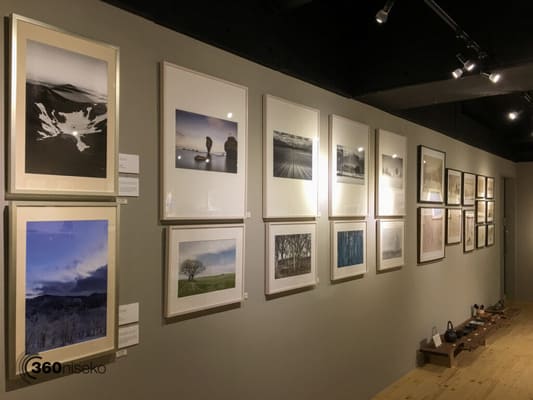}\kern1pt%
  \includegraphics[width=\dimexpr0.5\linewidth-0.5pt\relax]{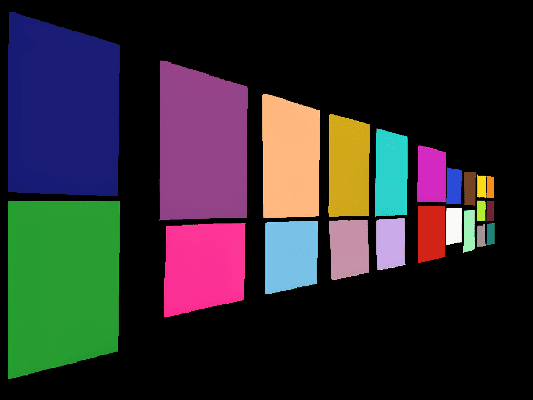}

  \vspace{1pt}\raggedright\scriptsize
  \texttt{[TASK: INSTANCE\_SEGMENTATION]} This image is a segmentation task derived from the input. The ``photograph'' instances are each represented by a unique, solid color on a black [0, 0, 0] background. \textit{$\langle$instance mapping$\rangle$}
\end{minipage}\hfill
\begin{minipage}[t]{\dimexpr0.5574\textwidth+1pt\relax}
  \centering
  \includegraphics[width=\dimexpr0.5\linewidth-0.5pt\relax]{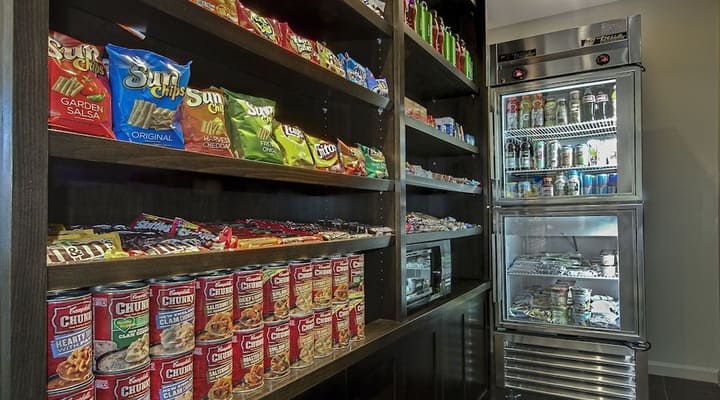}\kern1pt%
  \includegraphics[width=\dimexpr0.5\linewidth-0.5pt\relax]{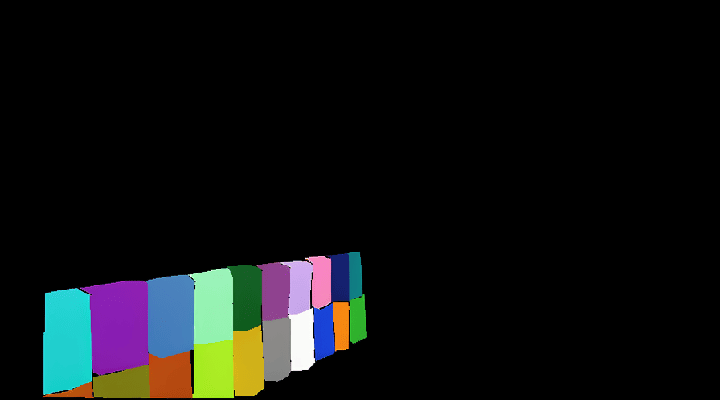}

  \vspace{1pt}\raggedright\scriptsize
  \texttt{[TASK: INSTANCE\_SEGMENTATION]} This image shows segmentation masks for a canned good from the input image. The background is set to black color [0, 0, 0]. Each a canned good instance is represented by a solid mask, and a different color is used for each mask. \textit{$\langle$instance mapping$\rangle$}
\end{minipage}

\vspace{4pt}

\begin{minipage}[t]{\dimexpr0.4159\textwidth+1pt\relax}
  \centering
  \includegraphics[width=\dimexpr0.5\linewidth-0.5pt\relax]{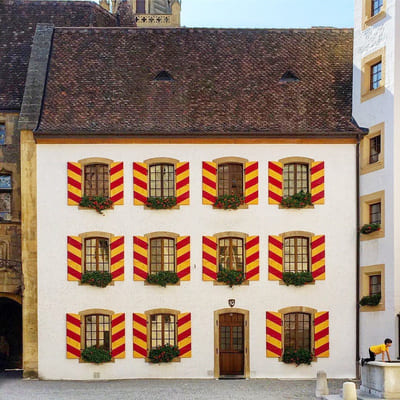}\kern1pt%
  \includegraphics[width=\dimexpr0.5\linewidth-0.5pt\relax]{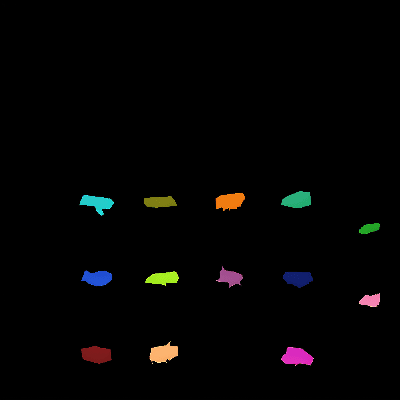}

  \vspace{1pt}\raggedright\scriptsize
  \texttt{[TASK: INSTANCE\_SEGMENTATION]} This image shows segmentation masks for greenery from the input image. The background is set to black color [0, 0, 0]. Each greenery instance is represented by a solid mask, and a different color is used for each mask. \textit{$\langle$instance mapping$\rangle$}
\end{minipage}\hfill
\begin{minipage}[t]{\dimexpr0.5541\textwidth+1pt\relax}
  \centering
  \includegraphics[width=\dimexpr0.5\linewidth-0.5pt\relax]{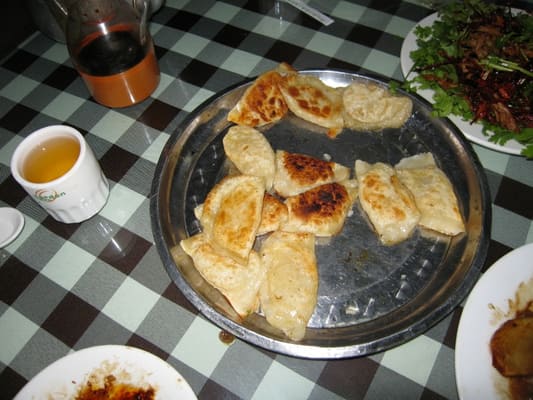}\kern1pt%
  \includegraphics[width=\dimexpr0.5\linewidth-0.5pt\relax]{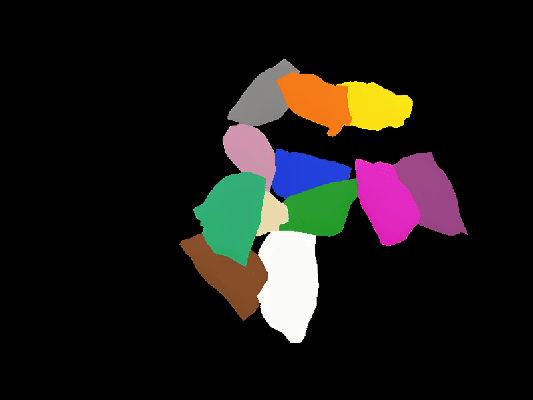}

  \vspace{1pt}\raggedright\scriptsize
  \texttt{[TASK: INSTANCE\_SEGMENTATION]} Generate an instance segmentation visualization of this image. Each ``Shui jiao'' is colored differently. Leave everything else black [0, 0, 0]. \textit{$\langle$instance mapping$\rangle$}
\end{minipage}

\vspace{4pt}

\begin{minipage}[t]{0.485\textwidth}
  \centering
  \includegraphics[width=0.497\linewidth]{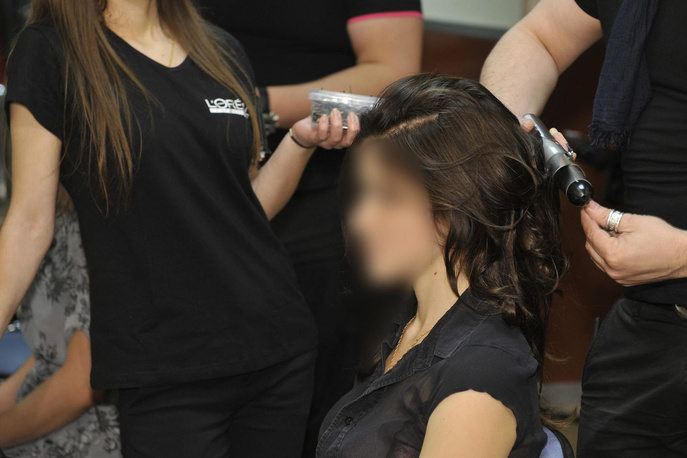}\hspace{1pt}%
  \includegraphics[width=0.497\linewidth]{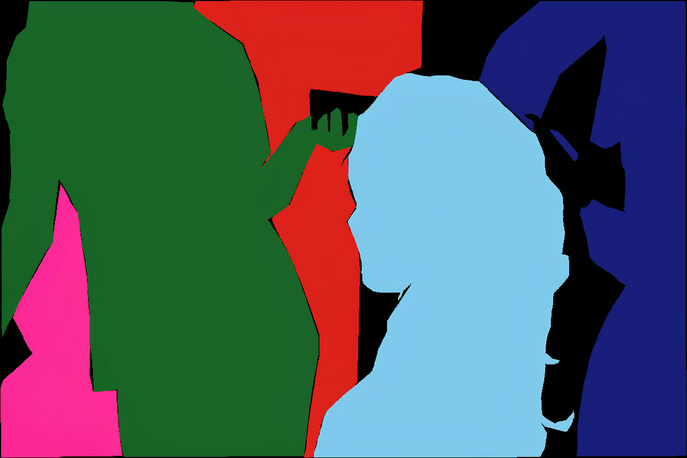}

  \vspace{1pt}\raggedright\scriptsize
  \texttt{[TASK: INSTANCE\_SEGMENTATION]} This image shows segmentation masks for the person from the input image. The background is set to black color [0, 0, 0]. Each the person instance is represented by a solid mask, and a different color is used for each mask. \textit{$\langle$instance mapping$\rangle$}
\end{minipage}\hfill
\begin{minipage}[t]{0.485\textwidth}
  \centering
  \includegraphics[width=0.497\linewidth]{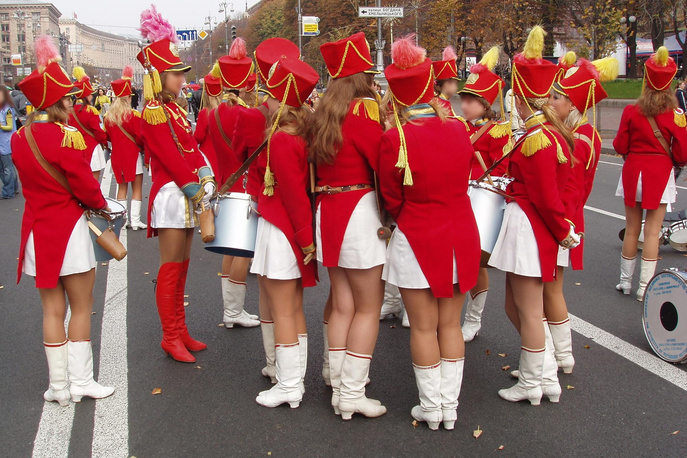}\hspace{1pt}%
  \includegraphics[width=0.497\linewidth]{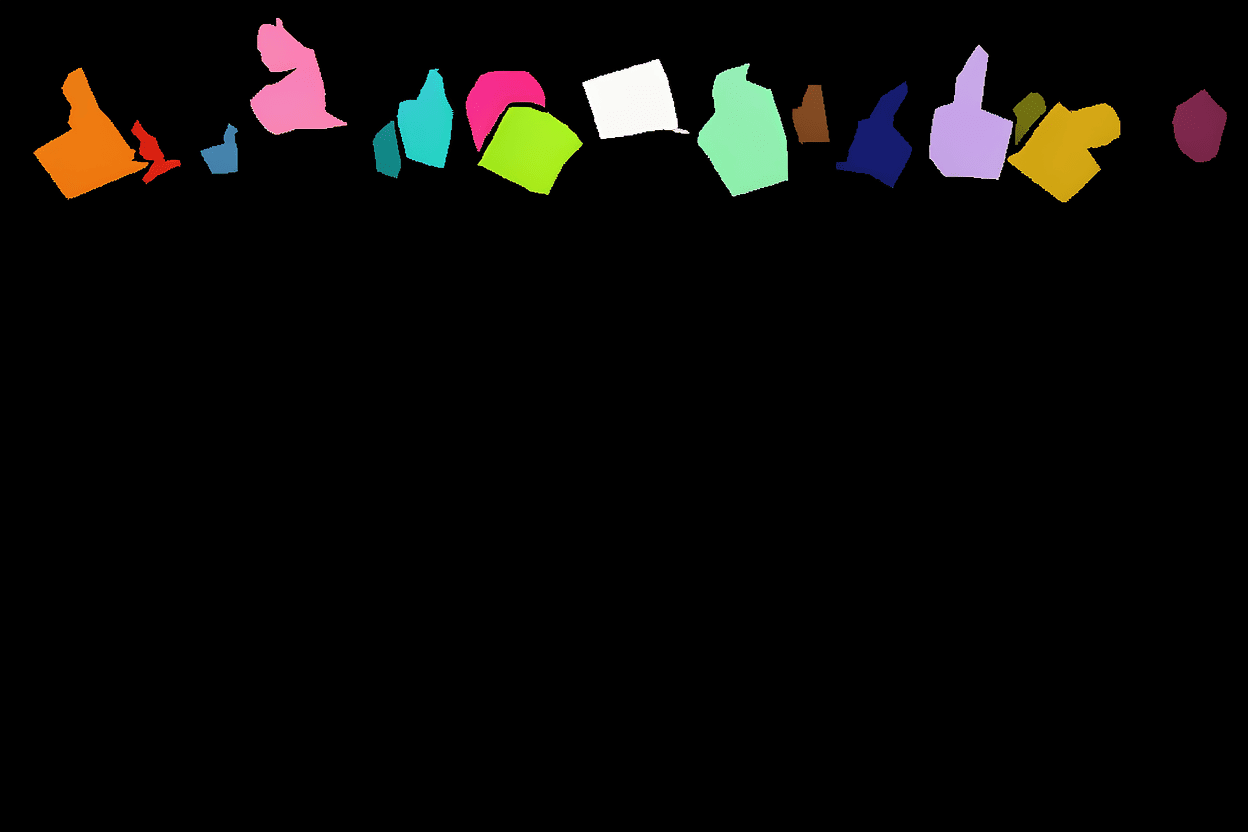}

  \vspace{1pt}\raggedright\scriptsize
  \texttt{[TASK: INSTANCE\_SEGMENTATION]} This image shows segmentation masks for a red hat from the input image. The background is set to black color [0, 0, 0]. Each a red hat is represented by a different color. \textit{$\langle$instance mapping$\rangle$}
\end{minipage}

\end{minipage}
\end{adjustbox}
\caption{\textbf{Additional instance segmentation results on SA-Co/Gold} with the zero-shot \model{} 9B model. Each pair shows the input and its raw generated prediction from a single generation (no multi-pass consensus). As in Fig.~\ref{fig:instanceseg-qualitative-mergedmodel}, the instruction template is drawn per image from a fixed pool of four paraphrases, and every prompt ends with the instance--color mapping, abbreviated as \textit{$\langle$instance mapping$\rangle$}.}
\label{fig:instanceseg-qualitative-supp}
\end{figure*}

}
\end{ovbgallerypage}

\clearpage
\begin{ovbgallerypage}
\ovbplacegallery{\textwidth}{

\begin{figure*}[!htbp]
\centering
\renewcommand{\arraystretch}{1}
\setlength{\tabcolsep}{1.5pt}

\ifdefined\supppairw\else\newlength{\supppairw}\fi
\ifdefined\scolw\else\newlength{\scolw}\fi
\setlength{\gap}{6pt}
\setlength{\scolw}{\dimexpr(\textwidth-\gap-4\tabcolsep)/4\relax}
\setlength{\supppairw}{\dimexpr2\scolw+2\tabcolsep\relax}

\newcommand{\simg}[1]{\includegraphics[width=\scolw]{#1}}

\begin{tabular}{@{}
  >{\centering\arraybackslash}p{\scolw}
  >{\centering\arraybackslash}p{\scolw}
  @{\hspace{\gap}}
  >{\centering\arraybackslash}p{\scolw}
  >{\centering\arraybackslash}p{\scolw}
@{}}

\textbf{Input} & \textbf{Prediction} & \textbf{Input} & \textbf{Prediction} \\[2pt]
\simg{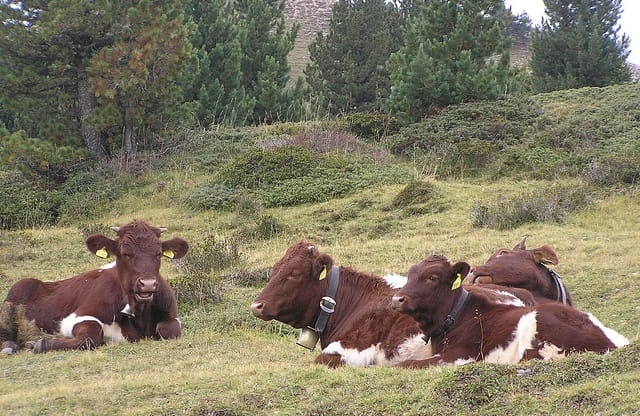} &
\simg{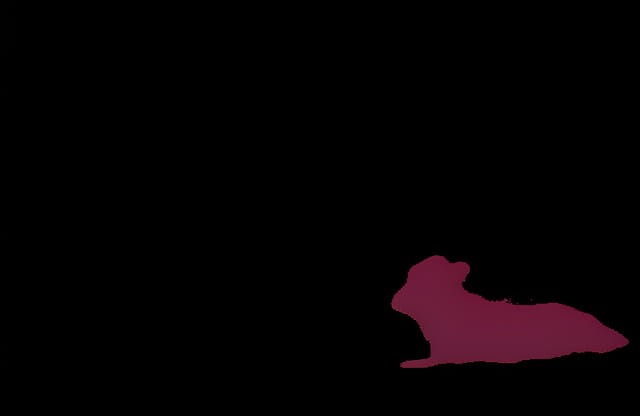} &
\simg{} &
\simg{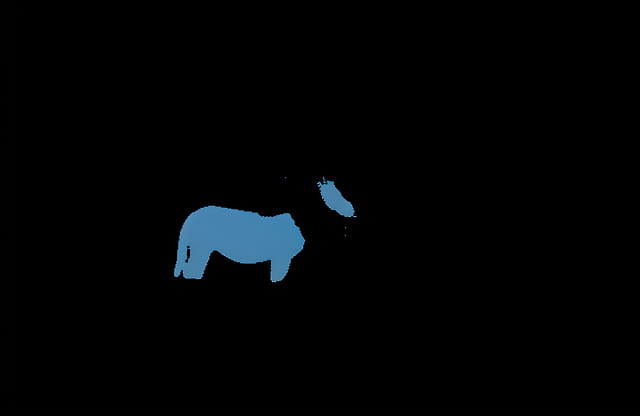} \\[3pt]
\multicolumn{2}{@{}>{\centering\arraybackslash}p{\supppairw}@{}}{\prompttext{The front most cow to the right of other cows.}} &
\multicolumn{2}{@{}>{\centering\arraybackslash}p{\supppairw}@{}}{\prompttext{A zebra that is on the left and is not completely visible.}} \\[6pt]

\textbf{Input} & \textbf{Prediction} & \textbf{Input} & \textbf{Prediction} \\[2pt]
\simg{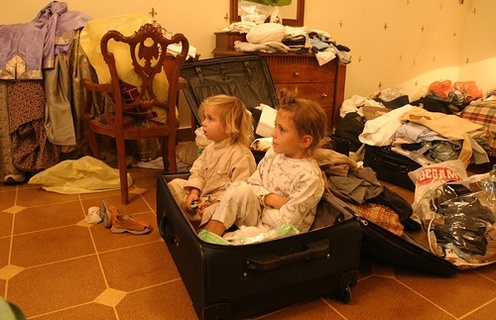} &
\simg{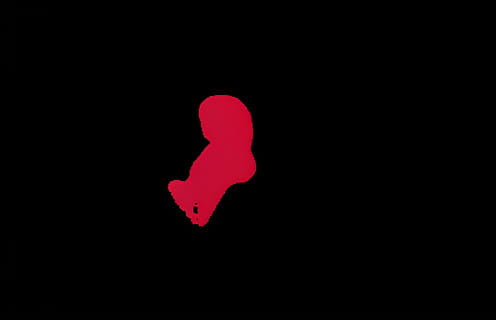} &
\simg{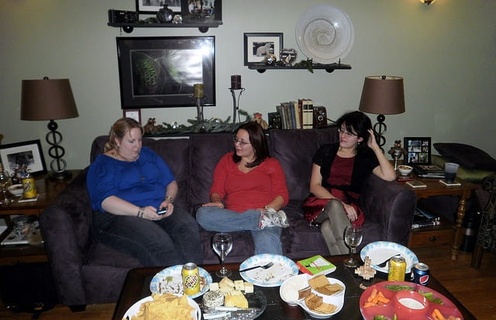} &
\simg{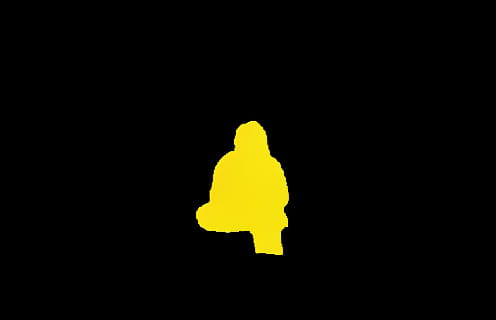} \\[3pt]
\multicolumn{2}{@{}>{\centering\arraybackslash}p{\supppairw}@{}}{\prompttext{Little girl sitting with toy in her hand.}} &
\multicolumn{2}{@{}>{\centering\arraybackslash}p{\supppairw}@{}}{\prompttext{the woman in the middle sitting on the couch.}}

\end{tabular}

\caption{\textbf{General referring-segmentation results on RefCOCOg with the non-zero-shot \model{} 9B model.}
Each example shows the input image and raw predicted mask, with its referring expression below.}
\label{fig:supp-qual-refcocog-nonzero}
\end{figure*}

}
\vspace{8pt}
\ovbplacegallery{\textwidth}{
\begin{figure*}[p]
  \centering
  \includegraphics[width=\linewidth]{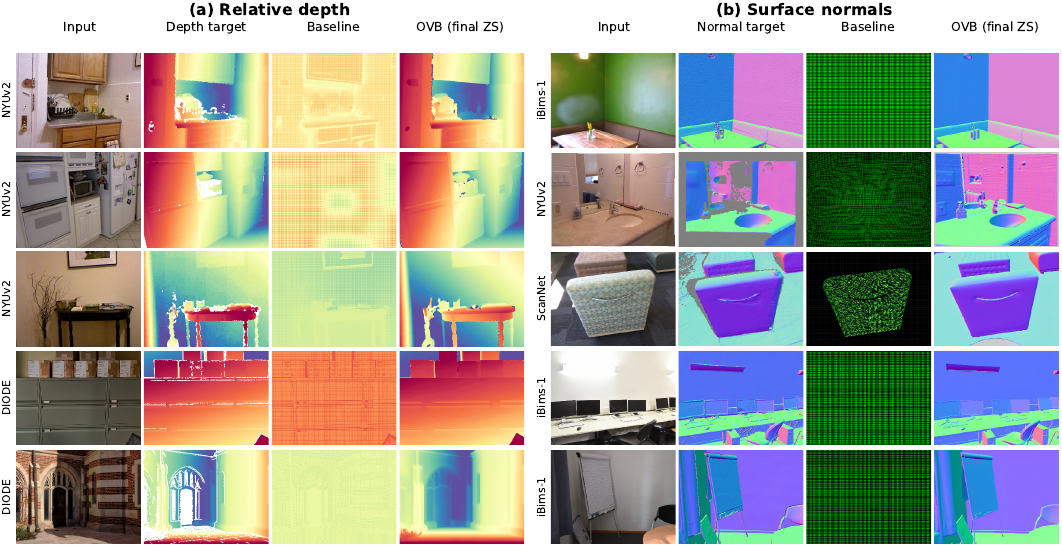}
  \caption{Selected examples of (a) relative depth and (b) surface normals from
  the baseline (pretrained Klein 9B) and final zero-shot OVB 9B.
  Both use the same inputs, task prompts, 28 sampling steps, guidance 3.5
  and seed 0. Depth rows show three NYUv2 and two DIODE examples;
  normal rows show iBims-1, NYUv2, ScanNet, then two more iBims-1 examples.
  Depth predictions use per-image disparity-affine alignment and share a
  Spectral color scale with the target in each row (red near, blue far),
  as in Figure~\ref{fig:final-geometry-qualitative}. White marks ground-truth
  pixels excluded by the evaluation mask; predictions have no validity
  overlay or artifact removal.
  The standard NYUv2 depth crop is applied to every column. Normals retain
  the declared camera-space RGB encoding. Examples were selected to show
  clear geometric structure and are not a random sample.}
  \label{fig:qual-klein-depth-finetuning}
  \label{fig:qual-klein-normal-finetuning}
\end{figure*}

}
\end{ovbgallerypage}
\endgroup

\end{document}